\documentclass[mnsc, nonblindrev, pdflatex]{informs3noheader}
\usepackage[margin=1in]{geometry}
\usepackage{graphicx}				
\usepackage{hyperref}
\usepackage{float}
\usepackage{amssymb}
\usepackage{authblk}
\usepackage{comment}
\usepackage[linesnumbered,ruled,vlined]{algorithm2e}
\SetCommentSty{mycommfont}

\SetKwInput{KwInput}{Input}                % Set the Input
\SetKwInput{KwOutput}{Output}              % set the Output

\usepackage{enumerate}
\usepackage{bbm}
\usepackage[normalem]{ulem}
\useunder{\uline}{\ul}{}

\OneAndAHalfSpacedXI % Current default line spacing

\usepackage[position=b]{subfig}
\usepackage{float}
\usepackage{changepage}
\usepackage{url}

\theoremstyle{TH}
\newcommand{\bea}{\begin{eqnarray}}
\newcommand{\eea}{\end{eqnarray}}
\newcommand{\ben}{\begin{equation*}}
\newcommand{\een}{\end{equation*}}
\newcommand{\bean}{\begin{eqnarray*}}
\newcommand{\eean}{\end{eqnarray*}}


\newcounter{parentnumber}

\newcolumntype{L}[1]{>{\raggedright\arraybackslash}b{#1}}
\newcolumntype{C}[1]{>{\centering\arraybackslash}b{#1}}
\newcolumntype{R}[1]{>{\raggedleft\arraybackslash}b{#1}}
\usepackage{natbib}
 \bibpunct[]{(}{)}{,}{a}{}{,}%
 \def\bibfont{\footnotesize}%
\usepackage{import}
\usepackage{booktabs} % for much better looking tables
\newcommand{\tabitem}{~~\llap{\textbullet}~~}

\usepackage{array} % for better arrays (eg matrices) in maths
\usepackage{paralist} % very flexible & customisable lists (eg. enumerate/itemize, etc.)
\usepackage{verbatim} % adds environment for commenting out blocks of text & for better verbatim
\usepackage{amsfonts}
\usepackage{amsmath}
\usepackage{dsfont}
\usepackage{multirow}
\usepackage{accents}
\usepackage{graphicx}
\usepackage{caption}
\usepackage[usenames,dvipsnames,svgnames,table]{xcolor}
\usepackage{mathtools}
\usepackage[framemethod=TikZ]{mdframed}
\usepackage{pdflscape}
\usepackage{epigraph} % for famous quotes
\usepackage{etoolbox}

\makeatletter
\patchcmd{\epigraph}{\@epitext{#1}}{\itshape\@epitext{#1}}{}{}
\makeatother

\usepackage{framed}

\definecolor{formalshade}{rgb}{0.95,0.95,1}

\TheoremsNumberedThrough     % Preferred (Theorem 1, Lemma 1, Theorem 2)
\ECRepeatTheorems

\EquationsNumberedThrough    % Default: (1), (2), ...
\MANUSCRIPTNO{}

\begin{document}

\RUNAUTHOR{Bastani, Bastani, and Sinchaisri}
\RUNTITLE{Improving Human Sequential Decision-Making with Reinforcement Learning}

% Full title.
\TITLE{Improving Human Sequential Decision-Making \\ with Reinforcement Learning}
\ARTICLEAUTHORS{%
\AUTHOR{Hamsa Bastani}
\AFF{Wharton School, Operations Information and Decisions, \EMAIL{hamsab@wharton.upenn.edu}}

\AUTHOR{Osbert Bastani}
\AFF{University of Pennsylvania, Computer and Information Science, \EMAIL{obastani@seas.upenn.edu}}

\AUTHOR{Wichinpong Park Sinchaisri}
\AFF{University of California, Berkeley, Haas School of Business, \EMAIL{parksinchaisri@haas.berkeley.edu}}
}

\ABSTRACT{
Workers spend a significant amount of time learning how to make good decisions. Evaluating the efficacy of a given decision, however, can be complicated---e.g., decision outcomes are often long-term and relate to the original decision in complex ways. Surprisingly, even though learning good decision-making strategies is difficult, they can often be expressed in simple and concise forms. Focusing on sequential decision-making, we design a novel machine learning algorithm that is capable of extracting ``best practices" from trace data and conveying its insights to humans in the form of interpretable ``tips". Our algorithm selects the tip that best bridges the gap between the actions taken by human workers and those taken by the optimal policy in a way that accounts for which actions are consequential for achieving higher performance. We evaluate our approach through a series of randomized controlled experiments where participants manage a virtual kitchen. Our experiments show that the tips generated by our algorithm can significantly improve human performance relative to intuitive baselines. In addition, we discuss a number of empirical insights that can help inform the design of algorithms intended for human-AI interfaces. For instance, we find evidence that participants do not simply blindly follow our tips; instead, they combine them with their own experience to discover additional strategies for improving performance.
}

\maketitle

\section{Introduction}

Workers spend a significant amount of time on the job learning how to make good decisions that improve their performance~\citep{chui2012social}. Yet, the impact of a current decision can be long-range---affecting future decisions/rewards in complex ways---making it difficult for them to evaluate the quality of a decision. This is exacerbated by the fact that multiple decisions are often made sequentially, making it hard to determine which decisions are responsible for good outcomes even in hindsight. Many jobs require such sequential decision-making, e.g., doctors ordering tests to optimize patient outcomes~\citep{kleinberg2015prediction}, or workers choosing jobs on gig economy platforms to optimize their daily profits~\citep{marshall2020, allon2023managing}. As a concrete example, physicians seek to learn good strategies for ordering lab tests, since obtaining test results in a timely fashion is necessary to minimize delays in patient visits; for instance, \cite{song2017closing} finds that experienced physicians have learned to order these tests early to avoid delays. Despite the simple description of the strategy---``order lab and radiology tests as early in the care delivery process as possible''---learning it on the job can be difficult because the connection between when tests are ordered and the overall quality of care is influenced by numerous other decisions made by the physician, as well as unrelated changes in the underlying environment (e.g., hospital congestion).

Learning on the job can significantly impact service quality, since workers likely make sub-optimal decisions during this time. For instance, when surgeons first use new devices, surgery duration increases by roughly a third, which can be costly to both patients and providers~\citep{ramdas2017variety}. Thus, when possible, workers seek alternative ways to acquire best practices on decision-making. Continuing our example on physician decisions for lab testing,~\cite{song2017closing} finds that physicians can learn strategies for reducing service time from their better-performing colleagues. This approach is effective precisely because the strategy is simple and easy to communicate, yet time-consuming to discover independently. However, learning from their peers is not always an option; for instance, some workers are comparatively isolated---e.g., physicians working in rural hospitals or independent workers in the gig economy. In these cases, workers must wastefully spend time independently rediscovering best practices that are already known to their colleagues.

Thus, a natural question arises: can we \emph{automatically} discover best practices and convey them to workers to help them improve their performance? In particular, over the past two decades, many domains have accumulated large amounts of \emph{trace data} on human decisions. For example, nearly every physician action is logged in electronic medical record data; every movement of a driver is recorded by gig economy platforms; even retail manager decisions on pricing and inventory management are recorded on a daily basis. This data implicitly encodes the collective knowledge acquired by numerous workers about how to effectively perform their jobs. However, trace data is often extremely noisy, granular, and of tremendous volume, rendering it unreadable to humans. At the same time, recent advances in reinforcement learning have enabled machines to achieve human-level or super-human performance at many challenging sequential decision-making tasks~\citep{mnih2015human, silver2016mastering}. Thus, we might hope to leverage these techniques to mine high-volume trace data to automatically identify key bottlenecks in current human decision-making, as well as promising tips/advice to improve their performance.

In this paper, we perform a large-scale behavioral experiment to study whether reinforcement learning can be used to infer tips that improve human performance in sequential decision-making tasks. There is now a large body of evidence that machine learning predictions can improve human performance in \textit{one-shot} decision-making---where the current decision does not affect future outcomes---e.g., bail decisions~\citep{green2019principles}, visual question answering~\citep{chandrasekaran2017takes,chandrasekaran2018explanations}, satellite image analysis~\citep{kneusel2017improving}, and detecting deceptive reviews~\citep{lai2019human}. In these settings, it often suffices to provide the model's prediction to the user, potentially in an interpretable way to improve trust and compliance. However, sequential decision-making settings pose qualitatively different challenges, since current decisions can have long-term consequences and affect future observed states. In particular, we must figure out in \textit{which} states we should intervene, which can be informed by examining bottlenecks in the current human policy. To this end, we devise a novel algorithmic framework for inferring simple tips that, if adopted, can improve the performance of the worker. Our algorithm aims to capture the \textit{discrepancy} between the existing human policy (as captured by historical trace data) and the optimal policy, which helps us identify the most performance-improving tips for key bottlenecks in current human decision-making.

An additional challenge in sequential decision-making is that for these tips to improve performance, the human needs to understand how to operationalize them into their broader workflow. Otherwise, even if they comply with the tip, there is no guarantee that they correctly understand what decisions to make on other time steps to achieve optimal performance. In principle, even if a tip suggests optimal actions for the worker to take, and the worker complies with the tip perfectly, the overall performance could degrade since the worker subsequently makes poor decisions. Thus, our search space of candidate tips must focus on interpretable and actionable information that workers can easily operationalize. Whether humans can actually do so is an empirical question; thus, we conduct a large-scale behavioral experiment that studies how humans perceive and improve their own decision-making over time (given tips from either our algorithm, or via peer feedback or simple descriptive statistics), how they adjust other portions of their workflow to accommodate these changes, and how humans may incorrectly perceive bottlenecks in their own decision-making.

To summarize, two criteria are needed to \textit{actually} improve human decision-making. First, our algorithm must identify sufficiently useful tips to improve performance (assuming humans comply with and effectively operationalize them). Second, humans must be able to understand and comply with our tip and, furthermore, effectively operationalize it by modifying their broader workflow.

\paragraph{Algorithm.} Our algorithm builds on the idea of model distillation~\citep{bucilua2006model,hinton2015distilling} for interpretable reinforcement learning~\citep{verma2018programmatically,bastani2018verifiable}, which involves first training a blackbox decision-making policy using reinforcement learning~\citep{sutton2018reinforcement}, and then training an interpretable policy to approximate the blackbox policy. However, unlike prior work, our goal is to infer an interpretable tip that best minimizes the discrepancy between the existing human policy and the blackbox policy, rather than to train the best-performing interpretable policy that is agnostic to the current human policy.
Thus, the chosen tip is tailored to current bottlenecks in the human decision-making policy, and accounts for which actions are consequential for achieving higher performance---i.e., following the tip is expected to improve the long-term performance of the human rather than to simply mimic the optimal policy.
In order to easily convey our insights to humans, we design the search space over tips to consist of if-then-else rules. Despite their simplicity, we find that these tips can capture useful insights that are challenging for humans to learn by themselves in complex sequential decision-making problems.

\paragraph{Game.} To study these issues, we designed and built a sequential decision-making game where human players manage a virtual kitchen, inspired by the popular game \emph{Overcooked}. Our game is based on the discrete-time job shop scheduling problem, where tasks need to be scheduled to virtual workers; each task consists of subtasks with dependencies (e.g., ingredients must be chopped before cooking) and workers have heterogeneous processing times (e.g., a chef is better at cooking, a server is better at plating). Players must assign subtasks to virtual workers in a way that minimizes the time it takes to complete a set of food orders. Our game is deterministic, making it easy for inexperienced players to learn the optimal strategy from a few interactions. Instead, the difficulty in achieving good performance comes from the game's combinatorial state space, encoding worker availability and subtask completion so far. For instance, they must make forward-looking trade-offs, e.g., deciding whether to greedily assign a worker to a subtask that they are slow to complete, or to leave them idle in anticipation of a more suitable subtask.

Our game captures challenges in a variety of operations problems encountered in the real world. For instance, when assigning tasks to health workers, there can be substitution when patient traffic is high, such as having a nurse practitioner perform tasks usually done by physicians. Another example is delivery workers on a grocery delivery platform choosing which orders to accept, where the worker must account for dependencies (e.g., orders must be picked up before delivery) as well as heterogeneous service times (e.g., bikers have an advantage over drivers in high-traffic locations). More broadly, our game can be viewed as a stylized model of any manager scheduling employees to perform tasks on a daily basis, a gig economy employee scheduling their daily workload, or a project manager assigning subtasks to workers to accomplish a longer-term goal. While these examples typically involve more complex challenges such as stochastic demands, we believe our experimental findings on worker learning and compliance can generalize well to these settings.

\begin{figure}[!htpb]
\centering
\includegraphics[width=0.95\textwidth]{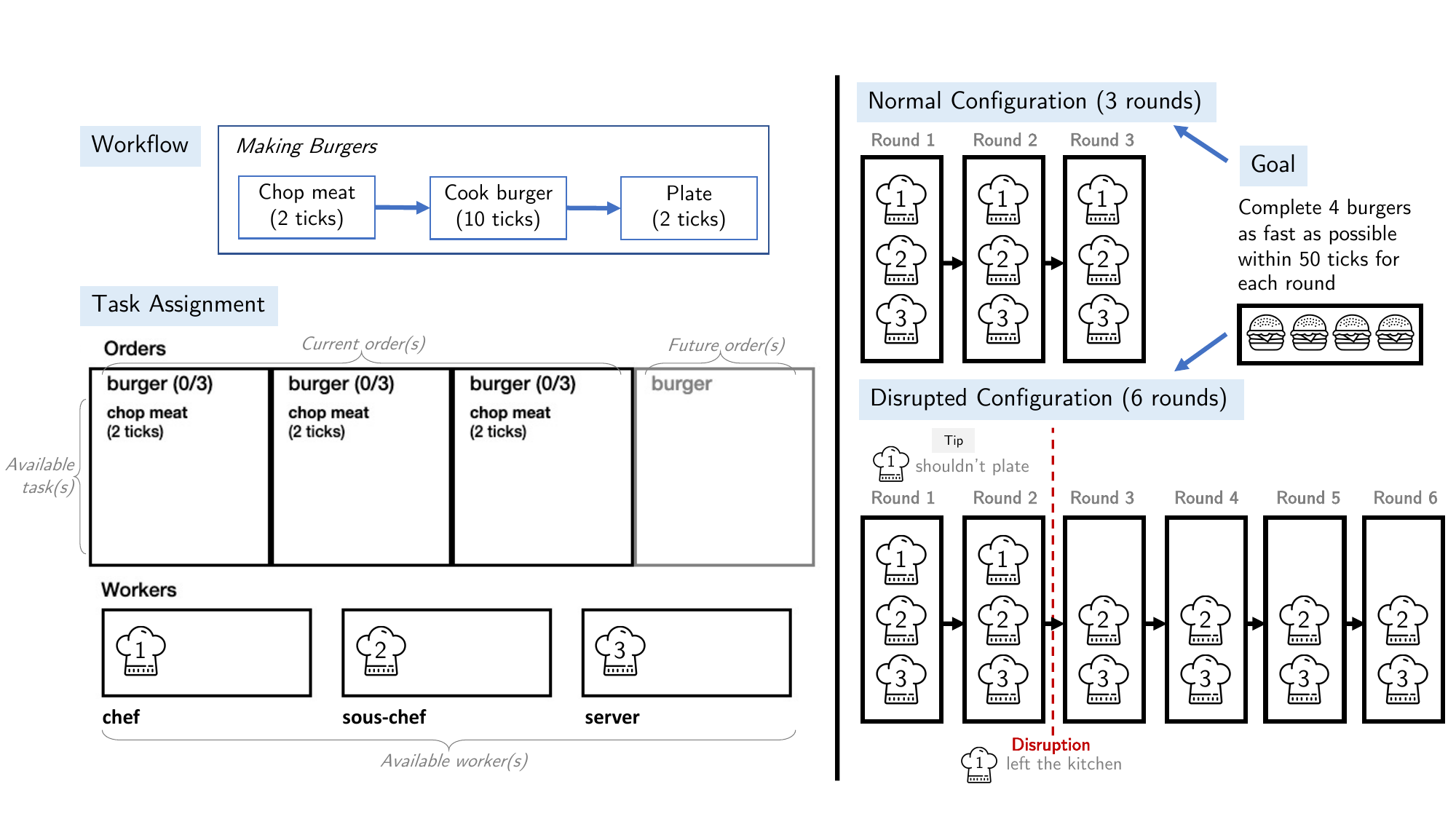}
\caption{Overview of kitchen management game. The left panel depicts what participants see: (i) the workflow required to complete a burger order, and (ii) the game screen that allows available tasks to be dragged and dropped to one of 3 virtual workers. The right panel depicts the study design: in the normal configuration, participants play the same game for 3 rounds; in the disrupted configuration, participants play the same game for 2 rounds, face a disruption in the kitchen (i.e., the chef leaves), and play the disrupted game for 4 rounds.}
\label{fig:gameflow}
\end{figure}

\paragraph{Experiment.} Our primary contribution is a large-scale randomized controlled experiment in the context of this game; Figure~\ref{fig:gameflow} illustrates the high-level setup and flow of the game and Section~\ref{sec:casestudy} provides a more detailed description. In particular, we perform a large-scale behavioral study on Amazon Mechanical Turk based on two different configurations of our virtual kitchen environment. In the \emph{normal} configuration, the participant plays three identical instantiations of the environment. In the \emph{disrupted} configuration, the first two instantiations of the environment are identical to the ones in the normal configuration, but the remaining four instantiations are modified so that a key worker (namely, the chef) is no longer available. These two configurations are visualized in the right panel of Figure~\ref{fig:gameflow}. The disrupted configuration is particularly challenging for the human participants, since they must un-learn preconceived notions about the optimal strategy acquired during the first two instatiations. For each of these configurations, we leverage our algorithm to learn interpretable tips, and then demonstrate how providing this decision-making rule improves the performance of the participants. 
Our results demonstrate that our algorithm can generate valuable insights that enable human participants to substantially improve their performance compared to counterparts that are not shown the tip or that are shown alternative tips derived from natural baselines. Importantly, we observe that participants do not naively adjust their policy by blindly following the tip. Instead, as they gain experience with the game, they increasingly understand the significance of the tip and improve their performance in ways beyond the surface-level meaning of the tip. Overall, our findings suggest that reinforcement learning can effectively leverage trace data to infer interpretable and useful insights, and furthermore, can successfully convey these insights to humans to improve their decision-making.

\subsection{Related Literature}\label{sec:relatedlit}

\paragraph{Identifying performance improvements for human workers.}

Process improvement has long been a focal point in operations management; scholars have especially identified various difficulties associated with sequential decision-making and learning. Thus, we study process improvement from the perspective of individual workers through sequential decision-making. 
When workers first experience a new work environment, they may have difficulty adjusting, resulting in various degrees of undesirable performance \citep{ramdas2017variety}, e.g., unexpected critical medical incidents slow down ambulance activation among paramedics \citep{bavafa2021recovering}.
The situation is exacerbated when inexperienced workers lack guidelines on how to manage their workflow, resulting in sub-optimal task prioritization and poor productivity~\citep{ibanez2018discretionary}. % lack of guideline v.nbvrf. g5
Complexity of workflows also plays a role. Workers tend to focus on immediate challenges and ignore opportunities for learning~\citep{tucker2002problem}; furthermore, switching between tasks can significantly hurt productivity~\citep{gurvich2020collaboration}.
Depending on the features of the sequential decision-making problem, workers may generally follow non-optimal policies~\citep{kagan2021dynamic}.

A common approach to increase reliability and reduce process variation is to standardize processes and offer best practices~\citep{nonaka1995knowledge, pfeffer2000knowing, spear2005fixing}. However, creating standards can be challenging \citep{szulanski1996exploring, argote2012organizational} and time-consuming~\citep{nonaka1995knowledge}. Workers can learn by trial and error~\citep{dorn2010learning}, but past experience sometimes makes it challenging to identify best practices~\citep{huckman2006firm, kc2012accumulating}. Workers can also learn through soliciting peer feedback~\citep{herkenhoff2018knowledge, jarosch2019learning, song2017closing, brattland2018learning} or working alongside experienced peers \citep{chan2014learning, tan2019you}; these mechanisms are especially salient when there is familiarity and collaborative experience between workers \citep{akcsin2021learning, kim2020admission}. However, these ingredients are often not available. Given well-documented difficulties in learning on the job and identifying best practices, our work proposes an effective approach to automatically extract best practices from logged trace data of historical decisions and outcomes. While recent work has leveraged trace data and machine learning to predict when humans make mistakes in decision-making~\citep{fudenberg2019predicting, mcilroy2020aligning, fudenberg2022measuring}, they do not offer tips to improve human performance.

\paragraph{Using machine learning to improve one-shot decision-making.}
As noted earlier, several recent papers have studied whether machine learning can improve human decision-making in the one-shot setting. Key challenges that arise are that humans often erroneously assess their own abilities \citep{fugener2022cognitive} as well as the predictive model's abilities \citep{chandrasekaran2017takes, chandrasekaran2018explanations, green2019principles}; this in turn can result in unwarranted algorithm aversion \citep{dietvorst2015algorithm} or algorithm appreciation \citep{logg2019algorithm}. This can be overcome by mechanisms such as enabling the predictive model to delegate tasks to humans in a user-aware manner \citep{fugener2022cognitive}, training workers on the success/failures of their specific predictive model \citep{chandrasekaran2018explanations}, capturing the uncertainty of the model's predictions \citep{kneusel2017improving}, or accounting for systematic human deviations from the model \citep{sun2022predicting}. Another important lever is improving the interpretability/explainability of the predictive model \citep{stites2021sage, lu2019good}, which allows workers to gain a deeper understanding of the environment and the potential improvement to be obtained~\citep{sull2015simple, gleicher2016framework}. This can be accomplished by using simple model families like decision trees \citep{breiman1984classification, bertsimas2017optimal} or rule lists \citep{wang2015falling, letham2015interpretable}, or by employing post-hoc explanation methods like LIME \citep{ribeiro2016should}.

In contrast to these approaches, we focus on sequential decision-making, which is representative of many real-world workflows and poses qualitatively different challenges. For example, adopting a recommended decision on the current time step affects future states/decisions faced by the worker; as a consequence, compliance with a tip may actually \textit{hurt} performance if the worker is unable to appropriately adjust their future workflow. Algorithmically, it is also more challenging to compute interpretable policies, since the entire sequence of recommended decisions needs to be interpretable. Thus, we propose a novel framework that adapts interpretable reinforcement learning techniques \citep{puiutta2020explainable, meyer2014machine} to compute interpretable tips that bridge the discrepancy between the human's current policy and the optimal policy. We build on a strategy that first trains a high-performance blackbox policy, and then use imitation learning~\citep{ross2011reduction} to distill this policy into an interpretable one~\citep{verma2018programmatically,bastani2018verifiable}.

\subsection{Contributions}\label{sec:contributions}

Our work contributes to the literature in two ways. First, we propose a novel algorithm for inferring tips for sequential decision-making. Our algorithm leverages techniques from interpretable reinforcement learning to capture the discrepancy
between the existing human policy (as captured by trace data) and the optimal policy, thereby identifying the best performance-improving tip targeted towards key bottlenecks in current human decision-making.

Second, to the best of our knowledge, we conduct the first large-scale behavioral experiment on Amazon Mechanical Turk to understand how reinforcement learning based tips can improve human performance in sequential decision-making problem. Unlike one-shot decision-making, in order to be effective, humans must understand not only the meaning of a tip, but also how to operationalize it into a broader workflow. Our experimental results demonstrate that workers are capable of inferring complex strategies from the limited recommendations provided by our algorithm's tips, but this is not always the case with tips inferred through peer feedback or simple descriptive statistics. We also provide a number of additional insights about how workers comply with tips, as well as how they perceive bottlenecks in their own workflows.

\section{Inferring Tips via Interpretable Reinforcement Learning}

Consider a human making a sequence of decisions to achieve some desired outcome. We study settings where current decisions affect future outcomes---for instance, if the human decides to consume some resources at the current time step, they can no longer use these resources in the future. These settings are particularly challenging for decision-making due to the need to reason about how current actions affect future decisions, making them ideal targets for leveraging tips to improve human performance. %In particular, our goal is to provide insights to the human that enable them to improve their performance.

We begin by formalizing the tip inference problem. We model our setting as the human acting to maximize reward in a standard, undiscounted Markov Decision Process (MDP) $\mathcal{M}=(S,A,R,P)$ over a finite time horizon $T$. Here, $S$ is the state space, $A$ is the action space, $R$ is the reward function, and $P$ is the transition function. Intuitively, a state $s\in S$ captures the current configuration of the system (e.g., available resources), and an action $a\in A$ is a decision that the human can make (e.g., consume some resources to produce an item). We represent the human as a decision-making policy $\pi_H$ mapping states to (possibly random) actions. At each time step $t\in\{1,...,T\}$, the human observes the current state $s_t$ and selects an action $a_t$ to take according to the probability distribution $p(a_t\mid s_t)=\pi_H(s_t,a_t)$. Then, they receive reward $r_t=R(s_t,a_t)$, and the system transitions to the next state $s_{t+1}$, which is a random variable with probability distribution $p(s_{t+1}\mid s_t,a_t)=P(s_t,a_t,s_{t+1})$, after which the process is repeated until $t=T$. A sequence of state-action-reward triples sampled according to this process is called a \emph{rollout}, denoted $\zeta=((s_1,a_1,r_1),...,(s_T,a_T,r_T))$. We measure the cumulative expected reward of a given policy $\pi$ as
\begin{align} \label{eq:J}
J(\pi)=\mathbb{E}_{\zeta\sim D^{(\pi)}}\left[\sum_{t=1}^T r_t\right],
\end{align}
where $D^{(\pi)}$ is the distribution of rollouts induced by using policy $\pi$. We denote the human policy $\pi_H$, which is not directly observed but can be estimated from historical trace data. It will also be useful to define the optimal policy, $\pi^* = \arg\max_{\pi} J(\pi)$, which maximizes cumulative reward.

\paragraph{Tips:} Now, given the MDP $\mathcal{M}$ and the human policy $\pi_H$, our goal is to learn a tip $\rho$ that, conditioned on adoption by the human, most improves the cumulative expected reward. Formally, a tip indicates that in certain states $s$, the human should use action $\rho(s)\in A$ instead of following their own policy $\pi_H$. Thus, we consider tips in the form of a single, interpretable rule:
\begin{align*}
\rho(s)=\text{if }\psi(s),\text{ then take action }a,
\end{align*}
where $a\in A$ is an action and $\psi(s)\in\{\text{true},\text{false}\}$ is a logical predicate over states $s\in S$ (e.g., $\psi(s)$ might be an indicator of whether a sufficient quantity of a certain resource is currently available). In other words, a tip $\rho=(\psi,a)$ says that if the logical predicate $\psi$ is true, then the human should use the action $a$ prescribed by the tip; otherwise, they should use their own policy $\pi_H$. 

If the human follows this tip exactly, then the resulting policy they use is $\pi_H\oplus\rho$, where we define the operation
\begin{align*}
(\pi\oplus\rho)(s,a')=
\begin{cases}
\mathbbm{1}(a'=a)&\text{if}~\psi(s) \\
\pi(s,a')&\text{otherwise}.
\end{cases}
\end{align*}
Here, $\mathbbm{1}$ is the indicator function; that is, the human takes action $a$ with probability one if $\psi(s)$ holds, and follows their existing policy otherwise.

\begin{remark}
In practice, we find that human adoption of tips varies. However, it is difficult to predict the rate of adoption of a tip prior to offering it. Instead, we focus on identifying the best performance-improving tip \textit{conditioned} on adoption. We find that this strategy works sufficiently well to improve performance in our experiments as long as the human can understand both the tip and its rationale. We give a detailed discussion of compliance with tips in Section~\ref{sec:compliance}. 
\end{remark}

Our goal is to compute the tip $\rho^*$ that most improves the human's performance---i.e.,
\begin{align}
\label{eqn:problem}
\rho^*=\operatorname*{\arg\max}_{\rho}J(\pi_H\oplus\rho).
\end{align}
This formulation ensures that the chosen tip is \textit{consequential} to improving performance $J$ in Eq.~\eqref{eq:J}. There are many other ways to choose tips, e.g., one can na\"{i}vely identify state-action pairs that frequently differ between the human and optimal policies. We illustrate the drawbacks to such an approach in our experiments (see Section~\ref{sec:results}).

\paragraph{Algorithm:} Next, we describe our algorithm for solving Eq.~\eqref{eqn:problem}. Note that we can simply loop through each candidate tip $\rho$, but we may lack the data to evaluate $J(\pi_H\oplus \rho)$ without additional assumptions. This is because showing the tip changes the human's behavior, changing the distribution of states $D^{(\pi)}$ they visit to $D^{(\pi\oplus\rho)}$. However, we do not have samples from $D^{(\pi\oplus\rho)}$, which are necessary to estimate Eq.~\eqref{eq:J}. One strategy would be to run an experiment with each tip to obtain these samples, but this is prohibitively expensive. Alternatively, one can consider approximating the unobserved distribution $D^{(\pi\oplus\rho)}$ with the observed distribution $D^{(\pi)}$ when evaluating $J(\pi_H\oplus \rho)$, but this has the unfortunate consequence of removing the dependence on the tip $\rho$ entirely from our optimization problem in Eq.~\eqref{eqn:problem}, rendering us unable to identify good tips.

Instead, we describe an approximation that is implementable given observed data, and effectively distinguishes between candidate tips; we find that this strategy works well in our experiments. To this end, we leverage the well-studied value- and $Q$-functions~\citep{watkins1992q} (denoted $V^*$ and $Q^*$, respectively), which can be defined recursively by the Bellman equation:
\begin{align*}
V^*(s) &= \max_{a\in A}Q^*(s,a), \\
Q^*(s,a) &= R(s,a)+\mathbb{E}_{s'\sim p(\cdot\mid s,a)}[V^*(s')].
\end{align*}
Intuitively, $V^*(s)$ is the cumulative expected reward accrued from state $s$ when using the optimal policy, and $Q^*(s,a)$ is the cumulative expected reward accrued from $s$ by first taking action $a$ and then using the optimal policy. We can compute both $V^*$ and $Q^*$ using $Q$-learning~\citep{watkins1992q}. Now, we can rewrite the objective $J(\pi_H\oplus\rho)$ in Eq.~\eqref{eqn:problem} as follows:
\begin{lemma}[Lemma 2.2, \citealp{bastani2018verifiable}]
For any policy $\pi$, we have
\begin{align*}
J(\pi^*)-J(\pi) &= \mathbb{E}_{\zeta\sim D^{(\pi)}}\left[\sum_{t=1}^TV_t^*(s_t)-Q_t^*(s_t,\pi(s_t))\right].
\end{align*}
\end{lemma}
%Then, we can decompose the gap in performance between $\pi_H\oplus\rho$ and $\pi_H$ into two terms:
Applying this lemma to both $\pi_H$ and $\pi_H\oplus\rho$, and taking the difference, we obtain
\begin{align*}
%\label{eq:decomposition}
J(\pi_H\oplus\rho)-J(\pi_H)
&= \mathbb{E}_{\zeta\sim D^{(\pi_H)}}\left[\sum_{t=1}^TV_t^*(s_t)-Q_t^*(s_t,\pi_H(s_t))\right] \\
&\qquad- \mathbb{E}_{\zeta\sim D^{(\pi_H\oplus\rho)}}\left[\sum_{t=1}^TV_t^*(s_t)-Q_t^*(s_t,\pi_H\oplus\rho(s_t))\right]. %&= \underbrace{\mathbb{E}_{\zeta\sim D^{(\pi_H)}}\left[\sum_{t=1}^TQ_t^*(s_t,\pi_H\oplus\rho(s_t))-Q_t^*(s_t,\pi_H(s_t))\right]}_{\text{first term}} \\
%&\qquad+ \underbrace{\left[\mathbb{E}_{\zeta\sim D^{(\pi_H)}}\left[\sum_{t=1}^TV_t^*(s_t)\right] - \mathbb{E}_{\zeta\sim D^{(\pi_H\oplus\rho)}}\left[\sum_{t=1}^TV_t^*(s_t)\right]\right]}_{\text{second term}}. \nonumber
\end{align*}
Letting $\bar{D}_t^{(\pi)}$ be the marginal distribution of $s_t$ in the distribution $D^{(\pi)}$ over rollouts, then
\begin{align*}
J(\pi_H\oplus\rho)-J(\pi_H)
&=\sum_{t=1}^T\mathbb{E}_{s_t\sim\bar{D}_t^{(\pi_H)}}\left[V_t^*(s_t)-Q_t^*(s_t,\pi_H(s_t))\right]-\mathbb{E}_{s_t\sim \bar{D}_t^{(\pi_H\oplus\rho)}}\left[V_t^*(s_t)-Q_t^*(s_t,\pi_H\oplus\rho(s_t))\right].
\end{align*}
Now, assuming that $\bar{D}_t^{(\pi_H)}\approx\bar{D}_t^{(\pi_H\oplus\rho)}$, we have
\begin{align}
J(\pi_H\oplus\rho)-J(\pi_H)
&\approx\sum_{t=1}^T\mathbb{E}_{s_t\sim \bar{D}_t^{(\pi_H)}}\left[V_t^*(s_t)-Q_t^*(s_t,\pi_H(s_t))\right]
-\mathbb{E}_{\zeta\sim \bar{D}_t^{(\pi_H)}}\left[V_t^*(s_t)-Q_t^*(s_t,\pi_H\oplus\rho(s_t))\right] \nonumber \\
&=\mathbb{E}_{\zeta\sim D^{(\pi_H)}}\left[\sum_{t=1}^TQ_t^*(s_t,\pi_H\oplus\rho(s_t))-Q_t^*(s_t,\pi_H(s_t))\right]. \label{eq:actioneffect}
\end{align}
Intuitively, this assumption says that the \textit{indirect} effect on performance due to the shift in the state distribution induced by the tip (i.e., from $\bar{D}_t^{(\pi_H)}$ to $\bar{D}_t^{(\pi_H\oplus\rho)}$) is small; instead, the main effect is due to the \textit{direct} effect on performance due to the change in the current human action induced by the tip, which is captured by Eq.~\eqref{eq:actioneffect}. In practice, we do not observe that the state distributions shift substantially, suggesting that this is a good approximation.
%thus, we focus on the first term:
%\begin{align*}
%J(\pi_H\oplus\rho)-J(\pi_H)
%&\approx
%\mathbb{E}_{\zeta\sim D^{(\pi_H)}}\left[\sum_{t=1}^TQ_t^*(s_t,\pi_H\oplus\rho(s_t))-Q_t^*(s_t,\pi_H(s_t))\right].
%\end{align*}
%

Next, we approximate the expectation in our objective using observed rollouts (i.e., historical trace data) $\zeta_1,...,\zeta_k\sim D^{(\pi_H)}$ from the human policy $\pi_H$. Thus, our algorithm computes the tip
\begin{align}
\label{eqn:algo}
\hat\rho = \operatorname*{\arg\max}_{\rho}\frac{1}{k}\sum_{i=1}^k\sum_{t=1}^T Q^*(s_{i,t}, (a_{i,t}\oplus\rho)(s_{i,t})).
\end{align}
Here, we have dropped the terms $J(\pi_H)$ and $\mathbb{E}_{\zeta\sim D^{(\pi_H)}}\left[\sum_{t=1}^TQ_t^*(s_t,\pi_H(s_t))\right]$ since they are constant in $\rho$; for a given tip $\rho=(\psi,a)$ and action $a'$, we have also defined the operation
\begin{align*}
(a'\oplus\rho)(s)=
\begin{cases}
a&\text{if}~\psi(s)=1 \\
a'&\text{otherwise.}
\end{cases}
\end{align*}
We optimize Eq.~\eqref{eqn:algo} by enumerating through candidate tips $\rho$, evaluating the objective, and selecting the tip $\hat\rho$ with the highest objective value.
%\begin{remark}
%We can weaken the assumption that the first term in Eq.~\eqref{eqn:decomposition} is small by iteratively gathering more data using the optimal tip $\hat\rho$, and then recomputing the optimal tip. When paired with dataset aggregation~\cite{TODO}, this strategy can provably converge to a good tip. This strategy is cheaper than the naive strategy of gathering additional data using every single tip since it only gathers data for a few iterations. However, we found that in practice, such a strategy was unnecessary.
%\end{remark}

\section{Virtual Kitchen Management Game}\label{sec:casestudy}

Our main empirical question is whether human workers can incorporate tips inferred using our algorithm into their broader decision-making policy. Specifically, our tips only provide partial information about the discrepancy between their policy and the optimal policy; thus, workers must not only comply with our tip (which is the usual challenge in improving human performance at one-shot decision-making problems), but they must implicitly infer additional information about the optimal policy in order to effectively operationalize our tip into their broader workflow. To achieve this goal, our environment was designed with two criteria in mind: (i) it should be possible for humans to compute the optimal policy given sufficient thought, but (ii) the optimal policy should not be obvious. We focused on deterministic environments, where inexperienced workers could reason about the optimal strategy from very few interactions with the environment. While we believe our insights extend to stochastic environments, they intuitively require more experience/interactions for humans to deduce optimal strategies. Finally, we deliberately designed a problem where we can compute the optimal policy (see Appendix~\ref{app:opt-policy} for a description of this policy), which enables us to evaluate human sub-optimality.

In particular, we build on the job shop scheduling problem, where the goal is to schedule jobs to machines in an optimal way, and where there are dependencies between different jobs. To ensure the problem is sufficiently challenging, we introduce additional complexity in the form of heterogeneous machines, where the processing time for different types of jobs varies depending on the machine. To make our problem intuitive to human users, inspired by the popular game \emph{Overcooked}, we represented our decision-making problem as a virtual kitchen management game that can be played by individual human players (see Figure \ref{fig:gameflow}). In this game, the player takes the role of a manager of several virtual workers (the ``machines'')---namely, chef, sous-chef, and server---serving burgers in a virtual kitchen. Each burger consists of a fixed set of subtasks (the ``jobs'') that must be completed in order---namely, chopping meat, cooking the burger, and plating the burger. The game consists of discrete time steps; on each time step, the player must decide which (if any) subtask to assign to each idle worker. The worker then completes the subtask across a fixed number of subsequent time steps, and then becomes idle again. A burger is completed once all its subtasks are completed, and the player completes the game once four burger orders are completed. The player's goal is to complete the game in as few time steps as possible.

\begin{figure}[!htpb]
\centering
\subfloat[The initial state where players observe available subtasks, median times to completion, and three idle virtual workers. The interface also shows the current tick, time limit, current progress, and potential tip.]{
\includegraphics[width=0.48\textwidth]{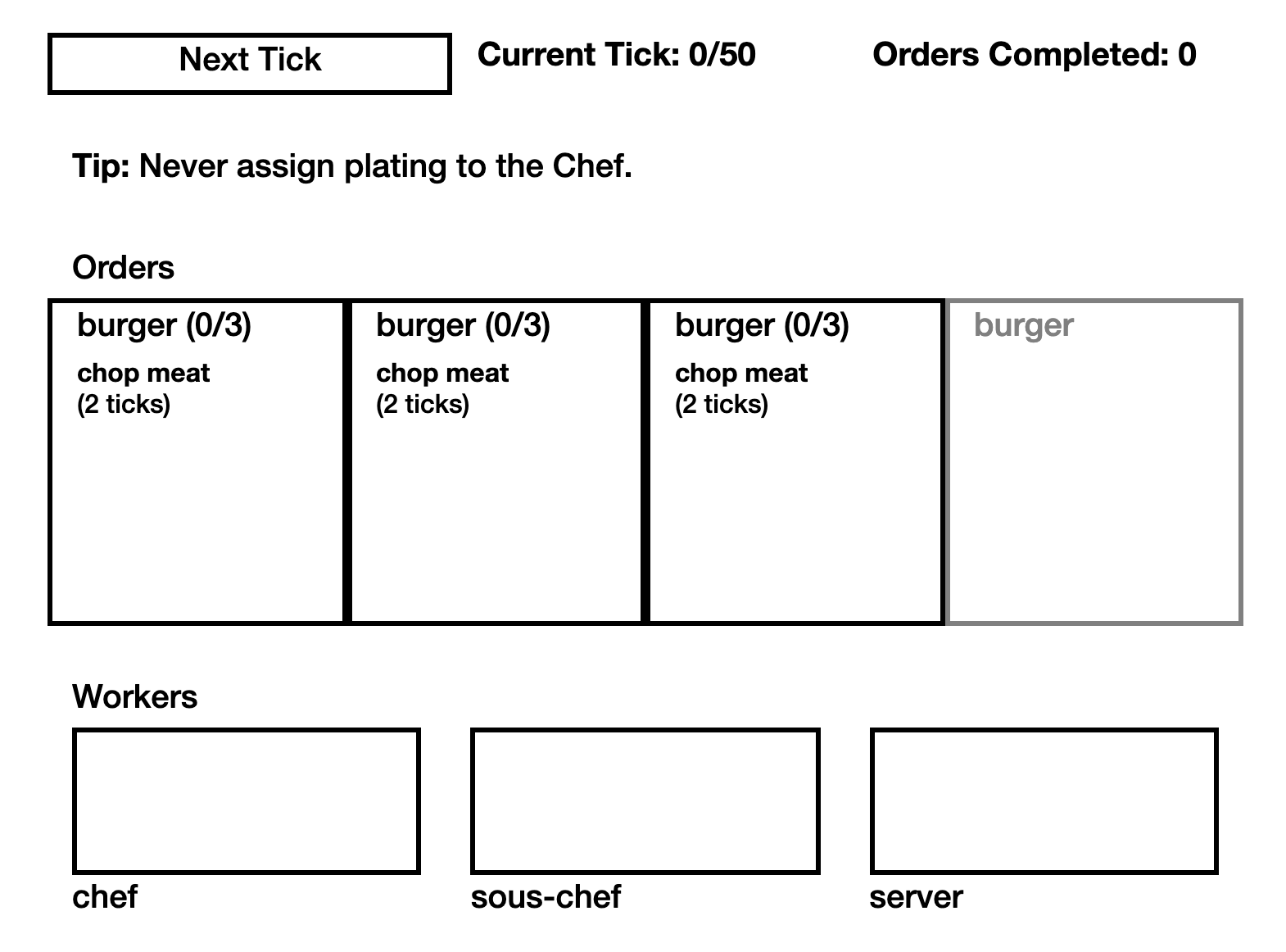}\label{screenshots1}}\quad
\subfloat[The next state after all three previously available subtasks were assigned to the virtual workers and the true completion times were realized, revealing different levels of virtual workers' skills.]{
\includegraphics[width=0.48\textwidth]{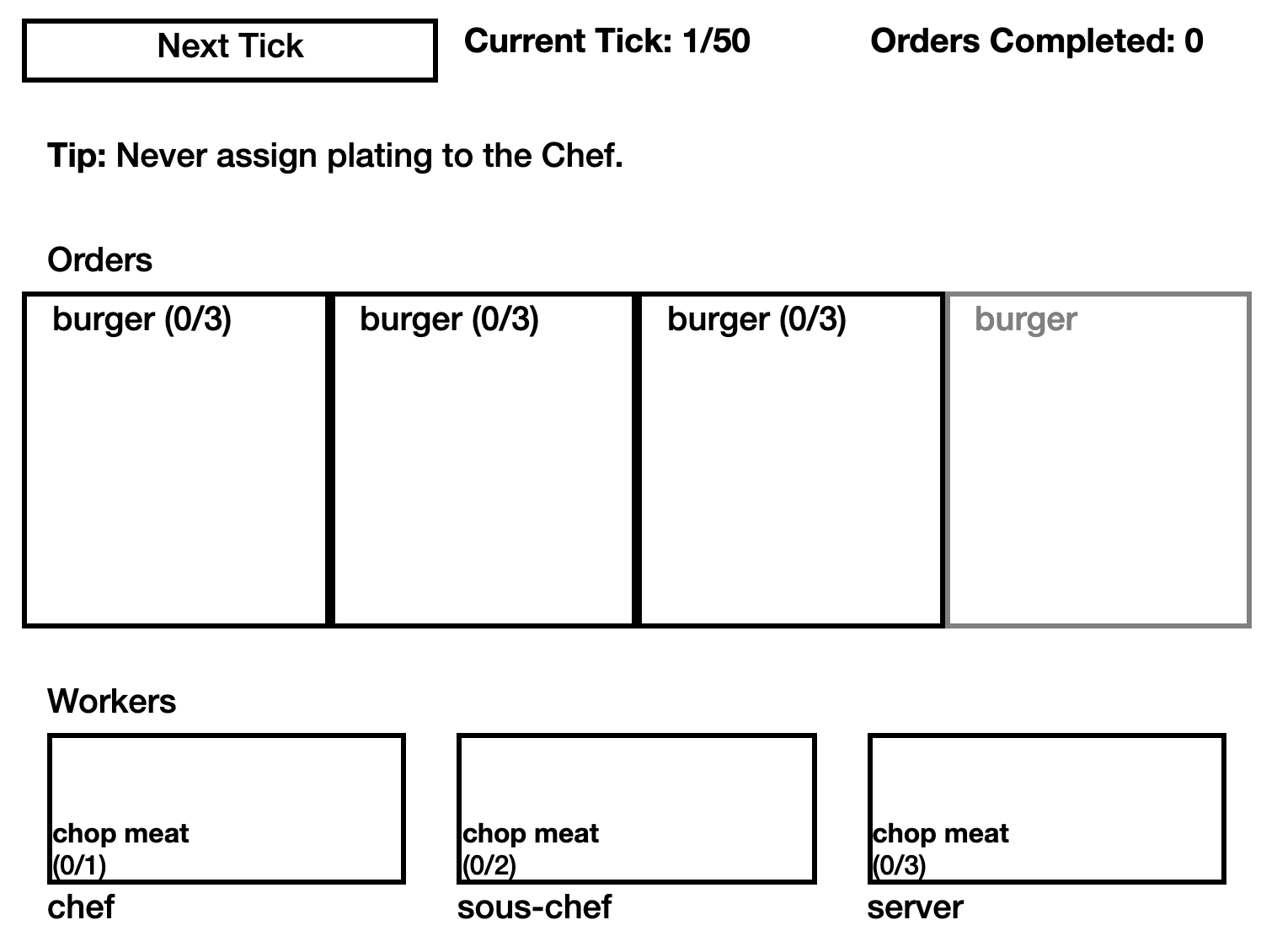}\label{screenshots2}}
\caption{Example screenshots from the game.}\label{fig:screenshots}
\end{figure}

There are two key aspects of the game that make it challenging. First, the subtasks have dependencies---i.e., a subtask can only be assigned once previous subtasks of the same order have already been completed. For example, the ``plate burger'' task can only be assigned once the ``cook burger'' task is completed. Second, the virtual workers have heterogeneous skills---i.e., different workers take different numbers of steps to complete different subtasks. For example, the chef is skilled at chopping/cooking but performs poorly at plating, while the server is the opposite, and the sous-chef has average skill on all subtasks; see Table \ref{tab:skillmatrix} in Appendix~\ref{app:exp-design} for details. Ideally, one would match workers to tasks that they are skilled at to reduce completion time. Thus, the player faces the following dilemma. When a worker becomes available but is not skilled at any of the currently available subtasks, then the player must decide between (i) assigning a suboptimal subtask to that worker, potentially creating a bottleneck, or (ii) leaving the worker idle until a more suitable subtask becomes available. For instance, if the server is idle but all available subtasks are ``cook burger'', then the player must either (i) assign cooking to the unskilled server, thereby slowing down completion of that burger and eliminating the possibility of assigning plating to the server for the near future, or (ii) leave the server idle until a ``plate burger'' subtask becomes available. Furthermore, players are not shown the number of steps a worker takes to complete a subtask until they assign the subtask to that worker (see Figure~\ref{fig:screenshots} and Appendix~\ref{app:screenshots} for example game screenshots); instead, they must experiment to learn this information. 

We consider two scenarios of the game, differing only in terms of worker availability. In the first scenario, the kitchen is \emph{fully-staffed}, where the human player has access to all three virtual workers (chef, sous-chef, and server). In the second scenario, the human player faces a disruption and the kitchen becomes \emph{understaffed}, with only two virtual workers (sous-chef and server). In both scenarios, the goal is to complete four burgers in as few time steps as possible. We describe how this decision-making problem can be formulated as an MDP and the resulting optimal policies in Appendix~\ref{supp-algo}. Note that the optimal policy completes four burgers in 20 and 34 time steps for the fully-staffed and understaffed scenarios, respectively.

\section{Experimental Design}

% overview
We investigate how humans interpret and follow the tips inferred by our algorithm in the context of our virtual kitchen management game, using pre-registered behavioral experiments involving Amazon Mechanical Turk (AMT) workers.\footnote{The full pre-registration document for our study is available at {\footnotesize \url{https://aspredicted.org/blind.php?x=8ye5cb}}} We describe our experimental design in this section.

\begin{figure*}
\centering
\includegraphics[width=0.95\textwidth]{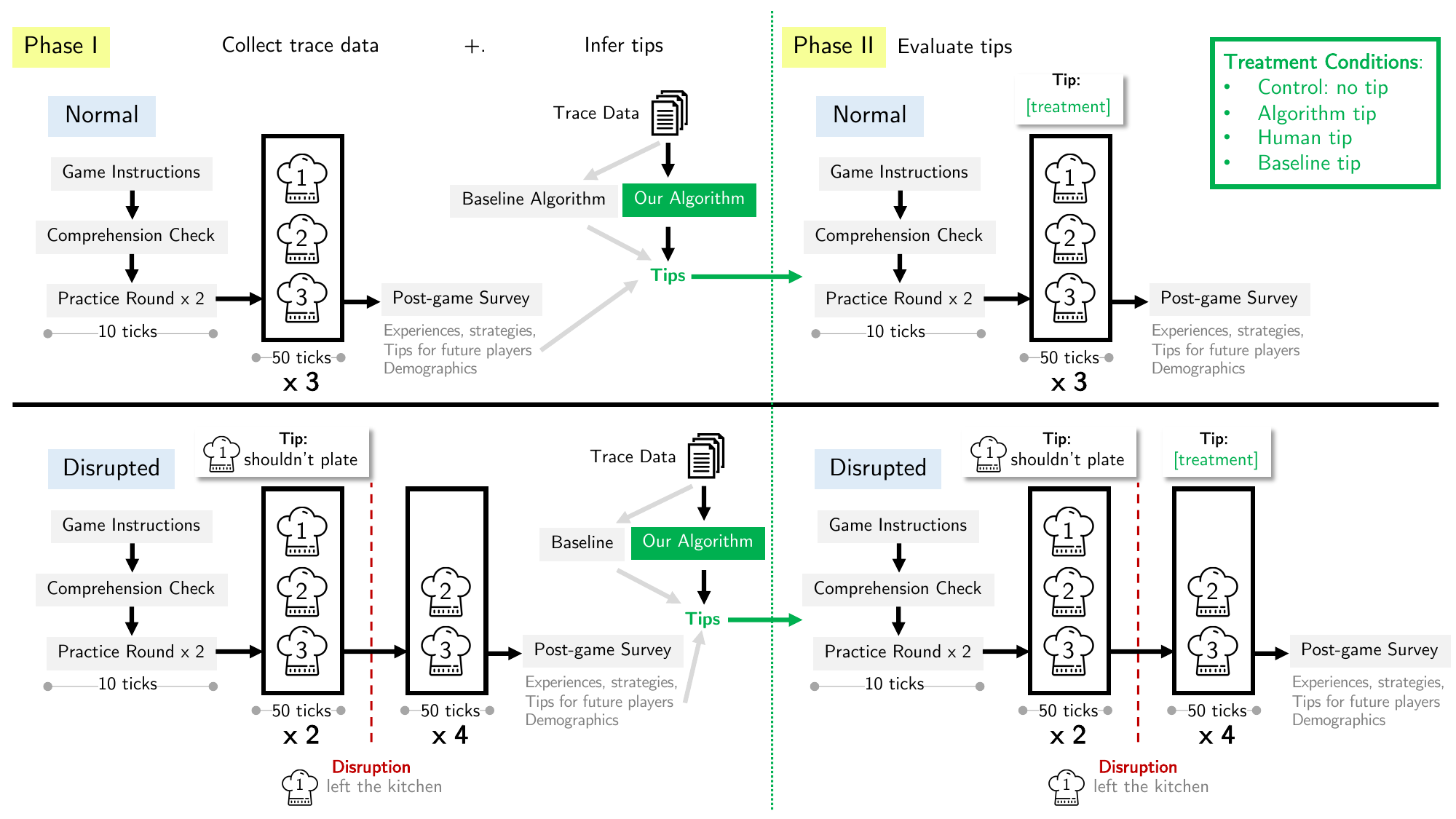}
\caption{Overview of experimental flow. The top two panels depict Phase I (left) and II (right) for the normal configuration, where each participant plays three fully-staffed scenarios. The bottom two panels depict Phase I (left) and II (right) for the disrupted configuration, where each participant plays two fully-staffed and four understaffed scenarios. Phase II participants are randomly assigned to one of four conditions (control, algorithm, human, and baseline). The set of participants across all four configuration-phase pairs is mutually exclusive.}
\label{fig:phasesflow}
\end{figure*}

\subsection{Overview}

Figure \ref{fig:phasesflow} summarizes our experiment, which proceeds in two phases. In Phase I, we recruit AMT workers to play our game without showing them any tips, and collect trace and survey data on their behavior. This phase enables us to collect historical data that would normally already be available for an existing decision-making task, which we use to infer tips.

Next, Phase II is our actual randomized controlled experiment; in this phase, we again recruit AMT workers to play our game, but this time, we randomize each participant into one of four \emph{advice conditions}, and show them a tip that depends on their advice condition (namely, the tip inferred using our algorithm, two alternative tips, and a control group where they are not shown any tip). We measure the performance of the participants, with the goal of determining whether our approach improves over the three alternatives. We describe the four advice conditions below.

In both phases, each participant plays a sequence of three or six \emph{rounds} of our virtual kitchen management game; each round is one instance of our game that is completely independent of the other rounds.  The number of rounds is determined by the \emph{game configuration} they are assigned to (normal vs. disrupted), which we described below. By having the participant play multiple rounds instead of a single one, we can study both how performance varies with the tip they are shown, as well as how it evolves across games as participants gain experience.

In summary, Phase I is purely to gather data for computing tips; in this phase, participants are randomly assigned to one of two conditions (game configuration). Then, Phase II is our main experiment, which uses a 2 (game configuration) $\times$ 4 (advice condition) between-subjects design; in this phase, participants are assigned randomly to the eight total conditions (two game configuration conditions times four advice conditions). See additional details on the experimental design (e.g., details on inferred tips, performance-based pay) in Appendix~\ref{app:exp-design}, participant demographics in Appendix~\ref{app:more-details}, and screenshots of our game in Appendix~\ref{app:screenshots}.

\paragraph{Game configurations.}

In both phases of our experiment, participants are randomized into one of two game configurations, each of which determines a sequence of rounds of our game:
\begin{itemize}
\item Normal configuration: Each participant plays three rounds of the fully-staffed scenario
\item Disrupted configuration: Each participant plays two rounds of the fully-staffed scenario, followed by four rounds of the understaffed scenario (i.e., the chef is no longer available), for a total of six rounds.
\end{itemize}
Intuitively, the normal configuration studies whether tips can help human participants fine-tune their performance. In contrast, the disrupted configuration is designed to show how tips can help participants adapt to novel situations where the optimal strategy substantially changes. The disrupted scenario is the more interesting one, since disruptions often cause workers to struggle to adapt~\citep{ramdas2017variety, bavafa2021recovering}, making tips especially useful.

\paragraph{Advice conditions.}

In Phase II, participants are randomly assigned not only to a game configuration, but also one of four advice conditions:
\begin{itemize}
\item ``Control group'' condition: Participants are not shown any tips.
\item ``Our algorithm'' condition: Participants are shown the tip inferred by our algorithm.
\item ``Human'' condition: Similar to peer feedback, participants are shown the tip most frequently suggested by Phase I participants after they have completed all rounds of our game.
\item ``Baseline algorithm'' condition: Participants are shown a tip derived by a baseline algorithm that leverages simple descriptive statistics to identify the state-action pair where human participants and the optimal policy most frequently differ.
\end{itemize}
These advice conditions, described in more detail in Section~\ref{sec:adviceconditions}, are chosen to illustrate how our algorithmic approach compares to and complements worker learning in practice.

\paragraph{Phase I details:} In Phase I, we have $N=183$ participants for the normal configuration, and $N=172$ participants for the disrupted configuration.

\paragraph{Phase II details:} In Phase II, we have $N=1,317$ participants for the normal configuration, and $N=1,011$ participants for the disrupted configuration. In the normal configuration, Phase II participants are shown the tip for their advice condition for the fully-staffed scenario on all rounds. In the disrupted configuration, they are shown the tip designated by their condition for the understaffed scenario (the last 4 rounds). In the first two rounds of the disrupted configuration, our goal is to quickly acclimate participants to the fully-staffed scenario in a way that is consistent across conditions. Thus, we show our algorithm tip for the fully-staffed scenario---``Chef should never plate''---across all conditions (including control) for the first two rounds; we choose this tip because, as we show in Section~\ref{sec:results}, it most quickly improves human performance in the fully-staffed scenario. After the disruption, we inform participants that the optimal strategy has now changed due to the chef's departure. 

\paragraph{Participant recruitment and pay.}

We recruited participants on the Amazon Mechanical Turk (AMT) platform. Each participant can only participate once across both phases and all conditions---i.e., no participant has prior experience with any version of the game. Participants are compensated a flat rate for completing the study, plus a relatively large performance-based bonus determined by how quickly they complete each round of the game (see Appendix~\ref{app:payschemes} for details).

\paragraph{Hypotheses.}

Our main outcomes of interest are the average performance in the final round of the game (i.e., the average number of time steps taken by participants to complete all orders in the final round they play), as well as the fraction of participants who ultimately learn the optimal policy. The final round is the the fourth round of the normal configuration and the sixth round of the disrupted configuration. Then, our main hypothesis is that for each of the two game configurations, participants in the ``our algorithm'' advice condition (i.e., shown the tip inferred using our algorithm) outperform participants in the other three advice conditions. In addition to our main hypothesis, we also examine participant behaviors in response to different tips, particularly their compliance, and how they learn to improve their decision-making beyond the provided tips.

\subsection{Advice Conditions}
\label{sec:adviceconditions}

\paragraph{Control group.}

The ``control group'' condition represents settings where best practices are not readily available, so workers must learn over time based on their own experience; indeed, we observe that performance improves over time without any tips in this condition.

\paragraph{Our algorithm.}

The ``our algorithm'' condition represents our approach. In particular, we use the tip $\hat{\rho}$ inferred using our algorithm (Eq.~\eqref{eqn:algo}) based on the trace data obtained in Phase I. Additional details are provided in Appendix~\ref{supp-algo}.

\paragraph{Human.}

The ``human'' condition represents settings where one can obtain advice on best practices from more experienced peers \citep[e.g., as in][]{song2017closing}. We use Phase I to do so. In particular, each participant in Phase I is shown a comprehensive list of candidate tips at after completing all rounds of our game, and is asked to select the tip they believe would most improve the performance of future players. This list is constructed by merging three types of tips:
\begin{enumerate}
\item all possible tips of the format described in Appendix~\ref{app:rulespace} (e.g., ``Chef should not plate''),
\item a small number of generic player tips that arose frequently in our exploratory pilot studies (e.g., ``Keep everyone busy at all times''), and
\item a small number of manually constructed tips obtained by studying the optimal policy (e.g., ``Chef should chop as long as there is no cooking task").
\end{enumerate}
Our algorithm's tip is always contained in this list, as part of the first category above. This list contained 13--14 tips (depending on the configuration), which we found to be a reasonable length that did not overwhelm participants in our pilot studies. We take the most frequently chosen tip as the ``human tip'', capturing the wisdom of the (experienced) crowd. We also considered several variations, such as taking the tip recommended by the best performing human participants, but these variations all resulted in the same tip; see Appendix~\ref{app:varyhumantip} for details.

The human tip is designed to demonstrate how our algorithmic approach can exceed the capabilities of humans to offer useful advice, capturing the limitations of relying on peers for advice.

\paragraph{Baseline algorithm.}

The ``baseline algorithm'' condition illustrates a na\"{i}ve use of descriptive statistics on historical trace data to provide tips---simply looking for frequent differences between the human and optimal policies, rather than leveraging interpretable reinforcement learning to identify the most consequential actions for improving performance. In particular, given rollouts $\zeta_1^*,...,\zeta_h^*\sim D^{(\pi^*)}$ sampled using the optimal policy, we let $C^*(s,a)$ denote the number of times state-action pair $(s,a)$ occurs across these rollouts. Then, given the observed rollouts (i.e., historical trace data from human decision-making) $\zeta_1,...,\zeta_k\sim D^{(\pi_H)}$, the baseline algorithm selects the tip
\begin{align}
\label{eqn:baseline}
\hat{\rho}_{\text{bl}}=\operatorname*{\arg\max}_\rho\frac{1}{k}\sum_{i=1}^k\sum_{t=1}^TC^*(s_{i,t},a_{i,t}).
\end{align}
In other words, our baseline optimizes the same objective but with $Q^*$ replaced with $C^*$. Intuitively, this baseline strategy tries to directly imitate the optimal policy, whereas our strategy prioritizes state-action pairs that are more relevant to achieving high rewards. In this condition, we show participants the tip $\hat{\rho}_{\text{bl}}$ inferred by the baseline algorithm (Eq.~\eqref{eqn:baseline}) based on the Phase I data.

This baseline algorithm ignores the sequential nature of our decision-making problem. It is designed to highlight the complexity of sequential structure compared to the one-shot decision-making setting studied in prior work, and in particular, the importance of accounting for this sequential structure when inferring tips.

\section{Experimental Results} \label{sec:results}

\begin{figure}
\centering
\subfloat[Tips for each condition and configuration]{
\includegraphics[width=0.97\textwidth]{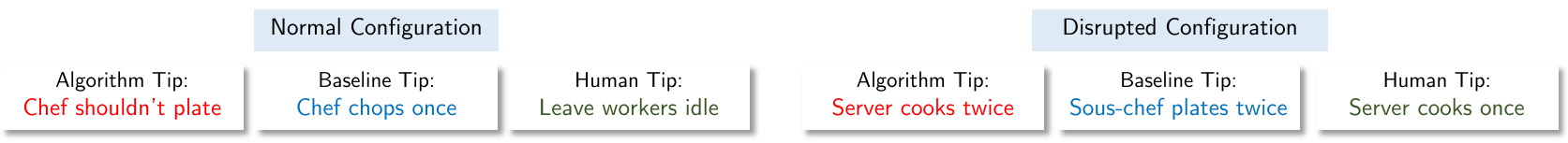}\label{toptips}}\\
\subfloat[Final Round Performance (Normal)]{
\includegraphics[width=0.47\textwidth]{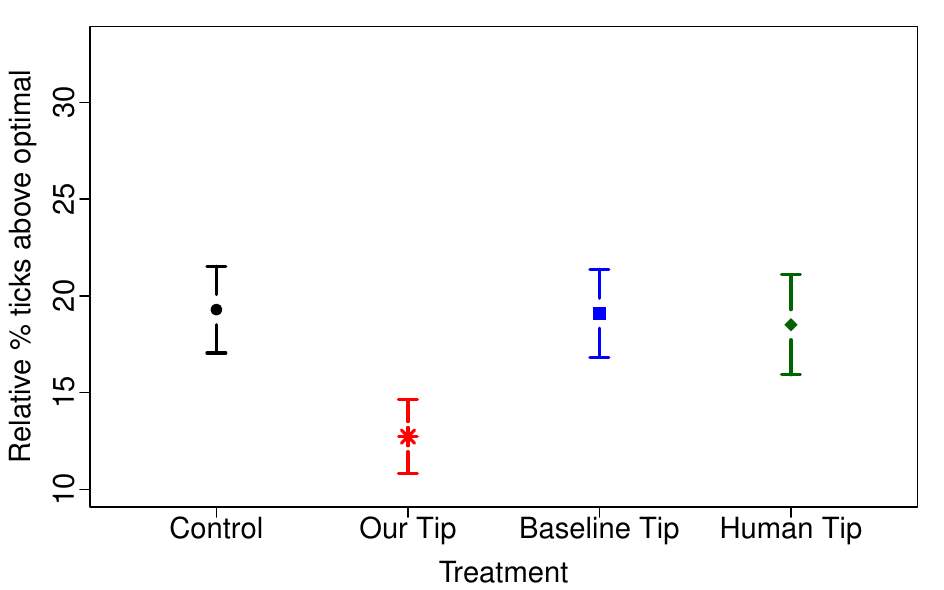}\label{phase2nfint}}\quad
\subfloat[Final Round Performance (Disrupted)]{
\includegraphics[width=0.47\textwidth]{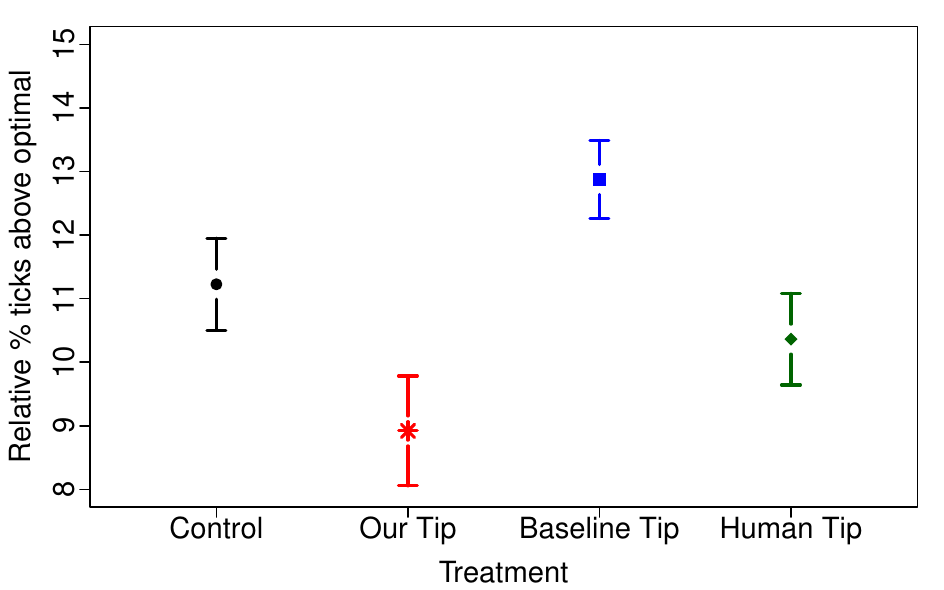}\label{phase2dfint}}\\
\subfloat[Performance over Time (Normal)]{
\includegraphics[width=0.46\textwidth]{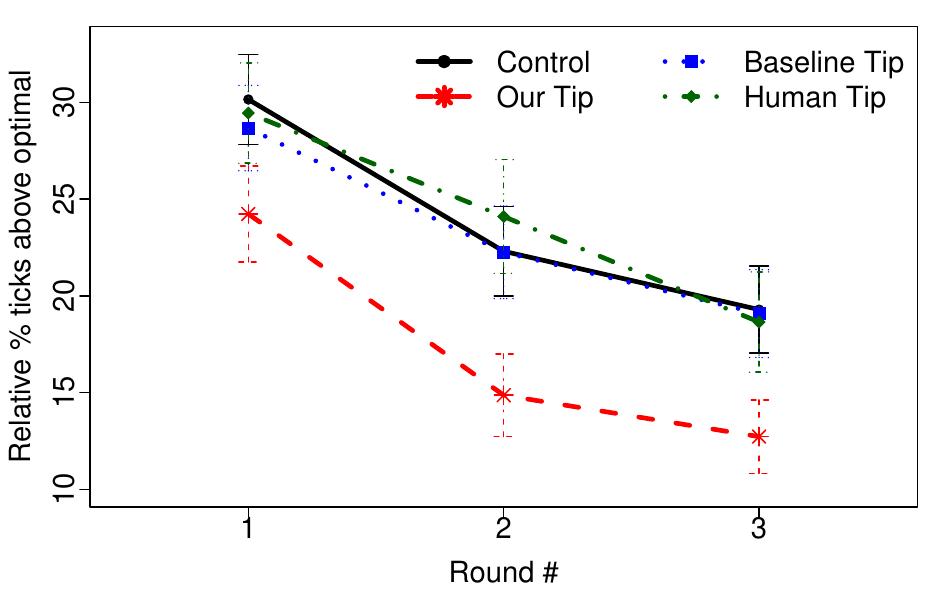}\label{phase2nticks}}\quad
\subfloat[Performance over Time (Disrupted)]{
\includegraphics[width=0.47\textwidth]{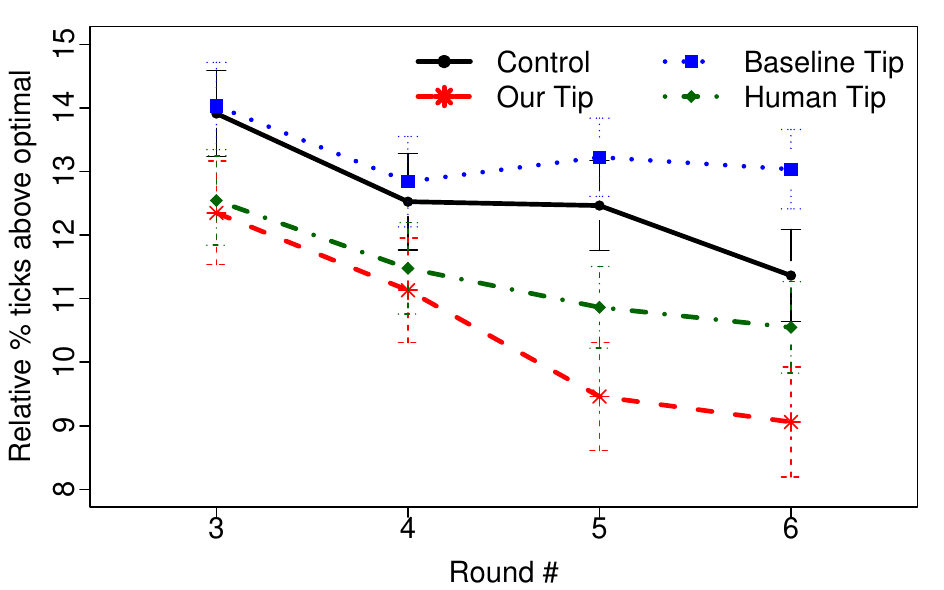}\label{phase2dticks}} \\
\subfloat[Fraction Achieving Optimal (Normal)]{
\includegraphics[width=0.46\textwidth]{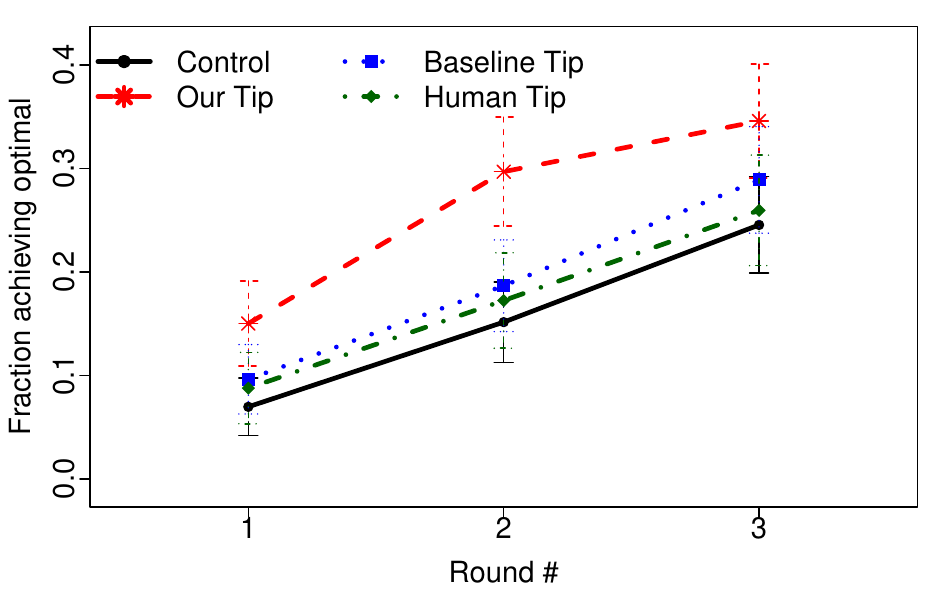}\label{phase2nopt}}\quad
\subfloat[Fraction Achieving Optimal (Disrupted)]{
\includegraphics[width=0.47\textwidth]{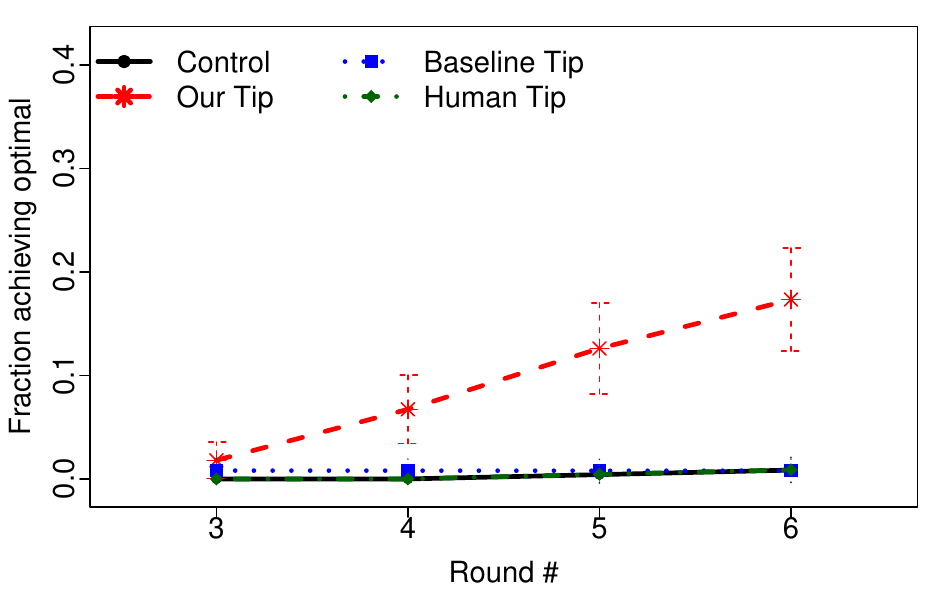}\label{phase2dopt}}
\caption{Phase II Participant Performance. \footnotesize The top row shows the tips derived for each condition and configuration based on Phase I data. Remaining rows depict various views of participant performance across conditions in the normal (left) and disrupted (right) configurations. The top row shows performance in the last round of the configuration, the second row shows how participant performance improves over time, and the third row shows the fraction of participants who execute an optimal policy over time.}
\label{fig:performance}
\end{figure}

Despite their simplicity and conciseness, we find that our tips can significantly improve participant performance since they capture strategies that are hard for participants to learn; in contrast, alternative tips have varying empirical shortcomings that reduce their effectiveness. We also describe how participant compliance---a key ingredient to ultimately improving decisions---varies across tips. Finally, we find evidence that participants do not blindly follow our tips, but combine them with their own experience to discover additional strategies beyond our tips. Figure~\ref{toptips} shows the tips inferred in each condition for each configuration using trace and survey data from Phase I.

\subsection{Performance: \emph{Our Tips Substantially Improve Performance}}\label{ssec:perf}

Figure \ref{fig:performance} shows performance results across all four conditions and both configurations. Figure \ref{phase2nfint} \& \ref{phase2dfint} show participant performance in the final round of our game,
Figure \ref{phase2nticks} \& \ref{phase2dticks} show how performance improves across rounds, and
Figure \ref{phase2nopt} \& \ref{phase2dopt} show the fraction of participants achieving optimal performance across rounds.
For each configuration, we report performance as the excess ticks (time) taken over the optimal policy, normalized by the optimal policy's ticks, i.e.,
\[ \frac{\text{\# ticks taken}-\text{optimal \# ticks}}{\text{optimal \# ticks}} \,.\]
Results in terms of the raw number of ticks are shown in  Figure~\ref{fig:totalticks} in Appendix~\ref{app:totalticks}.

The normal configuration is relatively easy for participants---a substantial fraction (24\%) discover the optimal policy by the final round without the aid of tips (control group). As shown in Figure \ref{phase2nfint}, participants shown our tip completed the final round in 22.5 steps on average, significantly outperforming participants in the control group ($t(329) = -4.397,~p < 10^{-4}$), those shown the human-suggested tip ($t(312) = -3.628,~p = 2\times10^{-4}$), and those shown the tip from the baseline algorithm ($t(334) = -4.232,~p < 10^{-4}$).\footnote{Results remain highly statistically significant under a Bonferroni correction for multiple hypothesis testing.} Our tip speeds up learning by at least one round compared to the other conditions---i.e., the performance of participants given our tip on round $k$ was similar to or better than the performance of participants in other conditions on round $k+1$ (Figure~\ref{phase2nticks}). Our tip also helped more participants (35\%) achieve optimal performance (20 steps) in the final round, compared to 24-29\% in other conditions.

The disrupted configuration is substantially harder, since participants must adapt to the more counter-intuitive understaffed scenario. 
Perhaps as a consequence, participants benefit much more from tips: those in the control group took four rounds to achieve the same level of performance as those shown our tip on the first round. Participants shown our tip completed the final round in 37.1 steps, again significantly outperforming participants in the control group ($t(243) = -4.361,~p < 10^{-4}$), those shown the human-suggested tip ($t(246) = -2.52,~p = 6\times10^{-3}$), and those shown the tip from the baseline algorithm ($t(246) = -7.348,~p < 10^{-4}$). In the disrupted configuration, the baseline tip actually \textit{reduces} participant performance, likely because participants struggle to operationalize it. More starkly, 19\% of participants shown our tip achieved optimal performance (34 steps) in the final round, compared to less than 1\% in all other conditions---i.e., our tip uniquely helps participants learn to play optimally. Note that there were no significant differences in performance across conditions when playing the two fully-staffed rounds in the disrupted configuration. Therefore, the relatively worse performance under other conditions reflect the informativeness of alternative tips.

\paragraph{Shortcomings of baseline tips.} As noted earlier, this tip tries to mimic the optimal policy rather than focusing on consequential actions; thus, we expect these tips to be less valuable to participants (for improving performance) than our algorithm's tips. Participants complied with both the baseline algorithm's tips and our algorithm's tips at similar rates (see Section~\ref{sec:compliance}). 

However, the baseline algorithm's tip is still derived from the optimal policy, so it is surprising that it performs \textit{worse} than the control condition in the disrupted configuration. In fact, in Section~\ref{sec:beyondtips}, we show that participants who received our algorithm's tips also learned the strategy encoded in the baseline algorithm's tip; however, participants who received the baseline algorithm's tip did not learn the strategy encoded in our algorithm's tip in both configurations. Thus, the problem is not with the \textit{content} of the baseline algorithm's tip, but rather that participants struggle to \textit{operationalize} the baseline tip into their workflow (without knowing our algorithm's tip).

In particular, when participants apply a tip, they shift to new unseen portions of the state space, and must also learn to act well in those states. By focusing on ``high-value'' states and critical performance bottlenecks, our algorithm more easily enables participants' off-distribution learning. 
For example, in the disrupted configuration, the baseline algorithm's tip ``Sous-chef should plate twice'' suggests actions that occur late in the game (hindering participants' ability to explore and alter their strategy) and does not focus on the critical performance bottleneck (cooking). In contrast, our algorithm's tip ``Server should cook twice" frees the sous-chef to plate later in the game (a strategy---not explicitly conveyed in our tip---that participants automatically learn when given our tip). However, targeting early decisions alone is not sufficient to help participants learn. In the normal configuration, although the baseline algorithm's tip targets an \textit{earlier} action (``Chef should chop once'') compared to our algorithm’s tip (``Chef shouldn’t plate''), it fails to help participants learn the entire optimal strategy (see Section~\ref{sec:beyondtips}) because it does not address the important bottleneck (keeping the chef from the lengthy task of plating).

\paragraph{Shortcoming of human tips.} While the human-suggested tips consistently improve performance compared to the control group, they can be overly general or incorrect. In the normal configuration, Phase I participants did not translate their strategy into a specific tip---i.e., their suggested tip ``Strategically leave some workers idle" captures a strategy needed to perform better but fails to convey any necessary details to identify the optimal strategy. Alternatively, in the disrupted configuration, Phase I participants provided an \textit{incorrect} tip, suggesting ``Server should cook once", whereas the optimal policy actually assigns the server to cook twice (as suggested by our tip)---i.e., participants identified the correct direction of change in response to the under-staffing disruption, but at an insufficient magnitude. The tips chosen by participants are remarkably consistent across different participant subgroups---e.g., top performers from Phase I vs. all participants (see Appendix C.5)---and generally fail to capture counter-intuitive properties of the optimal policy. Perhaps due to their more intuitive nature, participants are substantially \textit{more} likely to comply with the human tip than with our algorithm's tip (see Section~\ref{sec:compliance}). Thus, our results suggest that the worse performance of the human tip is due to the sub-optimal quality of the chosen tip.\footnote{Note that human participants have a slightly different tip search space than our algorithm. However, this discrepancy cannot be the source of the performance difference, since in the disrupted configuration, both our algorithm's tip and the human tip are present in both search spaces; participants then chose an incorrect tip.}

\subsection{Compliance: \emph{Humans Comply with Tips More over Time}}\label{sec:compliance}

As discussed earlier, the effectiveness of a tip critically depends on whether humans are able to understand it and implement it effectively. This involves both complying with the tip's suggested actions as well as modifying other portions of their strategy to make full use of the tip. First, we examine compliance with the tips. Note that participants were not informed of the source of the tip (i.e., algorithm or human), so any variation in compliance is due to the content of the tip, rather than behavioral reactions to its source~\citep[e.g., algorithmic aversion, see][]{dietvorst2015algorithm}.

\begin{figure}
\centering
\subfloat[Normal Configuration]{
\includegraphics[width=0.47\columnwidth]{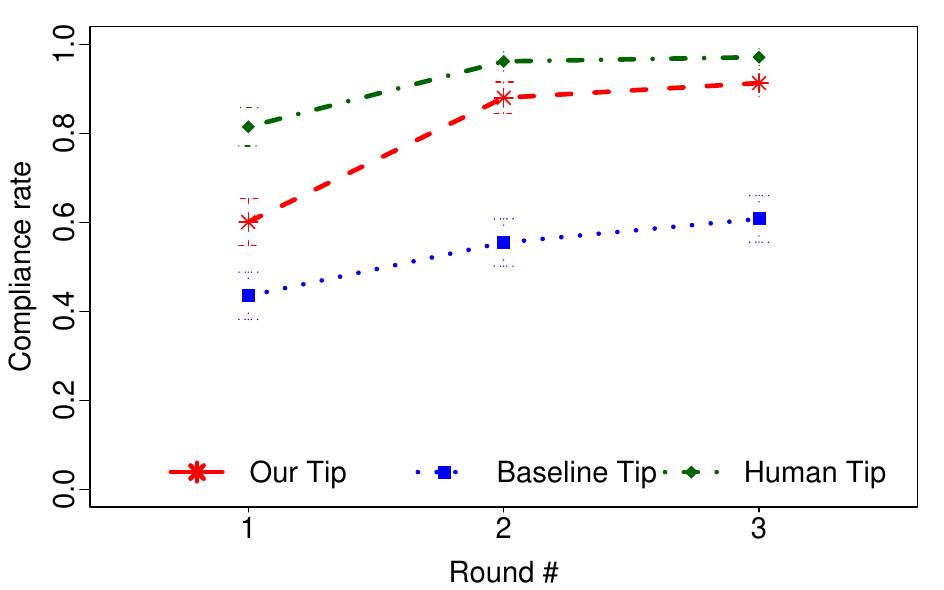}\label{nc}}\quad
\subfloat[Disrupted Configuration]{
\includegraphics[width=0.47\columnwidth]{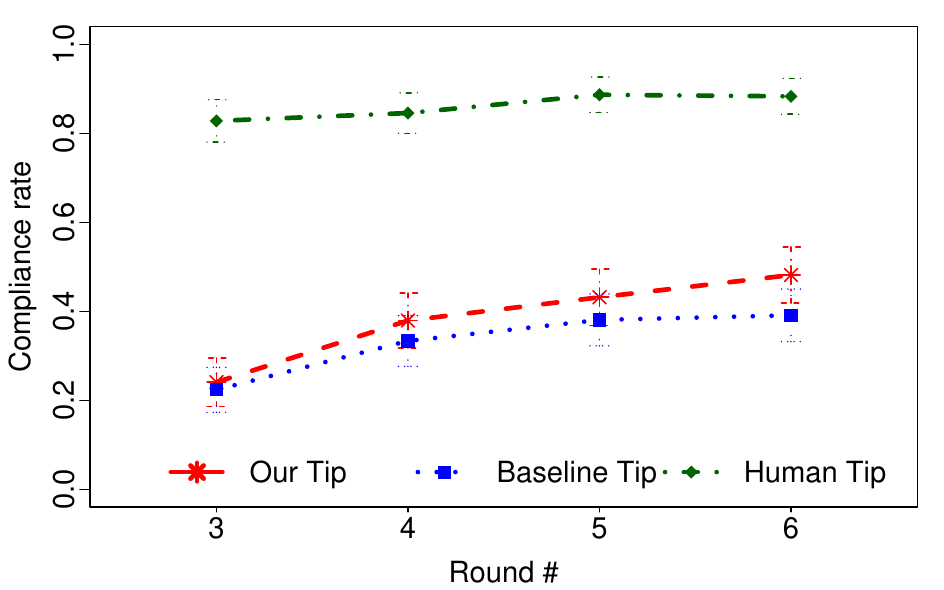}\label{dc}}
\caption{Compliance with Tips. Participant compliance in Phase II with the respective tip they were shown in each condition for the normal (left) and disrupted (right) configurations over time.} \label{compliance}
\end{figure}

Figure \ref{compliance} shows the fraction of participants that complied with the tip they were offered in each condition. Specifically, we measure the fraction of participants that act in a way that is consistent with the tip they are shown.\footnote{For the human tip in the normal configuration (``Strategically leave some worker(s) idle''), we measured compliance by identifying if the participant ever skipped a potential task assignment when at least one virtual kitchen worker was idle and there was at least one available subtask. Note that we cannot perfectly certain if such “skipping” was strategic, but given that participants were financially incentivized to complete each round as fast as possible, we expect that participants would
not skip an assignment unless they were being strategic.} We see that participants increasingly comply with the tips shown over time---as they gain experience, and better understand the significance of the tip---in all conditions. Compliance with the baseline algorithm's tip was relatively low in both configurations, suggesting that participants did not find it as useful. Alternatively, compliance with the human-suggested tip was higher than compliance with our algorithm's tip, particularly in the disrupted configuration. Based on participants' post-game feedback, we found that this is likely because the human-suggested tip better matches human intuition (since it is devised by humans).
The disrupted configuration is illustrative. Although our algorithm's tip is correct (unlike the human-suggested tip), it is highly counter-intuitive, hurting adoption. For example, in the disrupted scenario, our tip ``Server should cook twice'' may appear unreasonable since the server is very slow at cooking; in fact, participants \textit{just} learned to never assign the server cooking in the fully-staffed scenario prior to the disruption. Yet, having the server cook twice is the only way to achieve optimal performance in the understaffed scenario; in contrast, the human-suggested tip is to only have the server cook once, which is a less sharp departure from the previously employed policy. As participants gain experience with the new understaffed scenario, they grow to appreciate the value of our algorithm's tip (i.e., compliance with our algorithm's tip more than doubles over the four rounds). 

Our results suggest that participants do not blindly follow tips; instead, they only follow them if they believe that the suggested strategy is effective. Qualitative evidence from post-game survey responses suggested that participants ignored tips they did not agree with (see Section~\ref{sec:humancomments}). Thus, compliance is a function not just of the interpretability of the tip (which is unchanged across conditions), but also the strategy it encodes. When the optimal strategy is counter-intuitive, we observe an intrinsic trade-off between the optimality of the tip and compliance with the tip. Even in the disrupted configuration, our algorithm's tip succeeds despite much lower compliance (relative to the human-suggested tip) since it suggests a highly effective strategy; as seen in Figure~\ref{phase2dopt}, participants that understand this strategy can achieve optimal performance (whereas essentially none of the participants in the other conditions were able to do so). Interestingly, as we show in the next subsection, participants in the \textit{control group} also comply with the human-suggested tip at a high rate---i.e., the human-suggested tip largely captures behaviors that are likely to be adopted even in the absence of tips; in contrast, our algorithm's tip allows participants to learn new strategies that they may not learn otherwise. 

One may be able to improve compliance with algorithmic tips through additional levers. For instance, financial incentives~\citep{giuffrida1997should} and providing social nudges by setting norms~\citep{benjamin2010social} can improve human compliance with recommendations or guidelines; however, it is important to design these approaches in a context-specific way to avoid null or even negative effects~\citep[see, e.g.,][]{beshears2015effect, chang2021financial}. We perform a follow-up experiment testing these levers to improve compliance with our algorithm's tip (see Section~\ref{sec:algoaversion}).

\subsection{Learning Beyond Tips: \emph{Our Tips Help Humans Learn to Perform Optimally}}\label{sec:beyondtips}

One of the critical challenges in sequential decision-making is that the human must learn strategies \textit{beyond} the provided tip to achieve good performance throughout their workflow, since the tip only captures a portion of the optimal policy. To study whether humans learn the optimal policy, we examine what kinds of strategies they learn beyond the specific tips they were shown. More precisely, we study \emph{cross-compliance}, which is the compliance of the participant to tips other than the one they were shown. Na\"{i}vely, there is no reason to expect participants to cross-comply with a tip that we did not show them beyond the cross-compliance exhibited by the control group. Thus, any cross-compliance beyond that of the control group measures how a tip enables participants to learn strategies outside of the stated tip. Assuming these strategies are consistent with the optimal policy, cross-compliance serves as a way to measure participants' progress towards operationalizing the tip effectively throughout their broader workflow.

\begin{figure}
\centering
\subfloat[Our Tip:\\\emph{``Server Cooks Twice"}]{
\includegraphics[width=0.24\columnwidth]{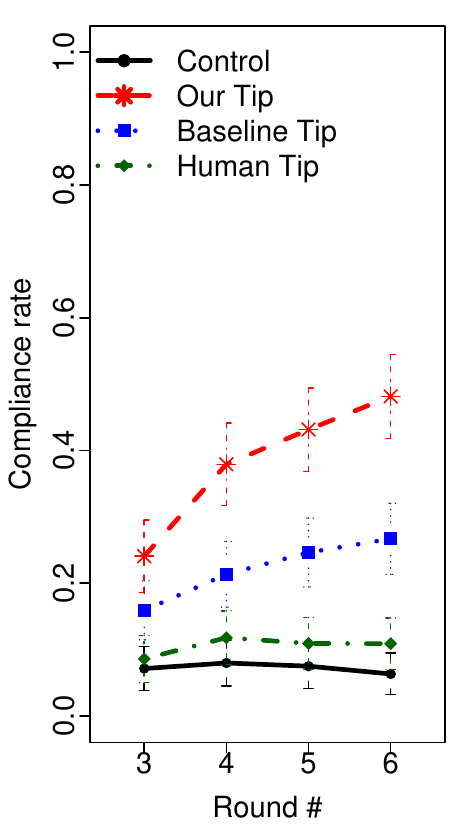}\label{beyond-alg}}
\subfloat[Human Tip:\\\emph{``Server Cooks Once"}]{
\includegraphics[width=0.24\columnwidth]{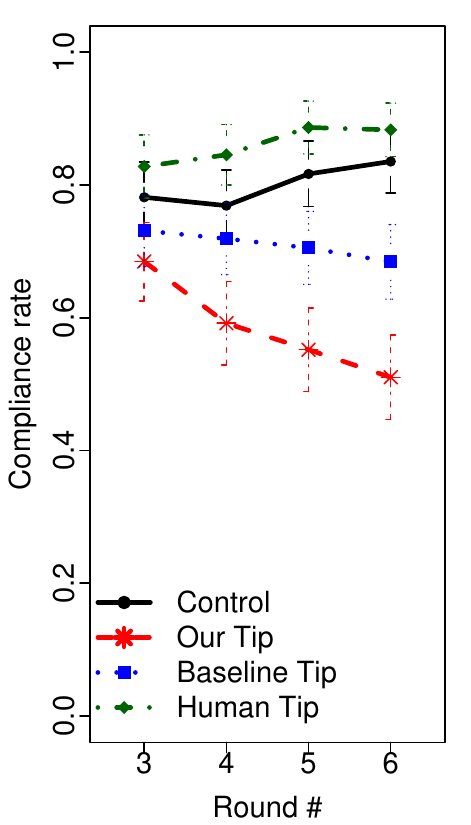}\label{beyond-human}}
\subfloat[Baseline Tip:\\\emph{``Sous-chef Plates Twice"}]{
\includegraphics[width=0.24\columnwidth]{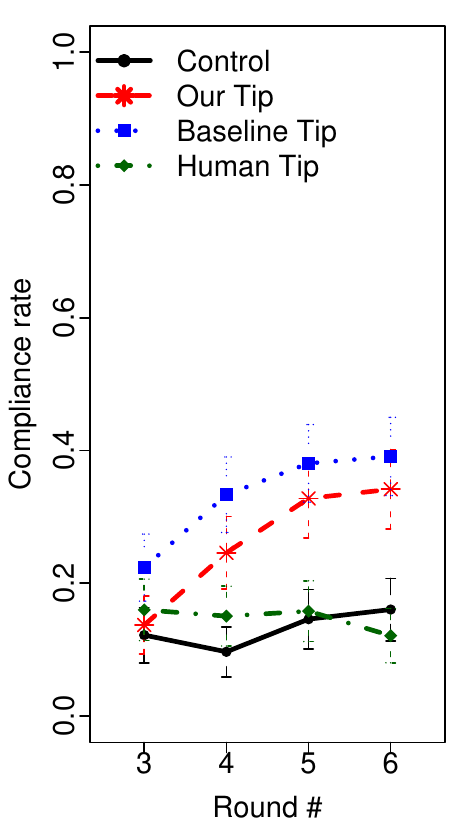}\label{beyond-baseline}}
\subfloat[Unshown Tip:\\\emph{``Server Chops Once"}]{
\includegraphics[width=0.24\columnwidth]{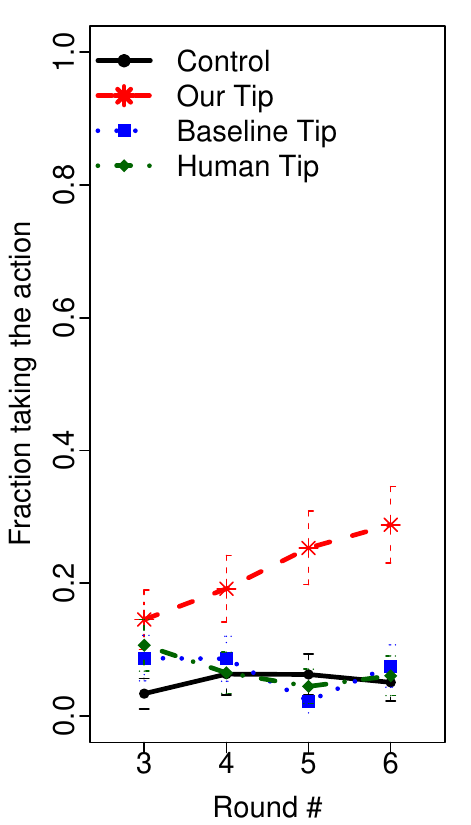}\label{beyond-unshown}}
\caption{Learning beyond Tips. Panels (a)-(c) show the rate at which participants in each condition cross-comply with each offered tip over time in the disrupted configuration. Panel (d) shows analogous results for a rule that is part of the optimal policy but was not shown as a tip in any condition.}
\label{fig:beyondtips}
\end{figure}

We focus on the disrupted configuration since it is more challenging for participants, leading to more interesting cross-compliance patterns.\footnote{In the easier normal configuration, participants in all conditions cross-comply with all other tips (which are all part of the optimal policy), but they achieve higher cross-compliance when shown our algorithm's tip; see Appendix~\ref{app:crossnormal}.} Figure~\ref{fig:beyondtips} shows the cross-compliance of participants in each condition with the different tips (algorithm, baseline, human), as well as a new tip (``Server chops once'') not shown to any participants. This new tip is part of the optimal policy for the understaffed scenario used in the disrupted configuration.
Participants in the human and control groups only comply with the human tip. Indeed, the human-suggested tip actually contradicts the optimal policy; thus, despite effectively operationalizing the tip, participants are prevented from learning the other tips which are part of the optimal policy.\footnote{Note that participants in the control and human conditions comply with the human tip at similar rates, i.e., the human tip suggests behaviors that are highly likely to be adopted even in the absence of tips.}
Participants shown the baseline tip only have high compliance with the baseline tip, indicating that the baseline tip could not help participants uncover the rest of the optimal policy; although the baseline tip is part of the optimal policy, it fails to help participants discover strategies beyond the tip itself, since it does not focus on high-value states and critical bottlenecks (see our earlier discussion in Section~\ref{ssec:perf}). In contrast, participants who received our algorithm's tip have high cross-compliance with \textit{all} parts of the optimal policy (i.e., the baseline tip and the unshown tip); furthermore, our algorithm is the only condition where cross-compliance with the sub-optimal human tip decreases over time. That is, our tip uniquely enables participants to combine the tip with their own experience to discover useful strategies (that are consistent with the optimal policy) beyond what is stated in the tip.

\subsection{Compliance: \emph{Algorithm Aversion and Interventions to Improve Compliance}}\label{sec:algoaversion}

Compliance in sequential decision-making depends on many factors. Thus far, we have found evidence that intuitive tips can improve compliance, while difficulty operationalizing tips into one's broader workflow can hurt compliance. In practice, additional levers such as trust and incentives may also play a significant role in compliance. We conduct two follow-up studies to examine how these levers may affect our results in real-world scenarios.

\paragraph{Trust in Algorithms.} While our experiments thus far did not reveal the \textit{source} of the tip (i.e., whether it is generated by an algorithm), workers may be able to infer this information in real-world contexts, potentially resulting in algorithm aversion~\citep{dietvorst2018overcoming}---i.e., where humans are mistrustful of algorithmic advice. To this end, we perform a pilot study in the disrupted configuration to evaluate the impact of algorithm aversion on compliance. We randomly assign participants into one of two conditions: ``Our Algorithm: No Source'' and ``Our Algorithm: With Source''. The ``Our Algorithm: No Source'' condition is identical to the ``Our Algorithm'' condition in our main study, i.e., the participant is shown ``Tip: Server should cook twice'' during the under-staffed rounds. In contrast, in the ``Algorithm: With Source'' condition, the participant is instead shown: ``Tip from AI Algorithm: Server should cook twice. The AI algorithm analyzes past players' strategies and chooses the best tip that would help improve your performance.''

\begin{figure}[htpb]
\centering
\subfloat[Compliance Rate with Our Algorithm's Tip]{
\includegraphics[width=0.47\columnwidth]{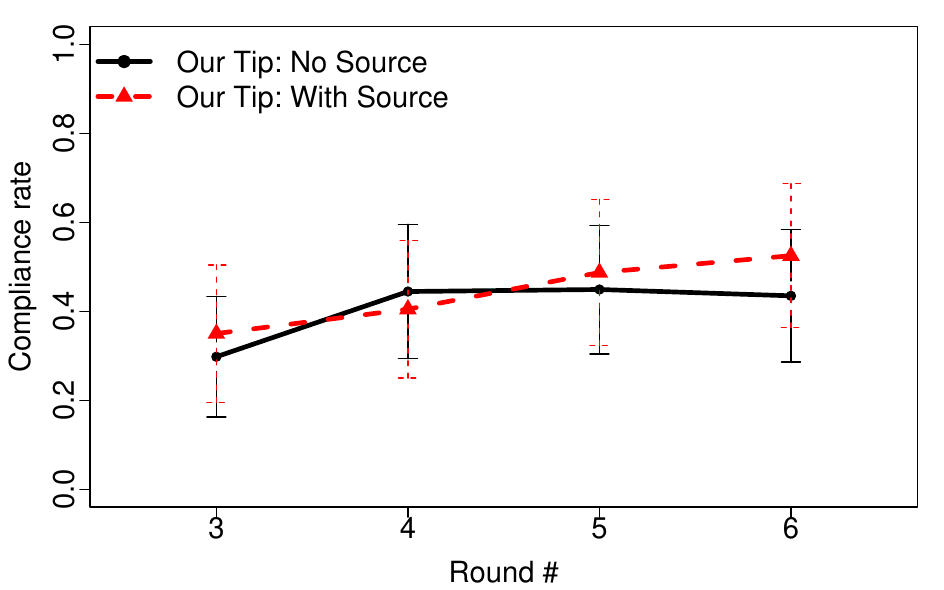}\label{2a-compliance}}\quad
\subfloat[Performance over Time]{
\includegraphics[width=0.47\columnwidth]{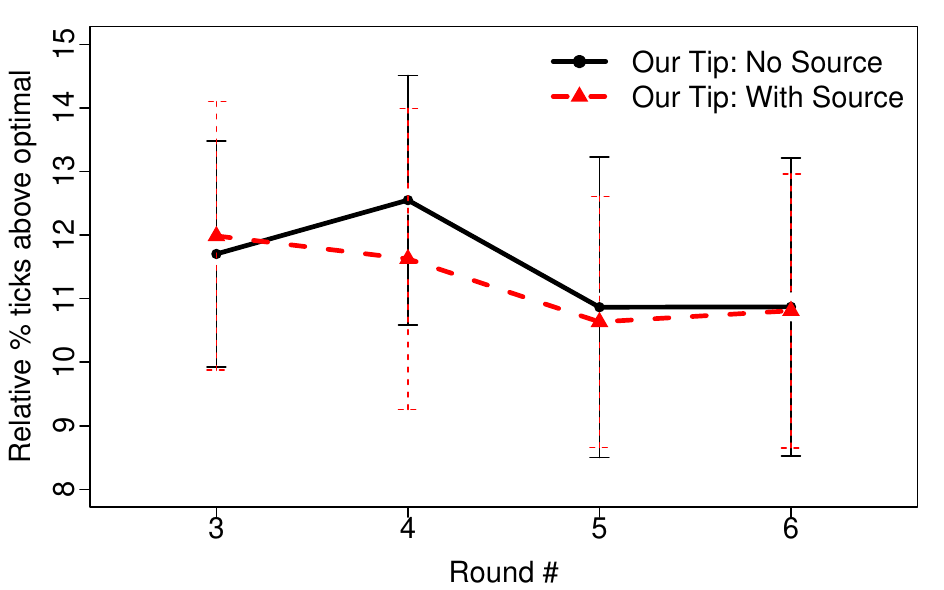}\label{2a-tick}}
\caption{Source of Tip. Participant compliance with our algorithm's tip (``Server should cook twice'') (left) and participant performance (right) whether the information about the source of the tip was provided.}
\label{2a}
\end{figure}

We recruited 200 participants via AMT, of which 90 successfully completed the study and passed all comprehension and attention checks. We find that there are no statistically or economically significant differences in the results between the two conditions according to compliance rate (Figure~\ref{2a-compliance}) or final round performance (Figure~\ref{2a-tick}). Providing the source of the tip in fact has a directionally \textit{positive} impact on compliance, suggesting that it is unlikely that we would observe algorithm aversion. One potential explanation is that, given the complex nature of the task, knowing that the tip came from an algorithm could increase humans’ likelihood in adopting the tip similar to the phenomenon of algorithm appreciation documented by \cite{logg2019algorithm}. Note that the final round performance is essentially identical whether or not the source was provided.

\begin{figure}[htpb]
\centering
\subfloat[Compliance Rate with Our Algorithm's Tip]{
\includegraphics[width=0.47\columnwidth]{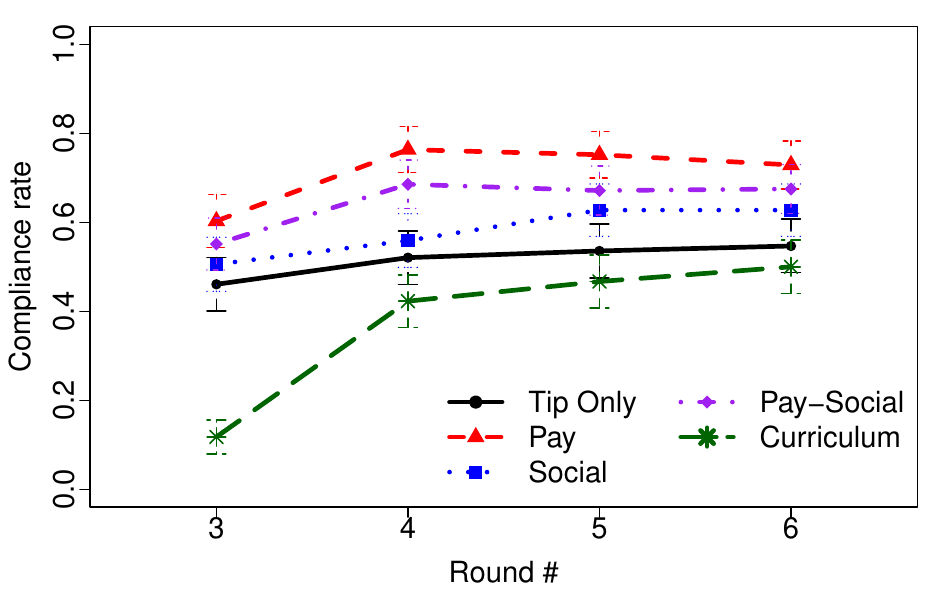}\label{2b-compliance}}\quad
\subfloat[Performance over Time]{
\includegraphics[width=0.47\columnwidth]{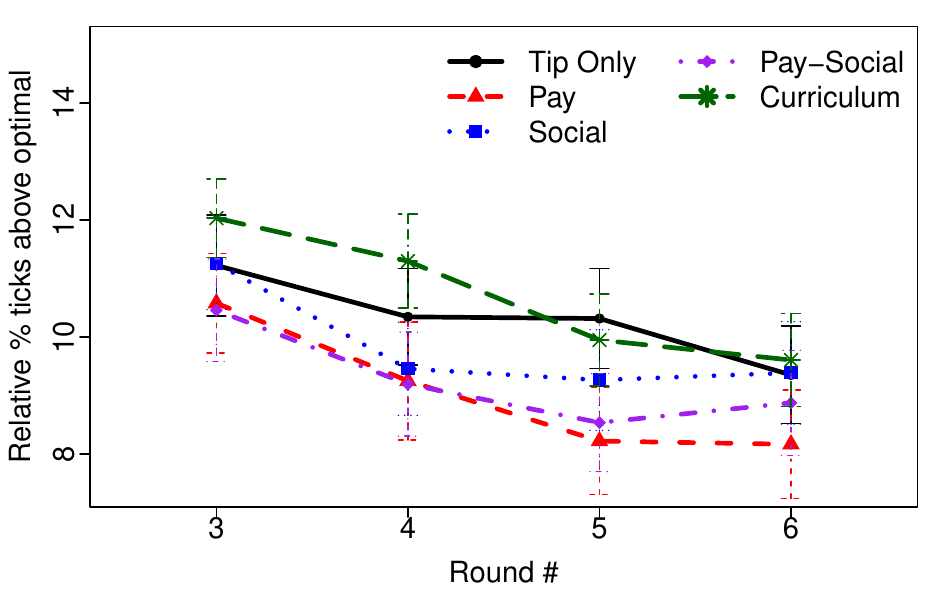}\label{2b-tick}}
\caption{Interventions for Compliance. Participant compliance with our algorithm's tip (``Server should cook twice") (left) and participant performance (right) in each intervention across the four disrupted rounds.}
\label{2b}
\end{figure}

\paragraph{Financial/Social Incentives.} Next, we conducted a follow-up full-scale experiment to determine whether we can improve compliance with our algorithm's tip through financial or social interventions.\footnote{This follow-up study is pre-registered at {\footnotesize \url{https://aspredicted.org/blind.php?x=85D_1RB}}} Focusing again on the disrupted configuration, we investigated the following four interventions in the under-staffed rounds of the disrupted configuration:
\begin{enumerate}
\item ``Pay'' condition: For rounds 3 and 4 (first two under-staffed rounds), we pay the participant the maximum pay for each round \textit{if} they successfully complied with the tip (i.e., server cooked twice). The pay scheme returns to the original performance-based one in rounds 5 and 6.
\item ``Social'' condition: We add the following to the tip for all four under-staffed rounds---``While this tip may appear counter-intuitive, the majority of best players adopted this rule, enabling them to achieve the optimal performance of 34 ticks.'' Past research has shown that information about social norms can change human behavior, e.g., residents consumed less energy after learning that their neighbors had better energy consumption ratings~\citep{allcott2011social}.
\item ``Pay-Social'' condition: Participants receive both the Pay and Social interventions.
\item ``Curriculum'' condition: In round 3 (e.g., the first disrupted round), we present the Human tip from the original study (``Server should cook once'') instead of our algorithm's tip. Then, in rounds 4 through 6, we present our algorithm's tip (``Server should cook twice''). The rationale of this intervention is to slowly move the participant's strategy from not letting the server cook any burgers to something in between (letting the server cook once) before telling them the more counter-intuitive algorithm tip (letting the server cook twice).
\end{enumerate}
We recruited 1,967 participants via AMT, of which 1,496 successfully completed the study and passed all the comprehension and attention checks. Participants were randomly assigned into one of five conditions: ``Tip Only'' (identical to the Algorithm arm in the original study), ``Pay'', ``Social'', ``Pay-Social'', and ``Curriculum'' interventions. The main outcome of interest here is the compliance rate in the final round of the game. Figure~\ref{2b-compliance} shows compliance rates by condition across all four rounds. We find that any combination of ``Pay'' and ``Social'' interventions improves compliance with our algorithm's tip. Paying people to follow the tip for two rounds appears to be effective at getting them to try out the seemingly counter-intuitive tip and such compliance sticks around when we no longer compensate merely for their compliance. Both ``Pay'' and ``Pay-Social'' interventions significantly improve compliance in round 6 compared to the ``Tip Only'' condition. Providing social information alone does improve compliance, but the effect is not statistically significant. Finally, slowly easing people toward our algorithm's tip in the ``Curriculum'' intervention does not seem to improve compliance by the end of the game; providing an intermediate step between human intuition and the optimal action might even backfire as it may slow down the ability of humans to adapt to the new environment. We note that the performances in the final round of the game across all five conditions are not statistically significantly different (see Figure~\ref{2b-tick}); however, the ``Pay'' condition appears to have the best performance.

\subsection{Participant Comments on The Provided Tips}\label{sec:humancomments}
\begin{figure}
\centering
\subfloat[Normal Configuration]{
\includegraphics[width=0.49\columnwidth]{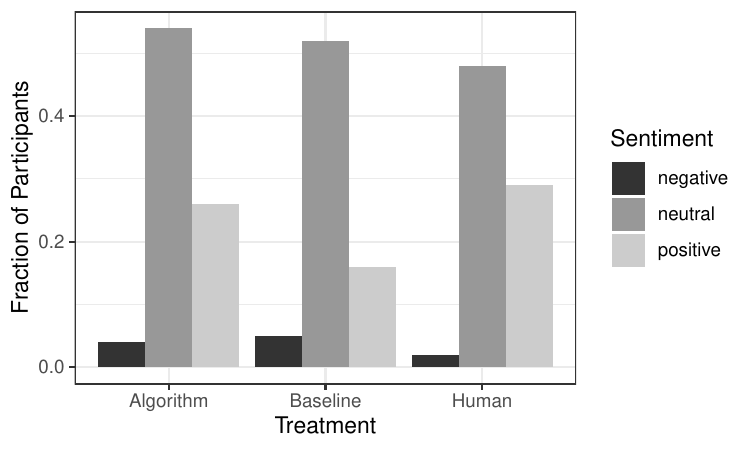}\label{normal-sentiment}}
\subfloat[Disrupted Configuration]{
\includegraphics[width=0.49\columnwidth]{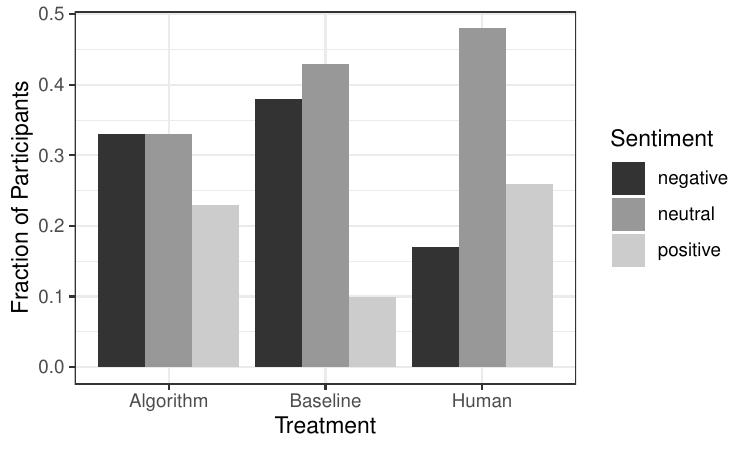}\label{disrupted-sentiment}}\\
\caption{Participant sentiment on the provided tips in post-game survey.}
\label{tip-sentiment}
\end{figure}

Lastly, we manually and independently code participants' sentiment towards the tip they received---positive, neutral, or negative---based on their post-game survey responses to the question \emph{``What did you think about the tip for these last [three/four] rounds and how did you incorporate it in your strategy?''}. Figure~\ref{tip-sentiment} exhibits the breakdown of responses in each condition and configuration.\footnote{We excluded unrelated/uninformative participant responses, so fractions per condition may not add to 1.} For robustness, we also performed these sentiment analyses using two natural language processing approaches, VADER and BERTweet, and find qualitatively similar insights (see Appendix~\ref{app:vader}).

We first observe that, for both configurations, more (fewer) participants in the human condition responded positively (negatively) to the tips compared to the other conditions. In other words, human participants selected tips that would likely be accepted by other humans; these tips offer natural strategies that match human intuition, e.g., we observe that they are often adopted even in the control group. On the other hand, the algorithmic and baseline tips are necessarily part of the optimal (rather than human) policy, and can therefore be counter-intuitive; this is especially apparent in the disrupted configuration, where these tips received substantially more negative feedback (and therefore lower compliance rates, as observed in Figure~\ref{dc}). However, performance and compliance improved over time, implying that it took participants time and effort to correctly incorporate these tips into their workflow and execute an optimal strategy. This is supported by selected excerpts from participant comments presented in Table~\ref{comments-disrupted}.

\begin{table}[!htpb]
\centering
\small
\begin{tabular}{cll}
 \toprule
\begin{tabular}[c]{@{}c@{}}Disrupted\\ Configuration\end{tabular} &
  \multicolumn{1}{c}{\begin{tabular}[c]{@{}c@{}}Our Algorithm's Tip\\ \emph{``Server should cook twice"}\end{tabular}} &
  \multicolumn{1}{c}{\begin{tabular}[c]{@{}c@{}}Human Tip\\ \emph{``Server should cook once"}
  \end{tabular}} \\
  \midrule
{\small Positive} &
{\small
  \begin{tabular}[c]{@{}l@{}} \tabitem ``It was very helpful. It made me focus on \\ making sure the server cooked more even if \\ that was not his obvious strength."\\ 
  \tabitem ``I ignored the tip at first, but later I used the \\ tip and it helped me complete the tasks quickly."\\ 
  \tabitem ``At first I didn't follow it because it seemed \\ counter intuitive since they're slow. But then I \\ had trouble, so I tried it and came out ahead." \\ 
  \tabitem ``I did not listen to it at first because I didn't \\ believe that it would actually help but it did."\\ 
  \tabitem ``The tip was helpful. Without it, I think I \\ would have tried to complete the task without \\ the Server cooking, which would have left \\ someone idle for a long time." \end{tabular}} &
 {\small  \begin{tabular}[c]{@{}l@{}} \tabitem ``It seemed pretty much essential \\ to have server cook once." \\ 
 \tabitem ``I thought it was smart and I \\ used it exclusively."\\ 
 \tabitem ``It was accurate, and I \\ implemented the tip."\\ 
 \tabitem ``I felt that tip was valid, as the \\ server primarily is useful plating/ \\ chopping. I only had him cook once."\\ 
 \tabitem ``It helped because she could cook \\ one burger but any more than that \\ and your ticks would be too high."\end{tabular} } \\
 \midrule
{\small Negative} &
 {\small \begin{tabular}[c]{@{}l@{}} \tabitem ``I think it was a bad tip. I couldn't figure \\ out how to incorporate it  successfully."\\ 
 \tabitem ``Seemed counterintuitive."\\ 
 \tabitem ``It did not help me. I did not use it for round \\ 1, I used it for round 2 and it made me do worse, \\ so round 3 I tried it again and was still unable to \\ do well, so the last round I ignored the tip."\\ 
 \tabitem ``I don't think it helped. I thought having the \\ sous chef cook 3 times would take too long and \\ the point at which I tried it, I decided last minute \\ to have the server cook twice. So I don't think it\\ told me anything useful."\\ 
 \tabitem ``It was not needed since the server took so \\ much longer to cook."\end{tabular} } &
 {\small \begin{tabular}[c]{@{}l@{}} \tabitem ``It was not helpful, because it \\ does not specify when the server \\ should cook."\\ 
 \tabitem ``I used the tip but I don't think \\ it was helpful. The server took long \\ to cook." \\ 
 \tabitem ``I don't agree with this tip."\\ 
 \tabitem ``It was not terribly helpful. I \\ tried to incorporate but it did not \\ seem to help"\\ 
 \tabitem ``It stunk honestly. The server \\ takes forever to cook."\end{tabular} } \\
\bottomrule
\end{tabular}
\caption{Selected excerpts from participant comments on the provided tips (disrupted configuration).}
\label{comments-disrupted}
\end{table}

Many participants felt that the human tip was more accurate since it better matched their intuition, and disagreed with tips that they found counter-intuitive. Some participants even found the human tip to be counter-intuitive since it did not match their intuition from the fully-staffed scenario in prior rounds (i.e., it asks the server to cook once instead of not at all); this result is matched by a post-game survey question for the normal configuration where participants were asked to imagine how their strategy would change in the under-staffed scenario (see Appendix~\ref{app:hypothetical}). As a consequence, compliance and performance suffered. Importantly, we observe that even participants who successfully understood our algorithm's tip (and viewed it favorably at the end) claimed that they did not comply with the tip in earlier rounds of the understaffed scenario. Rather, they needed time to experiment with and without the tip in order to learn its value.

Our results suggest that, to achieve high compliance, it is not sufficient for the participant to just understand the action suggested by the tip (a major focus of the literature on interpretable machine learning); they also have to believe that the suggested action will help improve performance, and be given sufficient time to learn how to correctly incorporate the tip into their workflow.

\section{Concluding Remarks}
We have proposed a novel reinforcement learning algorithm for automatically identifying interpretable tips designed to help improve human sequential decision-making. Our large-scale behavioral study demonstrates that the tips inferred by our algorithm can successfully improve human performance at challenging sequential decision-making tasks, speeding up learning by up to three rounds of in-game experience. Furthermore, we find evidence that participants combine our tips with their own experience to discover additional strategies beyond those stated in the tip. In other words, our algorithm is capable of identifying concise insights and communicating them to humans in a way that expands and improves their knowledge. To the best of our knowledge, our work is the first to empirically demonstrate that reinforcement learning based tips can be used to improve human sequential decision-making.

An important ingredient in our framework is incorporating trace data to identify succinct pieces of information that are most likely to help improve the performance of an average worker. Modern-day organizations have benefited from using customer data to inform new product strategies and to provide personalized offerings to their customers, but the data on their own employees is underused. Trace data is often noisy and too granular to be readable by and immediately useful to humans. Our machine learning framework provides techniques to leverage the largely untapped potential of readily available trace data in pinpointing areas of performance improvement and identifying new practices. Even when the true optimal strategy is unknown, trace data of experienced or high-performing workers can be used with reinforcement learning to identify good strategies.

Furthermore, we provide a number of insights that can aid the design of human-AI interfaces. First, a significant factor in the performance of a tip is whether humans comply with that tip.
Prior work has studied compliance from the perspective of \emph{algorithm aversion} (i.e., whether humans trust other humans more than algorithms)~\citep{eastwood2012people,dietvorst2015algorithm,dietvorst2018overcoming}, as well as interpretability (i.e., whether the human understands the tip)~\citep{doshi2017towards,lage2018human,rudin2019stop}.
Our results suggest that human compliance additionally depends on whether humans believe (based on their intuition and past experience) that the tip improves performance. Second, it takes time for humans to correctly operationalize and adopt the tip---humans need experience to understand why the tip is correct and to discover complementary strategies that further improve their performance. Thus, there is an opportunity for human-AI interfaces to help humans \textit{gradually} adapt their behavior to improve performance.
Third, as evidenced by the baseline tips, even tips that are part of the optimal policy can hurt participant performance if they focus on actions that are not consequential; avoiding such tips is important since it can cause participants to lose trust in machine learning models.
We anticipate that human-AI interfaces will become increasingly prevalent as machine learning algorithms are deployed in real-world settings to help humans make consequential decisions, and a better understanding of how to design trustworthy interfaces will be critical to ensuring that these interfaces ultimately improve human sequential decision-making.

\section*{Acknowledgments}
The authors are grateful to the Wharton Behavioral Lab for invaluable financial and staffing support. They also thank the Mack Institute for Innovation Management and the Wharton Risk Center for financial support.

\bibliographystyle{informs2014}
\bibliography{ms}

\begin{thebibliography}{72}
\providecommand{\natexlab}[1]{#1}
\providecommand{\url}[1]{\texttt{#1}}
\providecommand{\urlprefix}{URL }

\bibitem[{Ak{\c{s}}in et~al.(2021)Ak{\c{s}}in, Deo, J{\'o}nasson,
  \protect\BIBand{} Ramdas}]{akcsin2021learning}
Ak{\c{s}}in Z, Deo S, J{\'o}nasson JO, Ramdas K (2021) Learning from many:
  Partner exposure and team familiarity in fluid teams. \emph{Management
  Science} 67(2):854--874.

\bibitem[{Allcott(2011)}]{allcott2011social}
Allcott H (2011) Social norms and energy conservation. \emph{Journal of public
  Economics} 95(9-10):1082--1095.

\bibitem[{Allon et~al.(2023)Allon, Cohen, Moon, \protect\BIBand{}
  Sinchaisri}]{allon2023managing}
Allon G, Cohen MC, Moon K, Sinchaisri WP (2023) Managing multihoming workers in
  the gig economy. \emph{Available at SSRN 4502968} .

\bibitem[{Argote(2012)}]{argote2012organizational}
Argote L (2012) \emph{Organizational learning: Creating, retaining and
  transferring knowledge} (Springer Science \& Business Media).

\bibitem[{Bastani et~al.(2018)Bastani, Pu, \protect\BIBand{}
  Solar-Lezama}]{bastani2018verifiable}
Bastani O, Pu Y, Solar-Lezama A (2018) Verifiable reinforcement learning via
  policy extraction. \emph{Advances in neural information processing systems},
  2494--2504.

\bibitem[{Bavafa \protect\BIBand{} J{\'o}nasson(2021)}]{bavafa2021recovering}
Bavafa H, J{\'o}nasson JO (2021) Recovering from critical incidents: Evidence
  from paramedic performance. \emph{Manufacturing \& Service Operations
  Management} 23(4):914--932.

\bibitem[{Benjamin et~al.(2010)Benjamin, Choi, \protect\BIBand{}
  Strickland}]{benjamin2010social}
Benjamin DJ, Choi JJ, Strickland AJ (2010) Social identity and preferences.
  \emph{American Economic Review} 100(4):1913--28.

\bibitem[{Bertsimas \protect\BIBand{} Dunn(2017)}]{bertsimas2017optimal}
Bertsimas D, Dunn J (2017) Optimal classification trees. \emph{Machine
  Learning} 106(7):1039--1082.

\bibitem[{Beshears et~al.(2015)Beshears, Choi, Laibson, Madrian,
  \protect\BIBand{} Milkman}]{beshears2015effect}
Beshears J, Choi JJ, Laibson D, Madrian BC, Milkman KL (2015) The effect of
  providing peer information on retirement savings decisions. \emph{The Journal
  of finance} 70(3):1161--1201.

\bibitem[{Brattland et~al.(2018)Brattland, H{\o}iseth, Burkeland, Inderhaug,
  Binder, \protect\BIBand{} Iversen}]{brattland2018learning}
Brattland H, H{\o}iseth JR, Burkeland O, Inderhaug TS, Binder PE, Iversen VC
  (2018) Learning from clients: A qualitative investigation of
  psychotherapists’ reactions to negative verbal feedback.
  \emph{Psychotherapy Research} 28(4):545--559.

\bibitem[{Breiman(2001)}]{breiman2001random}
Breiman L (2001) Random forests. \emph{Machine learning} 45(1):5--32.

\bibitem[{Breiman et~al.(1984)Breiman, Friedman, Stone, \protect\BIBand{}
  Olshen}]{breiman1984classification}
Breiman L, Friedman J, Stone CJ, Olshen RA (1984) \emph{Classification and
  regression trees} (CRC press).

\bibitem[{Buciluǎ et~al.(2006)Buciluǎ, Caruana, \protect\BIBand{}
  Niculescu-Mizil}]{bucilua2006model}
Buciluǎ C, Caruana R, Niculescu-Mizil A (2006) Model compression.
  \emph{Proceedings of the 12th ACM SIGKDD international conference on
  Knowledge discovery and data mining}, 535--541.

\bibitem[{Chan et~al.(2014)Chan, Li, \protect\BIBand{}
  Pierce}]{chan2014learning}
Chan TY, Li J, Pierce L (2014) Learning from peers: Knowledge transfer and
  sales force productivity growth. \emph{Marketing Science} 33(4):463--484.

\bibitem[{Chandrasekaran et~al.(2018)Chandrasekaran, Prabhu, Yadav,
  Chattopadhyay, \protect\BIBand{} Parikh}]{chandrasekaran2018explanations}
Chandrasekaran A, Prabhu V, Yadav D, Chattopadhyay P, Parikh D (2018) Do
  explanations make vqa models more predictable to a human? \emph{arXiv
  preprint arXiv:1810.12366} .

\bibitem[{Chandrasekaran et~al.(2017)Chandrasekaran, Yadav, Chattopadhyay,
  Prabhu, \protect\BIBand{} Parikh}]{chandrasekaran2017takes}
Chandrasekaran A, Yadav D, Chattopadhyay P, Prabhu V, Parikh D (2017) It takes
  two to tango: Towards theory of ai's mind. \emph{arXiv preprint
  arXiv:1704.00717} .

\bibitem[{Chang et~al.(2021)Chang, Jacobson, Shah, Pramanik, \protect\BIBand{}
  Shah}]{chang2021financial}
Chang T, Jacobson M, Shah M, Pramanik R, Shah SB (2021) Financial incentives
  and other nudges do not increase covid-19 vaccinations among the vaccine
  hesitant. Technical report, National Bureau of Economic Research.

\bibitem[{Chui et~al.(2012)Chui, Manyika, \protect\BIBand{}
  Bughin}]{chui2012social}
Chui M, Manyika J, Bughin J (2012) The social economy: Unlocking value and
  productivity through social technologies. Technical report, McKinsey Global
  Institute.

\bibitem[{Dietvorst et~al.(2015)Dietvorst, Simmons, \protect\BIBand{}
  Massey}]{dietvorst2015algorithm}
Dietvorst BJ, Simmons JP, Massey C (2015) Algorithm aversion: People
  erroneously avoid algorithms after seeing them err. \emph{Journal of
  Experimental Psychology: General} 144(1):114.

\bibitem[{Dietvorst et~al.(2018)Dietvorst, Simmons, \protect\BIBand{}
  Massey}]{dietvorst2018overcoming}
Dietvorst BJ, Simmons JP, Massey C (2018) Overcoming algorithm aversion: People
  will use imperfect algorithms if they can (even slightly) modify them.
  \emph{Management Science} 64(3):1155--1170.

\bibitem[{Dorn \protect\BIBand{} Guzdial(2010)}]{dorn2010learning}
Dorn B, Guzdial M (2010) Learning on the job: characterizing the programming
  knowledge and learning strategies of web designers. \emph{Proceedings of the
  SIGCHI Conference on Human Factors in Computing Systems}, 703--712.

\bibitem[{Doshi-Velez \protect\BIBand{} Kim(2017)}]{doshi2017towards}
Doshi-Velez F, Kim B (2017) Towards a rigorous science of interpretable machine
  learning. \emph{arXiv preprint arXiv:1702.08608} .

\bibitem[{Eastwood et~al.(2012)Eastwood, Snook, \protect\BIBand{}
  Luther}]{eastwood2012people}
Eastwood J, Snook B, Luther K (2012) What people want from their professionals:
  Attitudes toward decision-making strategies. \emph{Journal of Behavioral
  Decision Making} 25(5):458--468.

\bibitem[{Fudenberg et~al.(2022)Fudenberg, Kleinberg, Liang, \protect\BIBand{}
  Mullainathan}]{fudenberg2022measuring}
Fudenberg D, Kleinberg J, Liang A, Mullainathan S (2022) Measuring the
  completeness of economic models. \emph{Journal of Political Economy}
  130(4):956--990.

\bibitem[{Fudenberg \protect\BIBand{} Liang(2019)}]{fudenberg2019predicting}
Fudenberg D, Liang A (2019) Predicting and understanding initial play.
  \emph{American Economic Review} 109(12):4112--41.

\bibitem[{F{\"u}gener et~al.(2022)F{\"u}gener, Grahl, Gupta, \protect\BIBand{}
  Ketter}]{fugener2022cognitive}
F{\"u}gener A, Grahl J, Gupta A, Ketter W (2022) Cognitive challenges in
  human--artificial intelligence collaboration: Investigating the path toward
  productive delegation. \emph{Information Systems Research} 33(2):678--696.

\bibitem[{Giuffrida \protect\BIBand{} Torgerson(1997)}]{giuffrida1997should}
Giuffrida A, Torgerson DJ (1997) Should we pay the patient? review of financial
  incentives to enhance patient compliance. \emph{Bmj} 315(7110):703--707.

\bibitem[{Gleicher(2016)}]{gleicher2016framework}
Gleicher M (2016) A framework for considering comprehensibility in modeling.
  \emph{Big data} 4(2):75--88.

\bibitem[{Green \protect\BIBand{} Chen(2019)}]{green2019principles}
Green B, Chen Y (2019) The principles and limits of algorithm-in-the-loop
  decision making. \emph{Proceedings of the ACM on Human-Computer Interaction}
  3(CSCW):1--24.

\bibitem[{Gurvich et~al.(2020)Gurvich, O’Leary, Wang, \protect\BIBand{}
  Van~Mieghem}]{gurvich2020collaboration}
Gurvich I, O’Leary KJ, Wang L, Van~Mieghem JA (2020) Collaboration,
  interruptions, and changeover times: Workflow model and empirical study of
  hospitalist charting. \emph{Manufacturing \& Service Operations Management}
  22(4):754--774.

\bibitem[{Herkenhoff et~al.(2018)Herkenhoff, Lise, Menzio, \protect\BIBand{}
  Phillips}]{herkenhoff2018knowledge}
Herkenhoff K, Lise J, Menzio G, Phillips G (2018) Knowledge diffusion in the
  workplace. Technical report, July 2018. Working Paper, University of
  Minnesota.

\bibitem[{Hinton et~al.(2015)Hinton, Vinyals, \protect\BIBand{}
  Dean}]{hinton2015distilling}
Hinton G, Vinyals O, Dean J (2015) Distilling the knowledge in a neural
  network. \emph{arXiv preprint arXiv:1503.02531} .

\bibitem[{Huckman \protect\BIBand{} Pisano(2006)}]{huckman2006firm}
Huckman RS, Pisano GP (2006) The firm specificity of individual performance:
  Evidence from cardiac surgery. \emph{Management Science} 52(4):473--488.

\bibitem[{Hutto \protect\BIBand{} Gilbert(2014)}]{hutto2014vader}
Hutto C, Gilbert E (2014) Vader: A parsimonious rule-based model for sentiment
  analysis of social media text. \emph{Proceedings of the international AAAI
  conference on web and social media}, volume~8, 216--225.

\bibitem[{Ibanez et~al.(2018)Ibanez, Clark, Huckman, \protect\BIBand{}
  Staats}]{ibanez2018discretionary}
Ibanez MR, Clark JR, Huckman RS, Staats BR (2018) Discretionary task ordering:
  Queue management in radiological services. \emph{Management Science}
  64(9):4389--4407.

\bibitem[{Jarosch et~al.(2019)Jarosch, Oberfield, \protect\BIBand{}
  Rossi-Hansberg}]{jarosch2019learning}
Jarosch G, Oberfield E, Rossi-Hansberg E (2019) Learning from coworkers.
  Technical report, National Bureau of Economic Research.

\bibitem[{Kagan et~al.(2021)Kagan, Leider, \protect\BIBand{}
  Sahin}]{kagan2021dynamic}
Kagan E, Leider S, Sahin O (2021) Dynamic decision-making in operations
  management. \emph{Johns Hopkins Carey Business School Research Paper}
  (21-13).

\bibitem[{Kc \protect\BIBand{} Staats(2012)}]{kc2012accumulating}
Kc DS, Staats BR (2012) Accumulating a portfolio of experience: The effect of
  focal and related experience on surgeon performance. \emph{Manufacturing \&
  Service Operations Management} 14(4):618--633.

\bibitem[{Kim et~al.(2020)Kim, Tong, \protect\BIBand{}
  Peden}]{kim2020admission}
Kim SH, Tong J, Peden C (2020) Admission control biases in hospital unit
  capacity management: How occupancy information hurdles and decision noise
  impact utilization. \emph{Management Science} 66(11):5151--5170.

\bibitem[{Kleinberg et~al.(2015)Kleinberg, Ludwig, Mullainathan,
  \protect\BIBand{} Obermeyer}]{kleinberg2015prediction}
Kleinberg J, Ludwig J, Mullainathan S, Obermeyer Z (2015) Prediction policy
  problems. \emph{American Economic Review} 105(5):491--95.

\bibitem[{Kneusel \protect\BIBand{} Mozer(2017)}]{kneusel2017improving}
Kneusel RT, Mozer MC (2017) Improving human-machine cooperative visual search
  with soft highlighting. \emph{ACM Transactions on Applied Perception (TAP)}
  15(1):1--21.

\bibitem[{Lage et~al.(2018)Lage, Ross, Kim, Gershman, \protect\BIBand{}
  Doshi-Velez}]{lage2018human}
Lage I, Ross AS, Kim B, Gershman SJ, Doshi-Velez F (2018) Human-in-the-loop
  interpretability prior. \emph{Advances in neural information processing
  systems} 31.

\bibitem[{Lai \protect\BIBand{} Tan(2019)}]{lai2019human}
Lai V, Tan C (2019) On human predictions with explanations and predictions of
  machine learning models: A case study on deception detection.
  \emph{Proceedings of the conference on fairness, accountability, and
  transparency}, 29--38.

\bibitem[{Letham et~al.(2015)Letham, Rudin, McCormick, Madigan
  et~al.}]{letham2015interpretable}
Letham B, Rudin C, McCormick TH, Madigan D, et~al. (2015) Interpretable
  classifiers using rules and bayesian analysis: Building a better stroke
  prediction model. \emph{The Annals of Applied Statistics} 9(3):1350--1371.

\bibitem[{Logg et~al.(2019)Logg, Minson, \protect\BIBand{}
  Moore}]{logg2019algorithm}
Logg JM, Minson JA, Moore DA (2019) Algorithm appreciation: People prefer
  algorithmic to human judgment. \emph{Organizational Behavior and Human
  Decision Processes} 151:90--103.

\bibitem[{Lu et~al.(2019)Lu, Lee, Kim, \protect\BIBand{} Danks}]{lu2019good}
Lu J, Lee D, Kim TW, Danks D (2019) Good explanation for algorithmic
  transparency. \emph{Available at SSRN 3503603} .

\bibitem[{Marshall(2020)}]{marshall2020}
Marshall A (2020) Uber changes its rules, and drivers adjust their strategies.
  \urlprefix\url{https://www.wired.com/story/uber-changes-rules-drivers-adjust-strategies/}.

\bibitem[{McIlroy-Young et~al.(2020)McIlroy-Young, Sen, Kleinberg,
  \protect\BIBand{} Anderson}]{mcilroy2020aligning}
McIlroy-Young R, Sen S, Kleinberg J, Anderson A (2020) Aligning superhuman ai
  with human behavior: Chess as a model system. \emph{Proceedings of the 26th
  ACM SIGKDD International Conference on Knowledge Discovery \& Data Mining},
  1677--1687.

\bibitem[{Meyer et~al.(2014)Meyer, Adomavicius, Johnson, Elidrisi, Rush,
  Sperl-Hillen, \protect\BIBand{} O'Connor}]{meyer2014machine}
Meyer G, Adomavicius G, Johnson PE, Elidrisi M, Rush WA, Sperl-Hillen JM,
  O'Connor PJ (2014) A machine learning approach to improving dynamic decision
  making. \emph{Information Systems Research} 25(2):239--263.

\bibitem[{Mnih et~al.(2015)Mnih, Kavukcuoglu, Silver, Rusu, Veness, Bellemare,
  Graves, Riedmiller, Fidjeland, Ostrovski et~al.}]{mnih2015human}
Mnih V, Kavukcuoglu K, Silver D, Rusu AA, Veness J, Bellemare MG, Graves A,
  Riedmiller M, Fidjeland AK, Ostrovski G, et~al. (2015) Human-level control
  through deep reinforcement learning. \emph{Nature} 518(7540):529--533.

\bibitem[{Nonaka \protect\BIBand{} Takeuchi(1995)}]{nonaka1995knowledge}
Nonaka I, Takeuchi H (1995) \emph{The knowledge-creating company: How Japanese
  companies create the dynamics of innovation} (Oxford university press).

\bibitem[{P{\'e}rez et~al.(2021)P{\'e}rez, Giudici, \protect\BIBand{}
  Luque}]{perez2021pysentimiento}
P{\'e}rez JM, Giudici JC, Luque F (2021) pysentimiento: A python toolkit for
  sentiment analysis and socialnlp tasks. \emph{arXiv preprint
  arXiv:2106.09462} .

\bibitem[{Pfeffer et~al.(2000)Pfeffer, Sutton et~al.}]{pfeffer2000knowing}
Pfeffer J, Sutton RI, et~al. (2000) \emph{The knowing-doing gap: How smart
  companies turn knowledge into action} (Harvard business press).

\bibitem[{Puiutta \protect\BIBand{} Veith(2020)}]{puiutta2020explainable}
Puiutta E, Veith EM (2020) Explainable reinforcement learning: A survey.
  \emph{International cross-domain conference for machine learning and
  knowledge extraction}, 77--95 (Springer).

\bibitem[{Ramdas et~al.(2017)Ramdas, Saleh, Stern, \protect\BIBand{}
  Liu}]{ramdas2017variety}
Ramdas K, Saleh K, Stern S, Liu H (2017) Variety and experience: Learning and
  forgetting in the use of surgical devices. \emph{Management Science}
  64(6):2590--2608.

\bibitem[{Ribeiro et~al.(2016)Ribeiro, Singh, \protect\BIBand{}
  Guestrin}]{ribeiro2016should}
Ribeiro MT, Singh S, Guestrin C (2016) " why should i trust you?" explaining
  the predictions of any classifier. \emph{Proceedings of the 22nd ACM SIGKDD
  international conference on knowledge discovery and data mining}, 1135--1144.

\bibitem[{Ross et~al.(2011)Ross, Gordon, \protect\BIBand{}
  Bagnell}]{ross2011reduction}
Ross S, Gordon G, Bagnell D (2011) A reduction of imitation learning and
  structured prediction to no-regret online learning. \emph{Proceedings of the
  fourteenth international conference on artificial intelligence and
  statistics}, 627--635 (JMLR Workshop and Conference Proceedings).

\bibitem[{Rudin(2019)}]{rudin2019stop}
Rudin C (2019) Stop explaining black box machine learning models for high
  stakes decisions and use interpretable models instead. \emph{Nature Machine
  Intelligence} 1(5):206--215.

\bibitem[{Silver et~al.(2016)Silver, Huang, Maddison, Guez, Sifre, Van
  Den~Driessche, Schrittwieser, Antonoglou, Panneershelvam, Lanctot
  et~al.}]{silver2016mastering}
Silver D, Huang A, Maddison CJ, Guez A, Sifre L, Van Den~Driessche G,
  Schrittwieser J, Antonoglou I, Panneershelvam V, Lanctot M, et~al. (2016)
  Mastering the game of go with deep neural networks and tree search.
  \emph{Nature} 529(7587):484--489.

\bibitem[{Song et~al.(2017)Song, Tucker, Murrell, \protect\BIBand{}
  Vinson}]{song2017closing}
Song H, Tucker AL, Murrell KL, Vinson DR (2017) Closing the productivity gap:
  Improving worker productivity through public relative performance feedback
  and validation of best practices. \emph{Management Science} 64(6):2628--2649.

\bibitem[{Spear(2005)}]{spear2005fixing}
Spear SJ (2005) Fixing health care from the inside, today. \emph{Harvard
  business review} 83(9):78.

\bibitem[{Stites et~al.(2021)Stites, Nyre-Yu, Moss, Smutz, \protect\BIBand{}
  Smith}]{stites2021sage}
Stites MC, Nyre-Yu M, Moss B, Smutz C, Smith MR (2021) Sage advice? the impacts
  of explanations for machine learning models on human decision-making in spam
  detection. \emph{International Conference on Human-Computer Interaction},
  269--284 (Springer).

\bibitem[{Sull \protect\BIBand{} Eisenhardt(2015)}]{sull2015simple}
Sull DN, Eisenhardt KM (2015) \emph{Simple rules: How to thrive in a complex
  world} (Houghton Mifflin Harcourt).

\bibitem[{Sun et~al.(2022)Sun, Zhang, Hu, \protect\BIBand{}
  Van~Mieghem}]{sun2022predicting}
Sun J, Zhang DJ, Hu H, Van~Mieghem JA (2022) Predicting human discretion to
  adjust algorithmic prescription: A large-scale field experiment in warehouse
  operations. \emph{Management Science} 68(2):846--865.

\bibitem[{Sutton \protect\BIBand{} Barto(2018)}]{sutton2018reinforcement}
Sutton RS, Barto AG (2018) \emph{Reinforcement learning: An introduction} (MIT
  press).

\bibitem[{Sutton et~al.(2000)Sutton, McAllester, Singh, \protect\BIBand{}
  Mansour}]{sutton2000policy}
Sutton RS, McAllester DA, Singh SP, Mansour Y (2000) Policy gradient methods
  for reinforcement learning with function approximation. \emph{Advances in
  neural information processing systems}, 1057--1063.

\bibitem[{Szulanski(1996)}]{szulanski1996exploring}
Szulanski G (1996) Exploring internal stickiness: Impediments to the transfer
  of best practice within the firm. \emph{Strategic management journal}
  17(S2):27--43.

\bibitem[{Tan \protect\BIBand{} Netessine(2019)}]{tan2019you}
Tan TF, Netessine S (2019) When you work with a superman, will you also fly? an
  empirical study of the impact of coworkers on performance. \emph{Management
  Science} 65(8):3495--3517.

\bibitem[{Tucker et~al.(2002)Tucker, Edmondson, \protect\BIBand{}
  Spear}]{tucker2002problem}
Tucker AL, Edmondson AC, Spear S (2002) When problem solving prevents
  organizational learning. \emph{Journal of Organizational Change Management}
  15(2):122--137.

\bibitem[{Verma et~al.(2018)Verma, Murali, Singh, Kohli, \protect\BIBand{}
  Chaudhuri}]{verma2018programmatically}
Verma A, Murali V, Singh R, Kohli P, Chaudhuri S (2018) Programmatically
  interpretable reinforcement learning. \emph{International Conference on
  Machine Learning}, 5045--5054 (PMLR).

\bibitem[{Wang \protect\BIBand{} Rudin(2015)}]{wang2015falling}
Wang F, Rudin C (2015) Falling rule lists. \emph{Artificial intelligence and
  statistics}, 1013--1022 (PMLR).

\bibitem[{Watkins \protect\BIBand{} Dayan(1992)}]{watkins1992q}
Watkins CJ, Dayan P (1992) Q-learning. \emph{Machine learning} 8(3-4):279--292.

\end{thebibliography}

\newpage

\setcounter{chapter}{0}
\renewcommand{\thechapter}{\Alph{chapter}}
\setcounter{section}{0}
\renewcommand{\theHsection}{\Alph{section}}
\setcounter{table}{0}
\renewcommand{\thetable}{\Alph{section}.\arabic{table}}
\setcounter{figure}{0}
\renewcommand{\thefigure}{\Alph{section}.\arabic{figure}}
\setcounter{equation}{0}
\renewcommand{\theequation}{\Alph{section}.\arabic{equation}}
\clearpage
\pagenumbering{arabic}% resets `page` counter to 1
\renewcommand*{\thepage}{\arabic{page}}

\begin{APPENDICES}
\section{Tip Inference Algorithm}\label{supp-algo}

We first discuss how we formulate the Markov Decision Process (MDP) for our virtual kitchen-management game and the overall structure of the optimal policies for both the fully-staffed and understaffed scenarios. Then, we provide detailed information on the design and implementation of our tip inference algorithm.

\subsection{MDP Formulation}\label{app:mdp}

In our virtual kitchen MDP, the states encode (i) which subtasks have been completed so far across all orders, and (ii) which subtask has been assigned to each virtual worker (if any), as well as how many steps remain to complete this subtask. The actions consist of all possible assignments of available subtasks (i.e., have not yet been assigned) to available virtual workers (i.e., not currently working on any subtask). The reward is $-1$ at each step, until all orders are completed; thus, the total number of steps taken to complete all orders is the negative reward.

\subsection{Optimal Policies}\label{app:opt-policy}

We summarize the optimal policy for each scenario. Note that the optimal policy for the understaffed scenario is more counter-intuitive than that for the fully-staffed scenario.

\textit{Fully-staffed scenario.} Here, the participant has access to all three virtual workers. The optimal number of ticks to complete this scenario is 20. The key insights to achieving optimality are: (i) all three workers should be assigned to chopping in the first time step, (ii) the chef must cook three of the burgers and the sous-chef must cook one (i.e., the second burger), (iii) the server should never cook and must be kept idle when the third burger becomes available for cooking; they should instead wait to be assigned to plating the first cooked burger, (iv) the chef should never plate, (v) the sous-chef must plate exactly one of the burgers, and (vi) none of the three workers should be left idle except in the previous cases.

\textit{Understaffed scenario.} Here, the participant has access to only two virtual workers (e.g., the sous-chef and the server). The optimal number of ticks to complete this scenario is 34. The keys insights to achieving optimality are: (i) both workers should be assigned to chopping in the first time step, (ii) the sous-chef and the server must cook two burgers each, even though the server is slow at cooking, (iii) the sous-chef must choose chopping over cooking after finishing her first chopping task, (iv) the server's first three tasks must be chopping, cooking, and cooking, in that order, (v) the sous-chef must chop three of the four burgers and the server must chop one, (vi) both workers must plate two burgers each, even though the sous-chef is slower at plating, (vii) the second cooked burger must not be served until the third and fourth burgers are cooked, and (viii) both workers must be kept busy at all times.

\subsection{Search Space of Tips}\label{app:rulespace}
Each tip is actually composed of a set of rules inferred by our algorithm. Recall that our algorithm considers tips in the form of an if-then-else statement that says to take a certain action in a certain state. One challenge is the combinatorial nature of our action space---there can be as many as $k!/(k-m)!$ actions, where $m$ is the number of workers and $k=\sum_{j=1}^nk_j$ is the total number of subtasks. The large number of actions can make the tips very specific---e.g., simultaneously assigning three distinct subtasks to three of the virtual workers. Instead, we decompose the action space and consider assigning a single subtask to a single virtual worker. More precisely, we include three features in the predicate $\phi$: (i) the subtask being considered, (ii) the order to which the subtask belongs, and (iii) the virtual worker in consideration. Then, our algorithm considers tips of the form
\begin{align*}
&\text{if }(\text{order}=o~\wedge~\text{subtask}=s~\wedge~\text{virtual worker}=w) \\
&\text{ then }(\text{assign }(o,s)\text{ to }w),
\end{align*}
where $o$ is an order, $s$ is a subtask, and $w$ is a virtual worker.

Even with this action decomposition, we found that these tips are still too complicated for human users to internalize. Thus, we post-process the tips inferred by our algorithm by aggregating over tuples $(o,s,w)$ that have the same $s$ and $w$.\footnote{We experimented with \textit{combinations} of tips in exploratory pilots, and found that AMT workers were unable to operationalize and comply with such complex tips even though they might be part of an optimal strategy.} 
In particular, consider a tip $\rho=(\psi,a)$ with state predicate $\psi$ and action $a$, where $a=(o,s,w)$ is a tuple consisting of a subtask $s$ of an order $o$ that is to be assigned to worker $w$. Our algorithm first aggregates all tips of the form $\rho=(\psi,(o,s,w))$ with the same subtask-worker pair $(s,w)$, to obtain a list $R_{s,w}=\{\rho_1,...,\rho_k\}$ for each $(s,w)$ pair. This $(s,w)$ pair is converted into a tip by counting the number of distinct orders $o$ that occur across $\rho\in R_{s,w}$; if $j$ different orders $o$ occur, then the tip becomes
\begin{align*}
\text{assign }s\text{ to }w,~j\text{ times}.
\end{align*}
For example, instead of considering two separate tips
\begin{align*} 
&\text{if }(\text{order}=\text{burger}_1~\wedge~\text{subtask}=\text{cooking}~\wedge~\text{virtual worker}=\text{chef}) \\
&\qquad\text{then }(\text{assign }(\text{burger}_1,\text{cooking})\text{ to }\text{chef}) \\
&\text{if }(\text{order}=\text{burger}_2~\wedge~\text{subtask}=\text{cooking}~\wedge~\text{virtual worker}=\text{chef}) \\
&\qquad\text{then }(\text{assign }(\text{burger}_2,\text{cooking})\text{ to }\text{chef}),
\end{align*}
we merge them into a tip
\begin{align*}
\text{assign cooking to chef 2 times}.
\end{align*}
Next, the score our algorithm assigns to the aggregated tip $R_{s,w}$ is $J(R_{s,w})=\sum_{\rho\in R_{s,w}}J(\rho)$. Finally, our algorithm chooses the tip $R_{s,w}$ with the highest score.

\subsection{Tip Inference Procedure}\label{app:tipinference}

Next, we describe how our algorithm computes optimal tips for our MDP. While our state space is finite, it is still too large for dynamic programming to be tractable. Instead, we use the policy gradient algorithm to (heuristically) learn an expert policy $\pi_*$~\citep{sutton2000policy}, which uses gradient descent to optimize a policy $\pi_{\theta}$ with parameters $\theta\in\Theta\subseteq\mathbb{R}^{d_{\Theta}}$; we choose $\pi_{\theta}$ to be a neural network. This approach requires that we construct a feature map $\phi:S\to\{0,1\}^d$. Then, $\pi_{\theta}$ takes as input the featurized state $\phi(s)$, and outputs a categorical distribution $\pi_*(a\mid \phi(s))$ over actions $a\in A$. Then, the policy gradient algorithm performs stochastic gradient descent on the objective $J(\pi_{\theta})$, and outputs the best policy $\pi_*=\pi_{\theta^*}$. For the kitchen game MDP, we use state features including whether each subtask of each order is available, the current status of each worker, and the current time step. We take $\pi_{\theta}$ to be a neural network with 50 hidden units; to optimize $J(\pi_{\theta})$, we take 10,000 stochastic gradient steps with a learning rate of $0.001$.

Once we have computed $\pi_*$, we use our tip inference algorithm to learn an estimate $\hat{Q}$ of the $Q$-function $Q^{(\pi_*)}$ for $\pi_*$. We choose $\hat{Q}$ to be a random forest~\citep{breiman2001random}. It operates over the same featurized states as the neural network policy---i.e., it has the form $\hat{Q}(\phi(s),a)\approx Q^{(\pi_*)}(s,a)$. Finally, we apply our algorithm to inferring tips on state-action pairs collected from observing human users playing our game. Since our goal is to help human users improve their performance, we restrict the training dataset to the bottom 25\% performing human users---indeed, our expected improvement is much higher for the bottom 25\% (3.6 tips faster for normal, 4.4 ticks faster for disrupted) than for everyone (2.1 ticks faster for normal, 1.8 ticks faster for disrupted), demonstrating that our tip is expected to be most effective for the bottom quartile of participants. In Appendix~\ref{app:varyhumantip}, we show that our algorithm is robust to this choice, i.e., it produces the same tips if we instead consider the bottom 50\% of participants or all participants.

In addition, we apply two post-processing steps to the set of candidate tips. First, we eliminate tips that apply in less than 10\% of the (featurized) states that occur in the human dataset. This step eliminates high-variance tips that may have large benefit, but are useful only a small fraction of the time; we omit such tips since our estimates of their quality tend to have very high variance. Second, we eliminate tips that disagree with the expert policy more than 50\% of the time---i.e., for a tip $(\psi,a)$, we have $\psi(s)=1$ and $a\neq\pi^*(s)$ for more than 50\% of state-action pairs in the human dataset. This step eliminates tips that have large benefits on average, but frequently offer incorrect advice that can confuse the human user or cause them to distrust our tips. In Appendix~\ref{app:varyhumantip}, we show that this second elimination step is robust to the choice of threshold.

\subsection{Adapting Our Tips to Other Domains}\label{app:otherdomains}

Broadly speaking, a challenge in interpretable machine learning is that the space of interpretable models must be tailored to each new domain, to ensure that the model captures insights relevant to that domain in a human-interpretable way. For our virtual kitchen management game, we have tailored our tips to convey useful information by first inferring if-then rules, and then aggregating these rules into useful tips. The design decisions include both the post-processing steps used to prune and aggregate tips as well as the feature map over states used to infer tips. We arrived at this tradeoff since we wanted tips that could be easily read and understood by human participants while conveying useful information for improving decision-making. The specific choices we made and the post-processing steps we used were informed by our pilot studies.

When applying our algorithm to a new domain, our approach must be adapted so it infers tips that are useful for that domain. In general, the goal should be to produce tips that are as informative as possible under the condition that a human worker can understand what the tip is trying to convey in a reasonable amount of time. For tasks where individual decisions must be made quickly, the tip must be very succinct and easy to understand; in these settings, post-processing strategies such as ours may be necessary to ensure the human understands the tip. Otherwise, more detailed tips such as the original if-then rules can be used.

Finally, we briefly comment on when we expect our algorithm's tip to outperform both the human tip and the baseline algorithm's tip. As our results demonstrate, the human tip tends to have higher compliance since it is usually more intuitive, yet it might be sub-optimal in settings where the optimal policy is complex/unintuitive. As a consequence, we expect the human tip to be more effective when the optimal strategy is intuitive; alternatively, one can imagine scenarios where the optimal policy is simply too complex for the human to determine (even with our algorithm's tip), making it better to go with a more intuitive but less effective strategy. For the baseline tip, we expect it to only be effective when the sequential structure is relatively unimportant for achieving good performance (e.g., in well-mixed MDPs), and the human can focus on achieving good immediate reward. In this case, a strategy that directly tries to maximize immediate rewards may also be effective.

\setcounter{table}{0}
\setcounter{figure}{0}
\section{Additional Details on Experimental Design}\label{app:exp-design}
We perform separate experiments for each of the two configurations of our game. The high-level structure of our experimental design for each configuration is the same; they differ in terms of when we show tips to the participant and which tips we show. Before starting our game, each participant is shown a set of game instructions and comprehension checks; then, they play a practice scenario twice (with an option to skip the second one). The practice scenario is meant to familiarize participants with the game mechanics and the user interface. In this scenario, they manage three identical chefs to make a single food order (different than the burger order used in the main game). Then, they proceed to play the scenarios for the current configuration. Table~\ref{tab:skillmatrix} exhibits the number of time steps needed for each of the virtual workers to complete each of the subtasks required to complete a single burger order.

\begin{table}[!htpb]
\centering
{\footnotesize
\begin{tabular}{lcccc}
\hline
\hline \\[-1.8ex]
       & Chopping meat & Cooking burger & Plating burger  \\  \\[-1.8ex] \hline 
\\[-1.8ex]
Chef      & 1             & 4              & 6             \\  \\[-1.8ex]
Souf-chef & 2             & 8              & 2             \\  \\[-1.8ex]
Server    & 3             & 12             & 1             \\  \\[-1.8ex]
 \hline \hline  
\end{tabular}}
\caption{The (deterministic) number of time steps each virtual worker requires to complete a given subtask.}
\label{tab:skillmatrix}
\end{table}

After completing all scenarios, we give each participant a post-game survey regarding their experience with the game. Each participant receives a participation fee of \$0.10 for each round they complete; they also receive a performance-based bonus based on the number of time steps taken to complete each round. The bonus ranges from \$0.15 to \$0.75 per round. Participants provided informed consent, and all study procedures were approved by our institution's Institutional Review Board.

\subsection{Phase I}

For each configuration, we recruited 200 participants via Amazon Mechanical Turk. 
As part of the post-game survey, we ask the participants to suggest a tip for future players. In particular, we show each participant a comprehensive list of candidate tips and ask them to select the one they believe would most improve the performance of future players. This list of tips is constructed by merging three types of tips: (i) all possible tips in the search space considered by our algorithm (e.g., ``Chef shouldn't plate.''), (ii) generic tips that arise frequently in our exploratory pilot studies (e.g., ``Keep everyone busy at all time.''), (iii) a small number of manually constructed tips obtained by studying the optimal policy (e.g., ``Chef should chop as long as there is no cooking task"). Importantly, this list always contains the top tip inferred using our algorithm.

\begin{figure}
\centering
\subfloat[Normal configuration]{
\includegraphics[width=0.55\columnwidth]{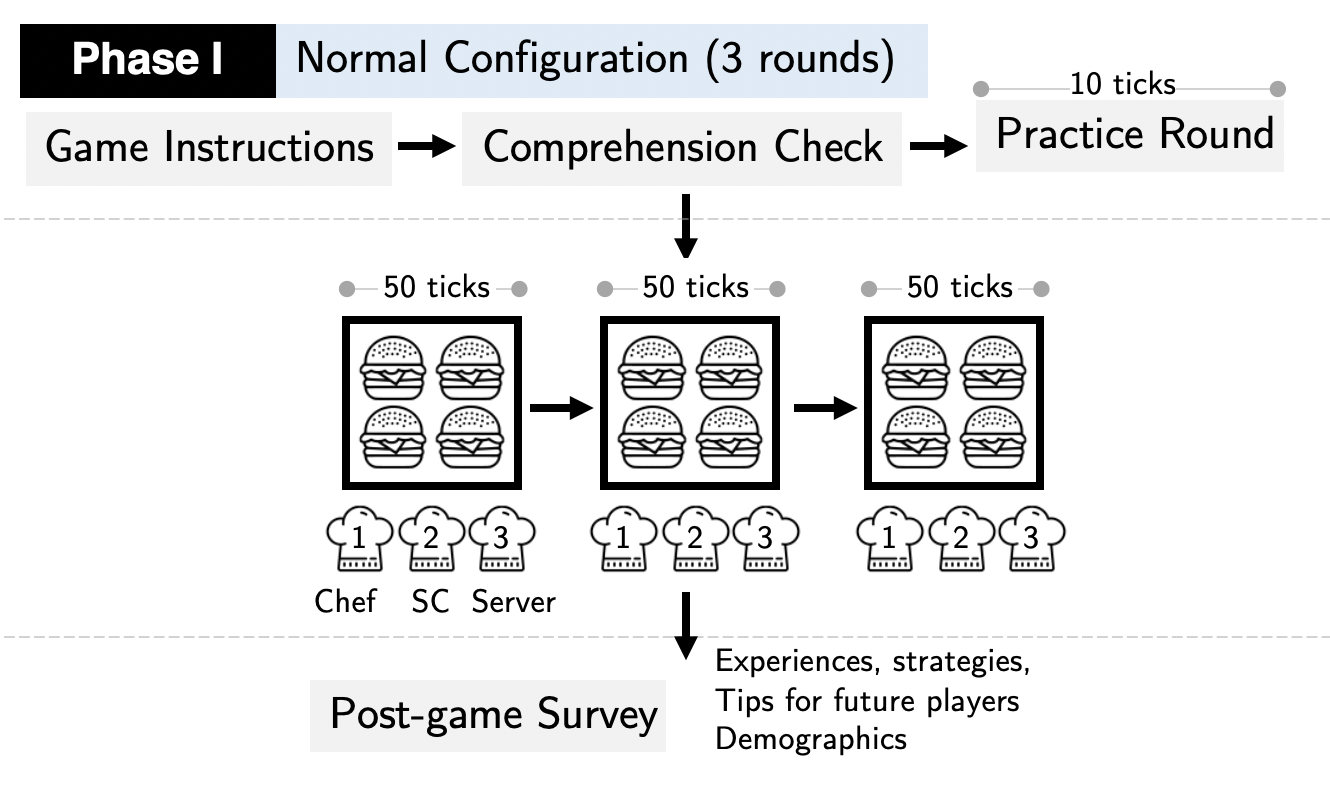}\label{flow1n}
}\\
\subfloat[Disrupted configuration]{
\includegraphics[width=0.985\columnwidth]{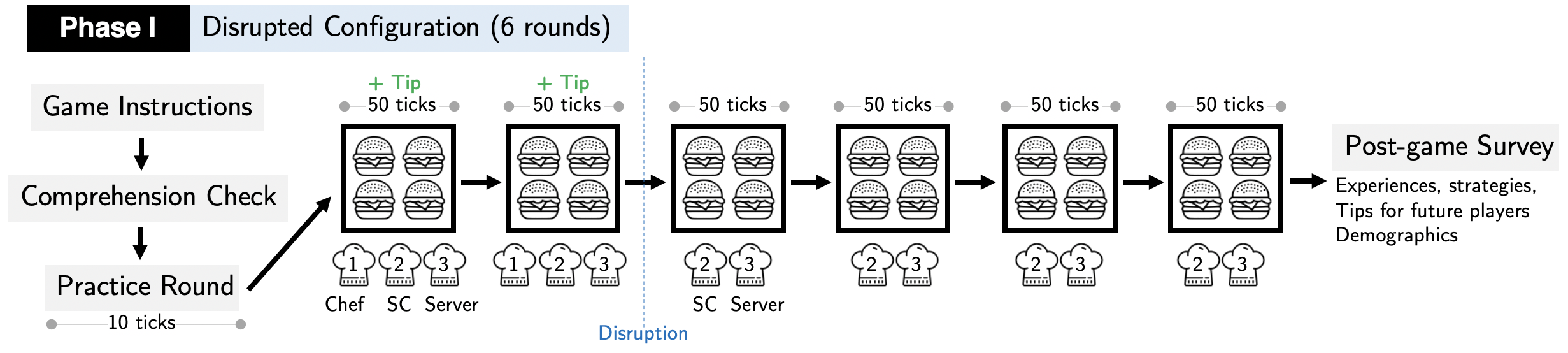}\label{flow1d}
}
\caption{Study flow for Phase I.}
\label{fig:flowphase1}
\end{figure}

\subsection{Inferred Tips}\label{app:inferredtips}

Next, we use participant data from the final round to infer tips in three ways: (i) use our tip inference algorithm in conjunction with the data from Phase I, (ii) do the same with the baseline algorithm, and (iii) rank the candidate tips in the post-game survey based on the number of votes by the participants. As shown in Appendix~\ref{app:varyhumantip}, the human tips are robust to the participant subgroup used to contruct them---i.e., we get the same tips if we restrict only to top performers.

For the normal configuration, 183 participants\footnote{They are 35 years old on average, 57\% are female, and 68\% have at least a two-year degree.} successfully completed the game. The top three tips inferred from each of the sources are reported in Table~\ref{tab:normaltips}. For the algorithm tip, ``Chef should never plate" is selected as it is expected to be the most effective at shortening completion time (2.43 steps). For the baseline tip, our na\"{i}ve algorithm selects ``Chef should chop once" as it is the most frequently observed state-action pair in the data. Finally, for the human tip, ``Strategically leave some workers idle" received the most votes among the participants (28.42\%). It is worth noting that all of the tips most voted by past players are in line with the optimal strategy. The first tip captures the key strategy that some virtual workers should be left idle rather than assigned to a time-consuming task. However, it is less specific than other tips. The second and third tips reflect the information participants could learn from assigning different tasks to different workers during the game: the server spends the most time cooking while the chef spends the most time plating.

\begin{table*}
\centering
{\footnotesize
\begin{tabular}{c|ccc}
\hline
\hline \\[-1.8ex]
Normal & \textbf{Tip \#1} & Tip \#2 & Tip \#3 \\  \\[-1.8ex] \hline  \\[-1.8ex]
Algorithm & \textbf{Chef should never plate} & Server plates three times & Server should skip chopping once \\ \\[-1.8ex]
Baseline & \textbf{Chef should chop once} & Server should plate three times & Sous-chef should plate twice \\ \\[-1.8ex]
\begin{tabular}[c]{@{}c@{}}Human\\ (\% voted)\end{tabular} & \begin{tabular}[c]{@{}c@{}}\textbf{Strategically leave}\\ \textbf{some workers idle}\\ (28\%)\end{tabular} & \begin{tabular}[c]{@{}c@{}}Server should never cook\\ (21\%)\end{tabular} & \begin{tabular}[c]{@{}c@{}}Chef should never plate\\ (13\%)\end{tabular}  \\[-1.8ex]\\ \hline \hline
\end{tabular}}
\caption{Top three tips inferred from different sources for the normal configuration.}
\label{tab:normaltips}
\end{table*}

For the disrupted configuration, 172 participants\footnote{They are 36 years old on average, 62\% are female, and 78\% have at least a two-year degree.} successfully completed the game. Table~\ref{tab:disruptedtips} reports the top three tips inferred from each of the sources. The best algorithm tip is ``Server should cook twice" with the expected completion time reduction of 2.32 steps.
The baseline algorithm chooses ``Sous-chef should plate twice" and the human tip ``Server should cook once" (equivalently ``Sous-chef should cook three times") got the most votes. Unlike the normal configuration, the top two human tips are not part of the optimal policy. In the optimal policy, sous-chef and server should each cook twice. The third human tip does align with the optimal policy; however, it is much less specific than the other tips. This highlights the increased difficulty for humans to identify the optimal strategy in the disrupted configuration compared to the normal configuration.

\begin{table*}
\centering
{\footnotesize
\begin{tabular}{c|ccc}
\hline
\hline \\[-1.8ex]
Disrupted & \textbf{Tip \#1} & Tip \#2 & Tip \#3 \\  \\[-1.8ex] \hline  \\[-1.8ex]
Algorithm & \textbf{Server should cook twice} & Sous-chef should plate once & Server should chop once \\ \\[-1.8ex]
Baseline & \textbf{Sous-chef should plate twice} & Sous-chef should chop three times & Server should cook twice \\ \\[-1.8ex]
\begin{tabular}[c]{@{}c@{}}Human\\ (\% voted)\end{tabular} & \begin{tabular}[c]{@{}c@{}}\textbf{Server should cook once}\\ (28\%)\end{tabular} & \begin{tabular}[c]{@{}c@{}}Server should never cook\\ (24\%)\end{tabular} & \begin{tabular}[c]{@{}c@{}}Keep everyone busy\\ (17\%)\end{tabular} \\[-1.8ex] \\ \hline \hline
\end{tabular}}
\caption{Top three tips inferred from different sources for the disrupted configuration.}
\label{tab:disruptedtips}
\end{table*}

\begin{figure}
\centering
\subfloat[Normal configuration]{
\includegraphics[width=0.55\columnwidth]{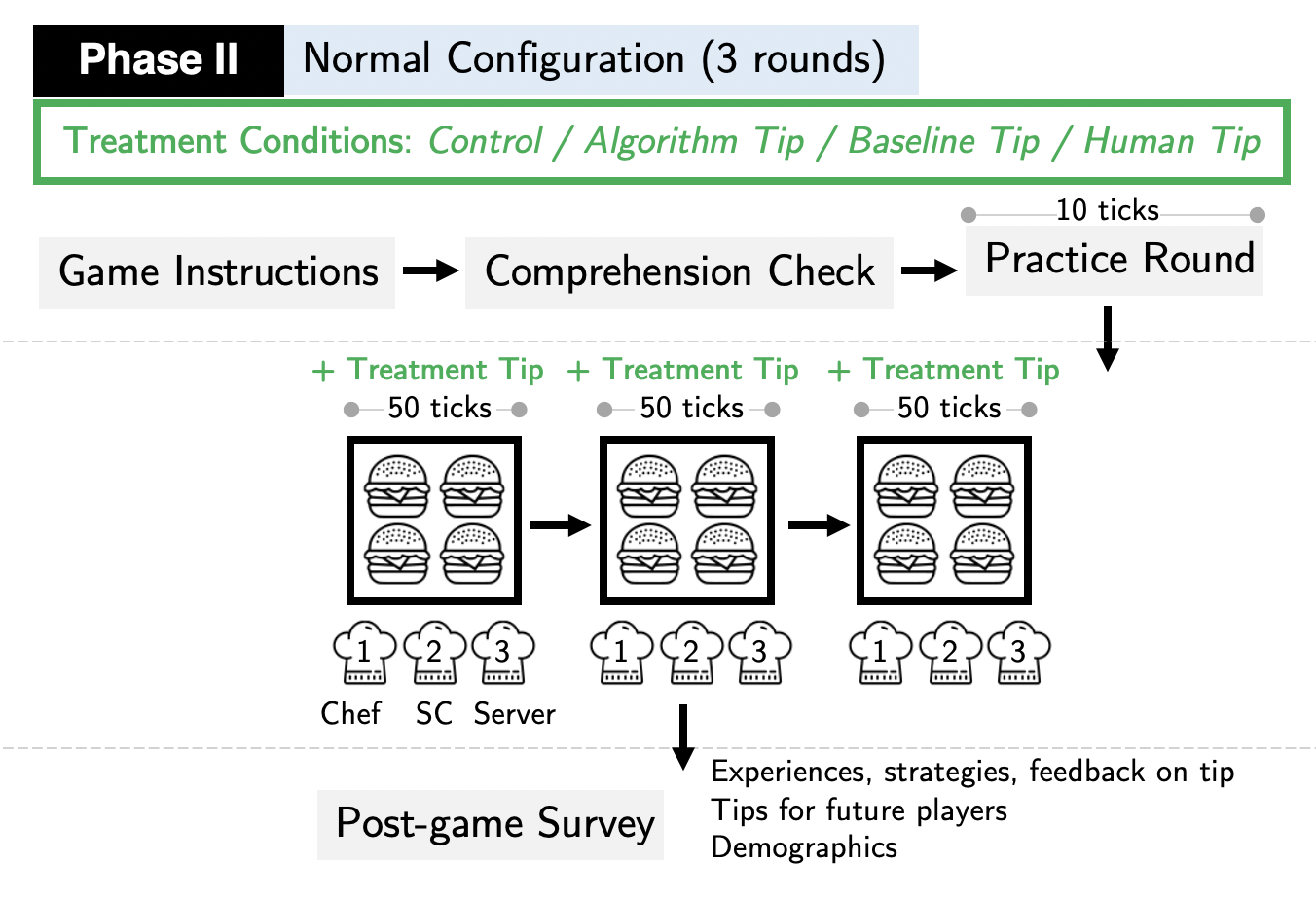}\label{flow2n}
}\\
\subfloat[Disrupted configuration]{
\includegraphics[width=0.985\columnwidth]{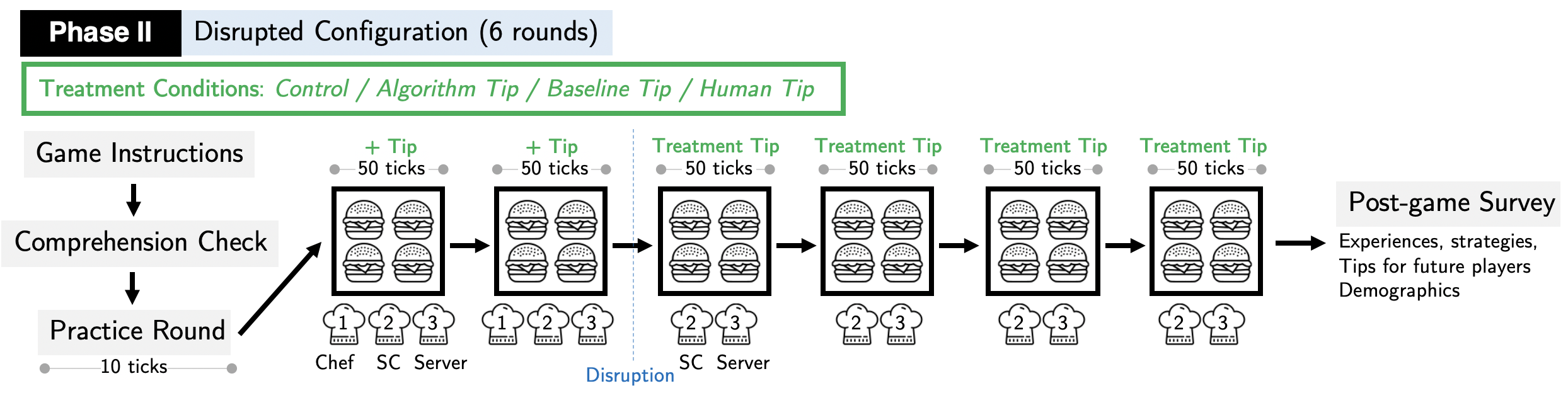}\label{flow2d}
}
\caption{Study flow for Phase II.}
\label{fig:flowphase2}
\end{figure}

\subsection{Phase II}

Next, we evaluate the effectiveness of these tips. 
In this phase, participants are randomly assigned to one of 4 conditions (control, baseline, algorithm, human). We recruited 350 AMT users to play each condition in each configuration, totaling to 2,800 users.  The specific tips we show in each round depends not just on the condition, but also varies from round to round depending on the configuration. For the normal configuration, we show the tip for the current condition in all three rounds. However, for the disrupted configuration, the tip for the current condition is specific to the understaffed scenario. Thus, we only show the tip for the current condition in rounds 3--6; in all conditions, for rounds 1 and 2, we show the tip inferred by our algorithm for the fully-staffed scenario from the normal configuration. By doing so, we ensure that the tip shown during the fully-staffed scenario does not bias our evaluation of the tip for the understaffed scenario.

\subsection{Pay Schemes}\label{app:payschemes}

\textit{Normal configuration.} In Phase I, participants received \$0.30 as a base pay for their participation. In addition, they could earn a performance-based bonus for each of the three rounds of the game. The optimal (e.g., shortest possible) completion time is 20 time steps and the maximum time allowed is 50 time steps. The bonus is as follows: \$0.75 if completing the round in exactly 20 time steps, \$0.35 if completing the round in 21 to 22 time steps, \$0.15 if completing the round in 23 to 26 time steps, or no bonus otherwise. The total pay ranges from \$0.30 to \$2.55, with a mean of \$1.00, a median of \$0.95, and a standard deviation of \$0.56. The sum of the total pay is \$182.15 (183 participants). In Phase II, which was conducted well into the COVID-19 pandemic, we kept the same base pay but slightly increased the tiered bonus: \$1.25 if completing the round in exactly 20 time steps, \$0.60 if completing the round in 21 to 22 time steps, \$0.25 if completing the round in 23 to 26 time steps, or no bonus otherwise.  The total pay ranges from \$0.30 to \$4.05, with a mean of \$1.63, a median of \$1.40, and a standard deviation of \$1.03. The sum of the total pay is \$2,149.70 (1,317 participants).

\textit{Disrupted configuration.} In both phases, participants received \$0.60 as a base pay for their participation. In addition, they could earn a performance-based bonus for each of the six rounds of the game. For the first two rounds, in which they managed a fully-staffed kitchen, the bonus scheme is the same as that of Phase I of the normal configuration. For the last four rounds, in which they managed an understaffed kitchen (optimal completion time is 34 time steps), the bonus is as follows: \$0.75 if completing the round in exactly 34 time steps, \$0.35 if completing the round in 35 to 36 time steps, \$0.15 if completing the round in 37 to 38 time steps, or no bonus otherwise. In Phase I, the total pay ranges from \$0.60 to \$3.30, with a mean of \$1.63, a median of \$1.55, and a standard deviation of \$0.60. The sum of the total pay is \$279.55 (172 participants). In Phase II, the total pay ranges from \$0.60 to \$4.50, with a mean of \$1.81, a median of \$1.75, and a standard deviation of \$0.68. The sum of the total pay is \$1,829.25 (1,011 participants).

\subsection{Hypothetical Disruption}\label{app:hypothetical}

In the post-game survey of both phases of the normal configuration, participants were asked to imagine a hypothetical understaffed scenario where the chef was no longer available in the kitchen and select the best tip that they believed would most help improve performance in such disruption. Note that these participants did not experience a disruption during their gameplay. The list of tips presented to them is the same as the one offered to the participants in the disruption configuration. Consistently in both phases, the tip that received the most votes is ``Server shouldn't cook". Again, this is likely due to the fact that, after three rounds of managing the virtual kitchen under the fully-staffed scenario, the participants potentially learned the optimal policy that the server should not be assigned to cook any burger. Without the actual experience of managing the disruption, they appeared to be biased towards their strategy learned in the fully-staffed scenario, which felt more intuitive to them. This observation highlights one of the key insights of our study that humans' intuition could be far away from the optimal policy, making them less likely to comply to the counter-intuitive tip inferred from our algorithm.

\setcounter{table}{0}
\setcounter{figure}{0}
\section{Additional Details on Experimental Results}\label{app:more-details}

\subsection{Pilot Studies}\label{app:pilot}

We ran small-scale pilot studies in late 2019 and early 2020 both online via AMT and in person at our institutions' behavioral lab. Following best practices, the main objectives of these pilots were to finalize the design of the game and ensure feasibility, but not to estimate treatment effects. For example, we investigated how long it would take the participant to finish each round of the game, calibrated the dishes and their cooking tasks/sequences, figured out how to portray different virtual kitchen workers, and improved the user interface of the game. We also experimented with different framing and presentation of tips to get a preliminary understanding of how participants would notice and respond to the tips. Once we identified our final design of the game, we pre-registered our main study and started collecting data in Summer 2020.

\subsection{Participant Demographics and Performances}\label{app:totalticks}

Table \ref{tab:adddemo} exhibits participants' demographic and gameplay information across our studies. The four groups of participants are not significantly different from one another, except that those playing the disrupted configuration spent slightly longer time to complete the game and found the game to be slightly more difficult, compared to the normal configuration. Tables \ref{tab:detailedperformancenormal} and \ref{tab:detailedperformancedisrupted} show the average performance in each round across phases and treatment conditions for normal and disrupted configurations, respectively. 

\begin{table}[!htpb]
\centering
{\footnotesize
\begin{tabular}{lcccc}
\hline
\hline \\[-1.8ex]
& Phase I: Normal & Phase II: Normal & Phase I: Disrupted & Phase II: Disrupted  \\  \\[-1.8ex] \hline  \\[-1.8ex]
Total & 183 & 1,317 & 172 & 1,011   \\ \\[-1.8ex]
Mean age [range] & 35 [18, 76]		& 33 [18, 74]	& 34 [19, 76]	& 35 [16, 84]	\\  \\[-1.8ex]
Female & 57\%					& 51\% 	& 62\%		& 60\%	\\  \\[-1.8ex]
$\geq$ 2-year degree & 73\%		& 68\%		& 78\%		& 70\% 	\\  \\[-1.8ex]
Median duration & 19 minutes		& 21 min	& 28 min	& 27 min  \\  \\[-1.8ex]
Found the game difficult & 61\%		& 50\%		& 71\%		& 65\% \\  \\[-1.8ex]
Never played similar games & 45\%	& 44\%		& 47\%		& 44\%  \\  \\[-1.8ex]
\hline \hline
\end{tabular}}
\caption{Participants' demographic and gameplay information.}
\label{tab:adddemo}
\end{table}

\begin{table}[!htpb]
\centering
{\footnotesize
\begin{tabular}{lccccc}
\hline
\hline \\[-1.8ex]
& Phase I & Phase II: Control & Phase II: Algorithm & Phase II: Baseline & Phase II: Human  \\  \\[-1.8ex] \hline  \\[-1.8ex]
Round 1 & 25.73 & 26.03 & 25.04 & 26.01 & 26.16  \\ \\[-1.8ex]
Round 2 & 25.02	& 24.46 & 23.29 & 24.71 & 25.06	\\  \\[-1.8ex]
Round 3 & 23.74	& 23.86 & 22.99 & 24.04 & 24.06	\\  \\[-1.8ex]
\hline \hline
\end{tabular}}
\caption{Average performance by treatment condition and round (normal configuration).}
\label{tab:detailedperformancenormal}
\end{table}

\begin{table}[!htpb]
\centering
{\footnotesize
\begin{tabular}{lccccc}
\hline
\hline \\[-1.8ex]
& Phase I & Phase II: Control & Phase II: Algorithm & Phase II: Baseline & Phase II: Human  \\  \\[-1.8ex] \hline  \\[-1.8ex]
Round 1 & 24.35 & 24.26 & 24.18 & 24.77 & 24.64  \\ \\[-1.8ex]
Round 2 & 22.87 & 22.38 & 22.95 & 23.08 & 22.69	\\  \\[-1.8ex]
Round 3 & 38.75 & 38.73 & 38.19 & 38.74 & 38.26	\\  \\[-1.8ex]
Round 4 & 38.39 & 38.21 & 37.77 & 38.38 & 37.85 \\ \\[-1.8ex]
Round 5 & 38.25 & 38.17 & 37.25 & 38.42 & 37.62	\\  \\[-1.8ex]
Round 6 & 37.96 & 37.82 & 37.14 & 38.37 & 37.62	\\  \\[-1.8ex]
\hline \hline
\end{tabular}}
\caption{Average performance by treatment condition and round (disrupted configuration).}
\label{tab:detailedperformancedisrupted}
\end{table}

\begin{figure}[!htpb]
\centering
\subfloat[Final Round Performance (Normal)]{
\includegraphics[width=0.47\textwidth]{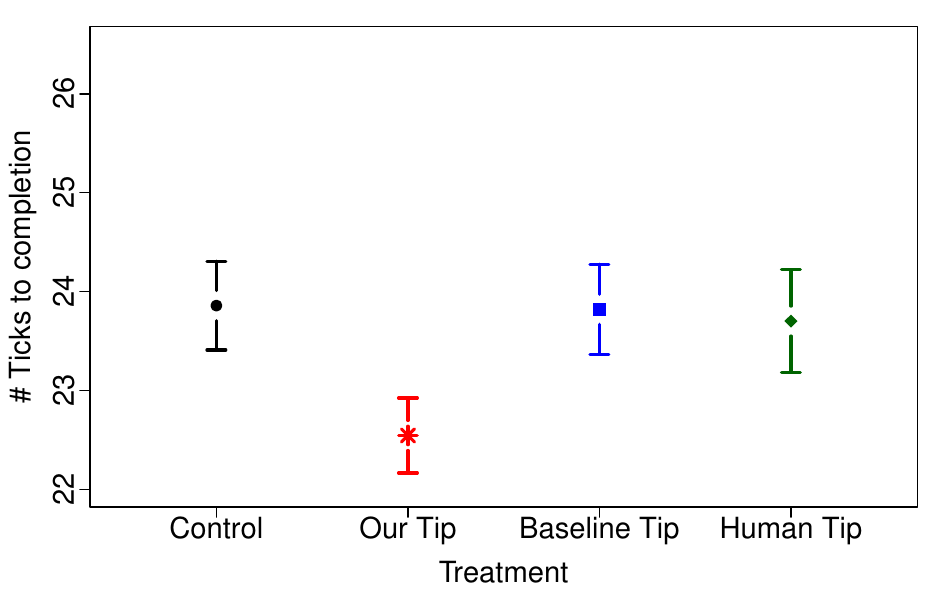}\label{phase2nfint2}}\quad
\subfloat[Final Round Performance (Disrupted)]{
\includegraphics[width=0.47\textwidth]{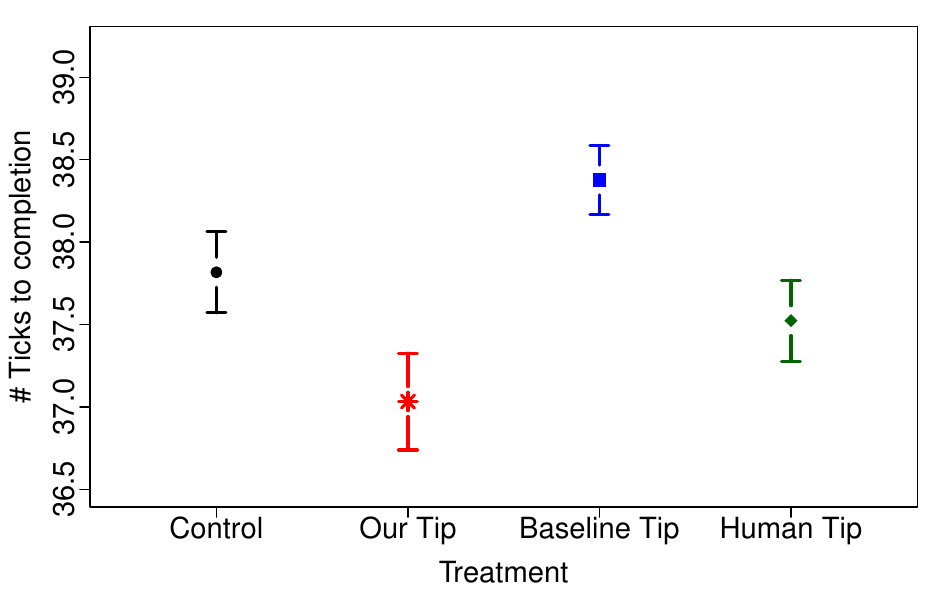}\label{phase2dfint2}}\\
\subfloat[Performance over Time (Normal)]{
\includegraphics[width=0.46\textwidth]{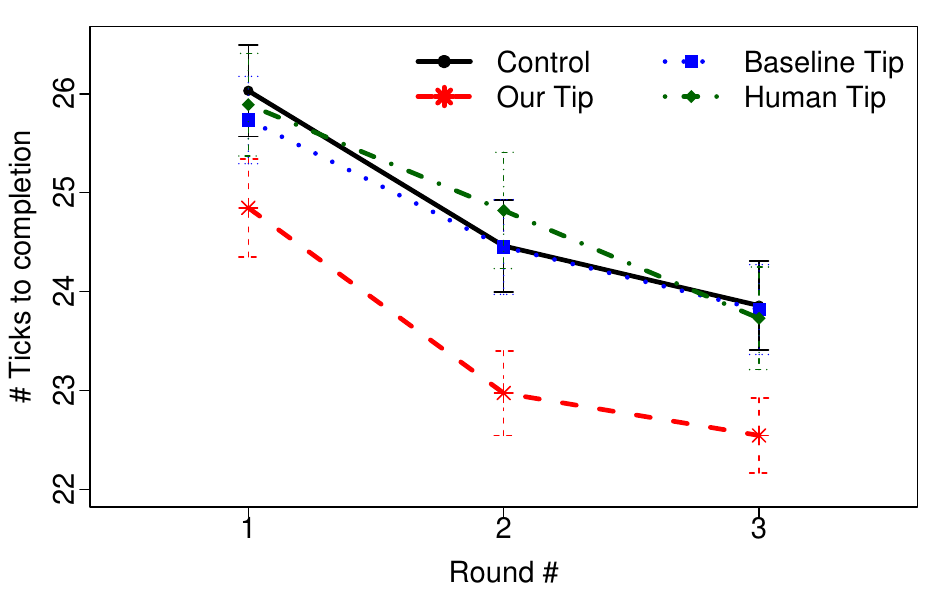}\label{phase2nticks2}}\quad
\subfloat[Performance over Time (Disrupted)]{
\includegraphics[width=0.47\textwidth]{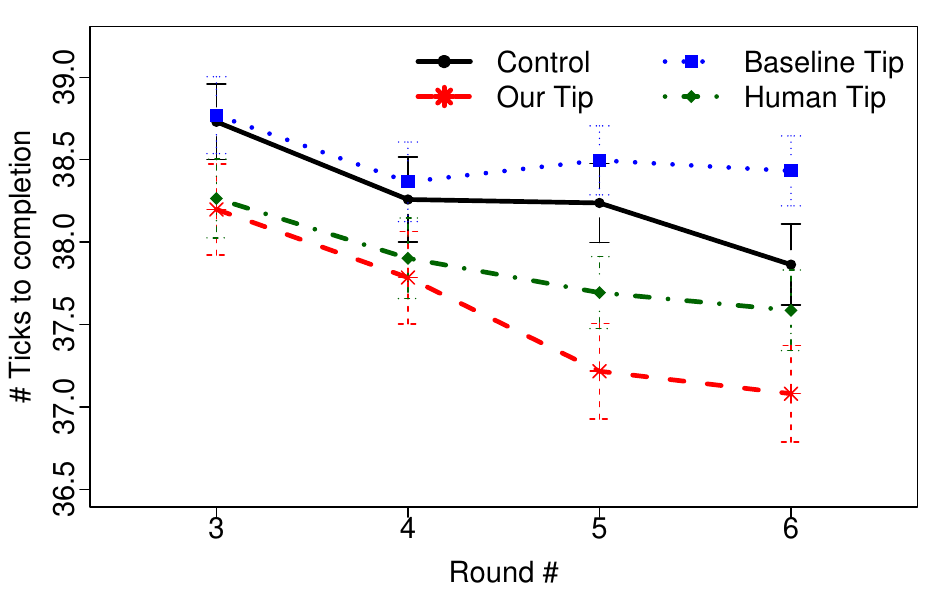}\label{phase2dticks2}}
\caption{Phase II Participant Performance. \footnotesize The subfigures depict various views of participant performance across conditions in the normal (left) and disrupted (right) configurations. The top row shows performance in the last round of the configuration, the bottom row shows how participant performance improves over time.}
\label{fig:totalticks}
\end{figure}

\begin{figure}[!htpb]
\centering
\subfloat[Source of Tip]{
\includegraphics[width=0.47\textwidth]{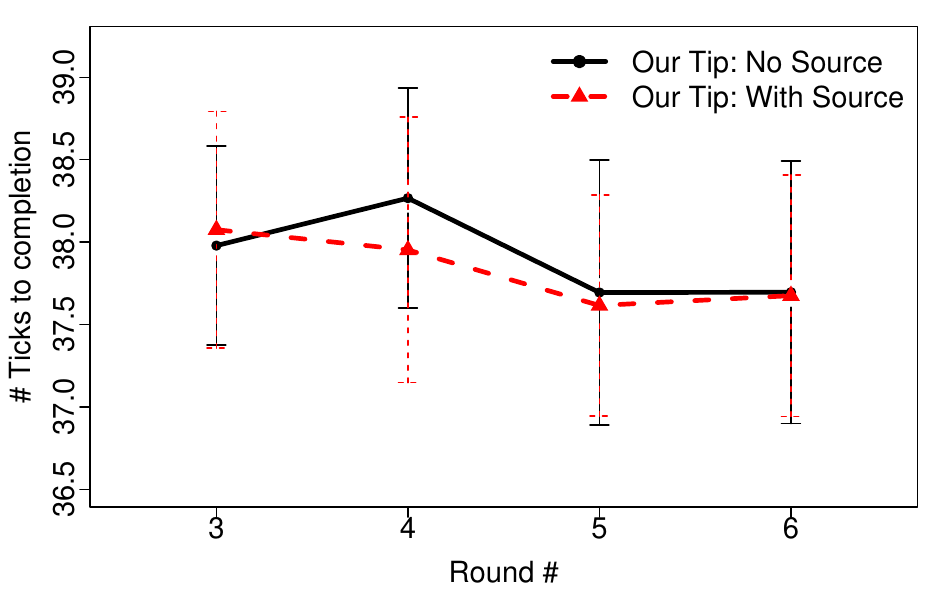}\label{fig:2aticks}}\quad
\subfloat[Interventions for Compliance]{
\includegraphics[width=0.47\textwidth]{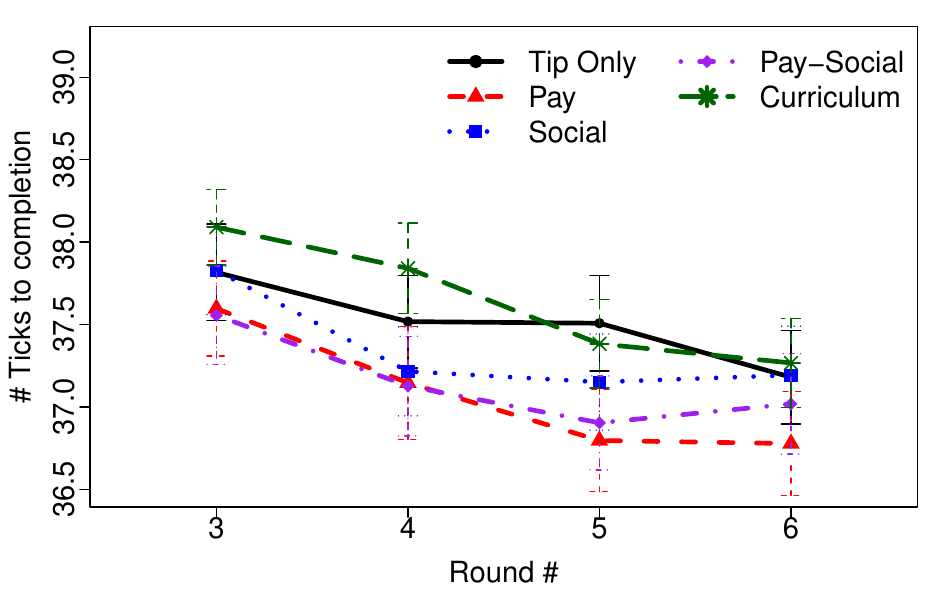}\label{fig:2bticks}}
\caption{Phase II Participant Performance for Follow-Up Studies. \footnotesize The subfigures depict performance over time across conditions in the Source of Tip (left) and the Interventions for Compliance (right) studies.}
\label{fig:study2ticks}
\end{figure}

Figure~\ref{fig:totalticks} illustrates participant performance in Phase II of our primary experiment in both the normal and disrupted configurations (same information as Figure~\ref{fig:performance} but in terms of raw ticks) but in the number of ticks instead of the relative percentage of ticks above optimal. Analogously, Figure~\ref{fig:study2ticks} illustrates participant performance in terms of raw ticks for our two follow-up experiments in Section~\ref{sec:algoaversion}.

\subsection{Learning Beyond Tips under The Normal Configuration}\label{app:crossnormal}
Compared to the disrupted configuration, the normal configuration is much easier. We find that participants across all conditions cross-comply with all other tips: not to assign plating to the chef (Figure \ref{nct}), strategically leave some virtual workers idle (Figure \ref{ncs}), and let the chef chop only once (Figure \ref{ncb}). These tips are all consistent with the optimal policy, suggesting that participants generally learn over time to improve their performance regardless of the condition. Interestingly, participants in the algorithm condition have similar or higher cross-compliance compared to the other conditions. This result suggests that our tip is the most effective as the information it encompasses the information conveyed by the other tips. At a high level, the optimal policy for the fully-staffed scenario has the chef cook most of the dishes, has the server plate most of the dishes, and never assigns the chef to plate or the server to cook. We observe that participants generally recover these optimal strategies as they gain more experience with the game. For instance, the fraction of participants in each arm that never assign cooking to the server in each round, as if they were following the tip ``Server shouldn't cook", increases over time and within each round the fractions are not statistically different among the arms (see Figure \ref{ncu}). This result suggests that participants can uncover this unshown rule by themselves across all conditions.

\begin{figure}
\centering
\subfloat[Our Tip:\\``Chef Shouldn't Plate"]{
\includegraphics[width=0.24\columnwidth]{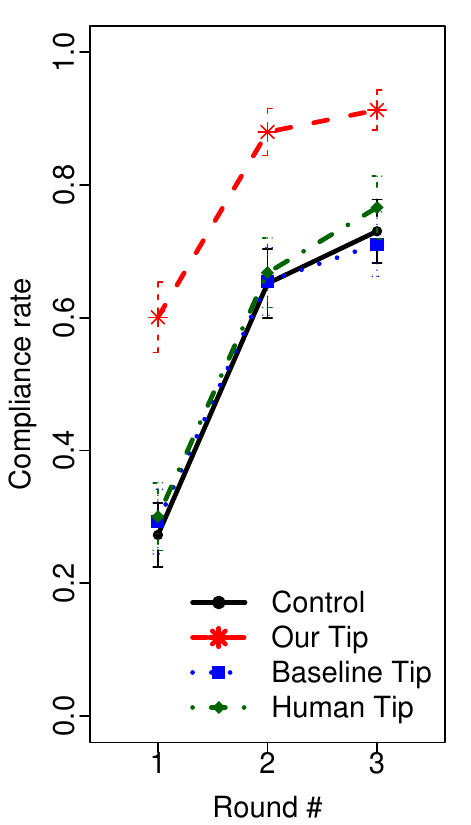}\label{nct}}
\subfloat[Human Tip:\\``Leave Some Workers Idle"]{
\includegraphics[width=0.24\columnwidth]{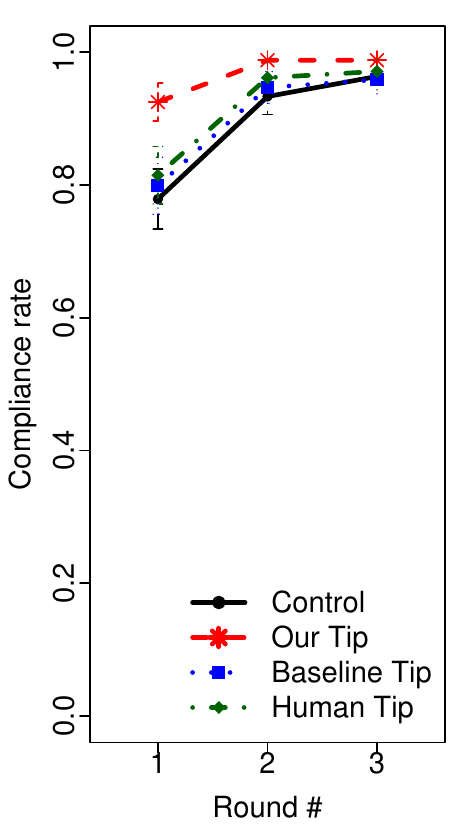}\label{ncs}}
\subfloat[Baseline Tip:\\ ``Chef Chops Once"]{
\includegraphics[width=0.24\columnwidth]{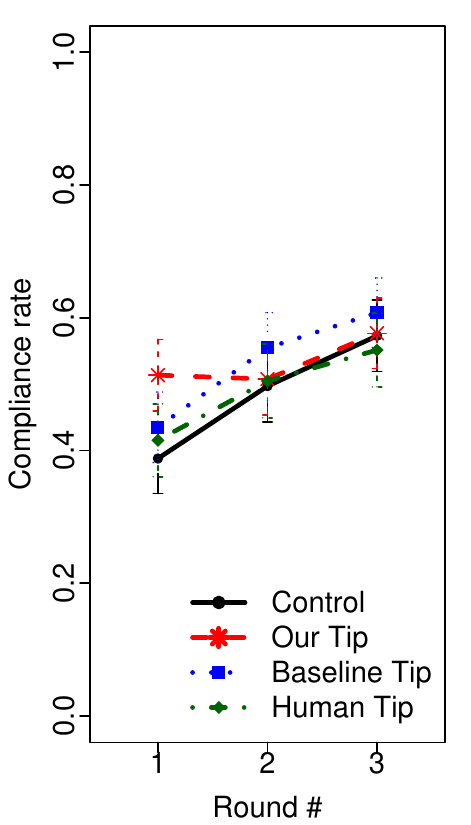}\label{ncb}}
\subfloat[Unshown Tip:\\ ``Server Shouldn't Cook"]{
\includegraphics[width=0.24\columnwidth]{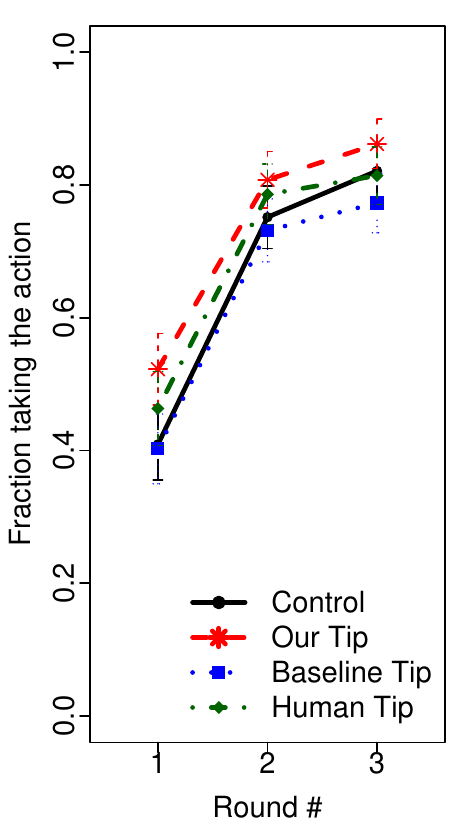}\label{ncu}}
\caption{Learning beyond Tip (Normal Configuration). Panels (a)-(c) show the rate at which participants in each condition cross-comply with each offered tip over time in the normal configuration. Panel (d) shows analogous results for a rule that is part of the optimal policy but was not shown as a tip in any condition.}
\label{ncc}
\end{figure}

\subsection{Robustness Checks for Tip Construction}
\label{app:varyhumantip}

We now examine the robustness of our results to several design choices.

\textit{Robustness to subpopulation used to construct human tip.} In our experiment, we desginated the human tip as the highest-voted tip among \textit{all} participants from Phase I. However, one may hypothesize that peer feedback is more effective if it is inferred from \textit{top} performers. To this end, we examine what tips we would have derived if we had restricted to the highest-voted suggestion by the (absolute top, top 5\%, top 10\%, and top 25\%) of Phase I performers. We define performance ``Top $X$\% performers'' refers to the Phase I participants who had the highest performance in Round 6 (final round). The human tip remained the \textit{same} across all of these subgroups in both configurations; in other words, focusing on best performers would not change our results, since we would still identify the same human tips (see Table~\ref{tab:newhumantips}).

\begin{table}[!htpb]
\centering
\small
\begin{tabular}{l|l|l}
\hline
\multicolumn{1}{c|}{Threshold for elimination} & \multicolumn{1}{c|}{Normal configuration} & \multicolumn{1}{c}{Disrupted configuration} \\ \hline
Everyone (Original) & Leave some idle & Server cooks once \\
Best performers & Leave some idle  & Server cooks once \\
Top 5\% performers & Leave some idle & Server cooks once \\
Top 10\% performers & Leave some idle & Server cooks once \\
Top 25\% performers & Leave some idle & Server cooks once \\ \hline
\end{tabular}
\caption{Top tips for the ``Human'' condition based on various subpopulations}
\label{tab:newhumantips}
\end{table}

\textit{Robustness of our algorithm's tip to varying quantiles of human trace data.} As described in Appendix~\ref{app:tipinference}, for computing our algorithm’s tip, we trained on the bottom 25\% of participants since the goal of our paper is to help improve the performance of workers who are weakest at the given problem. Indeed, our expected improvement is much higher for the bottom 25\% (3.6 tips faster for normal, 4.4 ticks faster for disrupted) than for everyone (2.1 ticks faster for normal, 1.8 ticks faster for disrupted), demonstrating that our tip is expected to be most effective for the bottom quartile of participants. To analyze the robustness of our approach, we considered two alternatives: using the bottom 50\% or using everyone. As shown in Table~\ref{tab:humantracethreshold}, our algorithm produces the same tips using these alternative strategies.

\begin{table}[!htpb]
\centering
\small
\caption{Top tips by our algorithm based on varying quantiles of Phase I human trace data}
\label{tab:humantracethreshold}
\begin{tabular}{l|l|l}
\hline
\multicolumn{1}{c|}{Criteria for tip selection} & \multicolumn{1}{c|}{Normal configuration} & \multicolumn{1}{c}{Disrupted configuration} \\ \hline
Bottom 25\% (Original) & Chef should never plate & Server cooks twice \\
Bottom 50\% & Chef should never plate & Server cooks twice \\
Everyone & Chef should never plate & Server cooks twice \\
\hline
\end{tabular}
\end{table}

\textit{Robustness of our algorithm's tip to elimination threshold.} As described in Appendix~\ref{app:tipinference}, for computing our algorithm's tip, we used a post-processing step where we pruned tips that that disagree with the optimal policy more than 50\% of the time. To evaluate robustness, we now consider constructing our algorithm's tip based on alternative thresholds of 30\% and 70\%. Results are shown in Table~\ref{tab:humantipthreshold}; as can be seen, the tip is robust to the choice of this threshold.

\begin{table}[!htpb]
\centering
\small
\begin{tabular}{l|l|l}
\hline
\multicolumn{1}{c|}{Threshold for elimination} & \multicolumn{1}{c|}{Normal configuration} & \multicolumn{1}{c}{Disrupted configuration} \\ \hline
30\% & Leave some idle & Server cooks once \\
50\% (Original) & Leave some idle  & Server cooks once \\
70\% & Leave some idle & Server cooks once \\
\hline
\end{tabular}
\caption{Top tips for the ``Human'' condition by various elimination criteria}
\label{tab:humantipthreshold}
\end{table}

\subsection{Sentiment Analysis of Participants' Qualitative Responses Using Machine Learning}\label{app:vader}

In Section~\ref{sec:humancomments}, we conducted a sentiment analysis for the qualitative responses to the post-study survey question ``How did you incorporate the provided tip into your gameplay?''; we presented results through manual coding by a human who is not part of the research team. To ensure robustness, we consider alternatively using two natural language processing approaches for the same task.

First, we use a lexicon and rule-based sentiment analysis tool called VADER~\citep{hutto2014vader}. VADER provides an API $polarity\_scores$ that returns four values for each set of texts, including $pos$ for positive sentiment, $neg$ for negative sentiment, $neu$ for neutral sentiment, and a compound score. The compound score is normalized to be between $-1$ (most extremely negative) and $+1$ (most extremely positive). According to the developer of VADER, among these four scores, the compound score is ``the one most commonly used for sentiment analysis by most researchers, including the authors.''

However, in practice, although VADER generates a reasonable score for most inputs, it does not perform as well for more complex content. For example, it might classify a response as positive because it sees the word ``like" repeatedly, when the word is actually being used as a preposition or conjunction. To address this issue, we instead consider a pre-trained BERT model to classify sentiments. Specifically, we used BERTweet Base Sentiment Analysis, a RoBERTa model trained on English tweets~\citep{perez2021pysentimiento}. We used the pipeline API provided by Hugging Face's Transformers library. Because the model returns three separate scores: \emph{Positive}, \emph{Neutral}, and \emph{Negative}, to get a final compound score as we had before, we calculated $Positive - Negative$.

Table~\ref{tab:vader} exhibits all the scores from our VADER and BERTweet analyses using the responses among participants in the disrupted configuration of Phase II. We find that our results are highly consistent with our initial approach using the human coder---i.e., the human tip consistently has higher positive, lower negative, and higher compound scores than our algorithm's tip across both approaches, suggesting that participants consistently perceive our algorithm's tip to be less favorable.

\begin{table}[!htpb]
\centering
\small
\begin{tabular}{|l|cccc|cccc|}
\hline
Disrupted & VADER$+$ & VADER$~$ & VADER$-$ & VADER  & B$+$ & B$~$ & B$-$ & BERTweet  \\ \hline
Algorithm & 0.0922    & 0.8515    & 0.0563    & 0.0256 & 0.2113       & 0.4712       & 0.3176       & $-$0.1063 \\
Human     & 0.1042    & 0.8601    & 0.0358    & 0.1305 & 0.2473       & 0.6100       & 0.1427       & 0.1046    \\
Baseline  & 0.0979    & 0.8339    & 0.0682    & 0.0600 & 0.1315       & 0.5029       & 0.3657       & $-$0.2342 \\ \hline
\end{tabular}
\caption{Positive, neutral, negative, and compound sentiment scores from VADER and BERTweet, using Phase II's disrupted configuration results. $+, ~, -$ refer to positive, neutral, and negative scores, respectively.}
\label{tab:vader}
\end{table}

\setcounter{table}{0}
\setcounter{figure}{0}
\section{Screenshots of Our Virtual Kitchen Management Game}\label{app:screenshots}
Finally, we provide screenshots to illustrate our experimental design. Figures~\ref{fig:intro} and~\ref{fig:intro2} show the introduction to the task shown to participants explaining various concepts in the game. Figures~\ref{fig:game1} and~\ref{fig:game3} show instructions for the fully-staffed and understaffed scenarios, respectively, shown to participants. Finally, Figure~\ref{fig:game4} shows the payment information shown to the participants.

\begin{figure}
\centering
\subfloat[Introduction to the interface]{
\includegraphics[width=0.46\columnwidth]{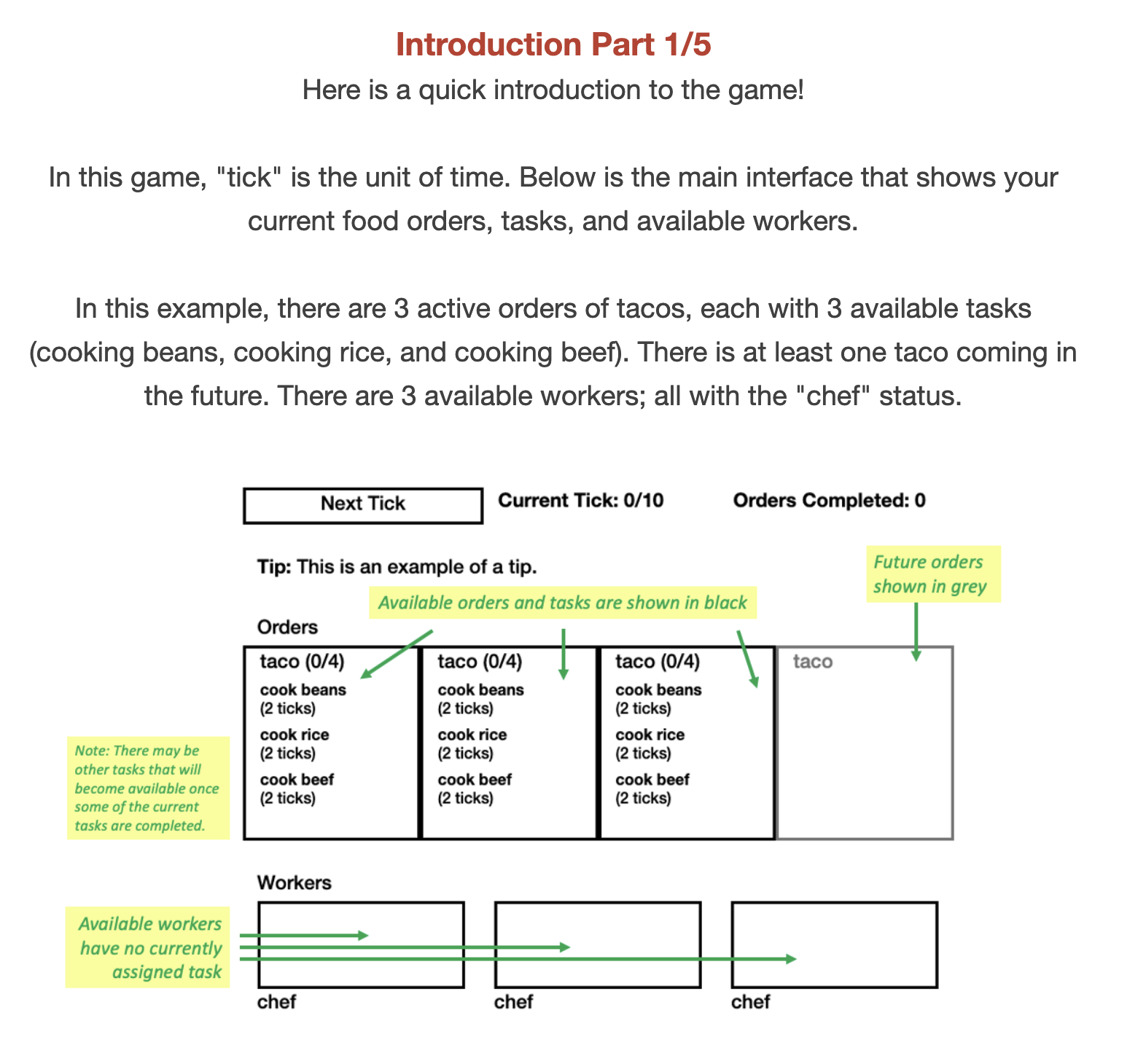}\label{i2}}\quad
\subfloat[Introduction to the subtasks]{
\includegraphics[width=0.46\columnwidth]{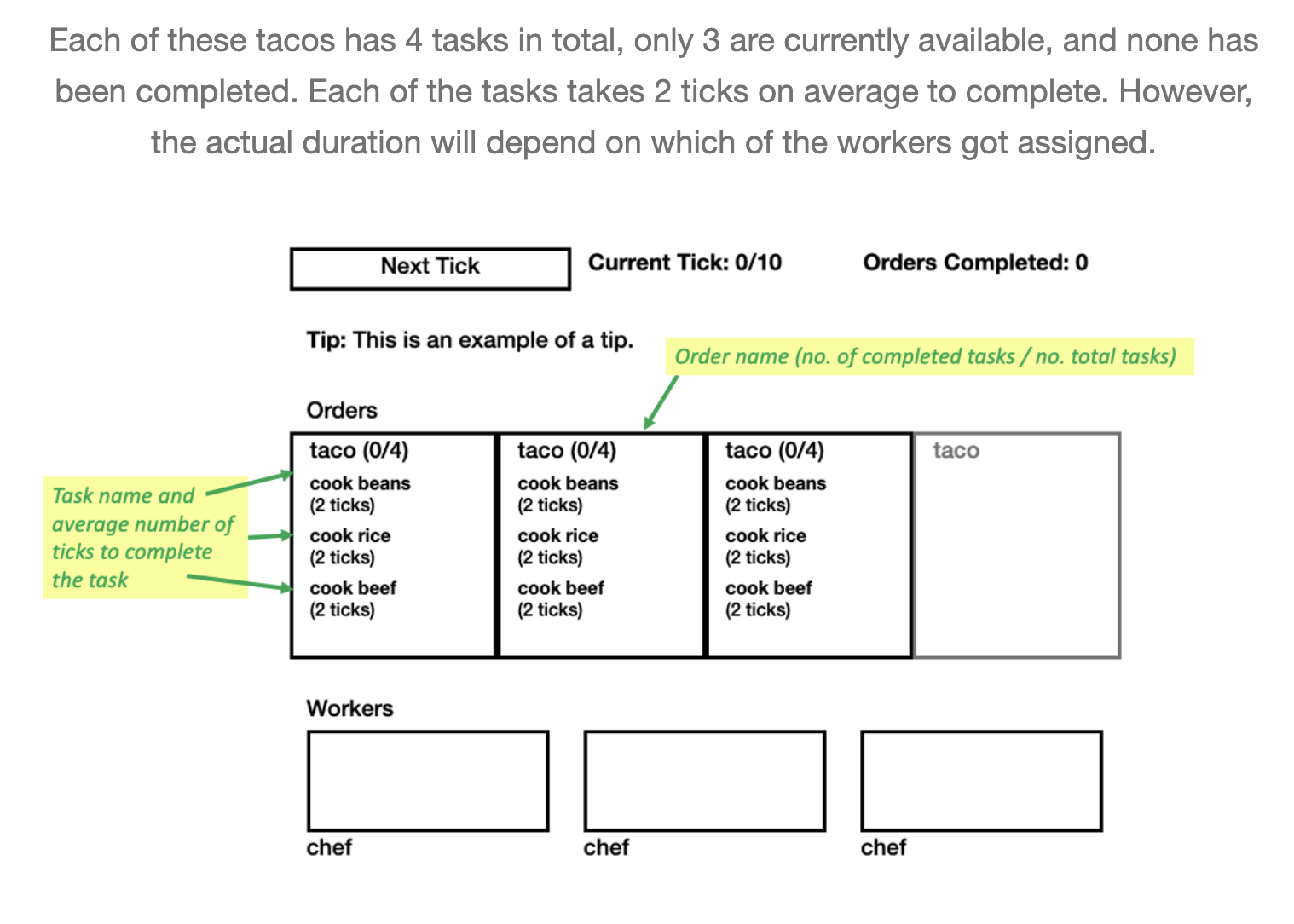}\label{i3}}\\
\subfloat[Introduction to task assignment]{
\includegraphics[width=0.46\columnwidth]{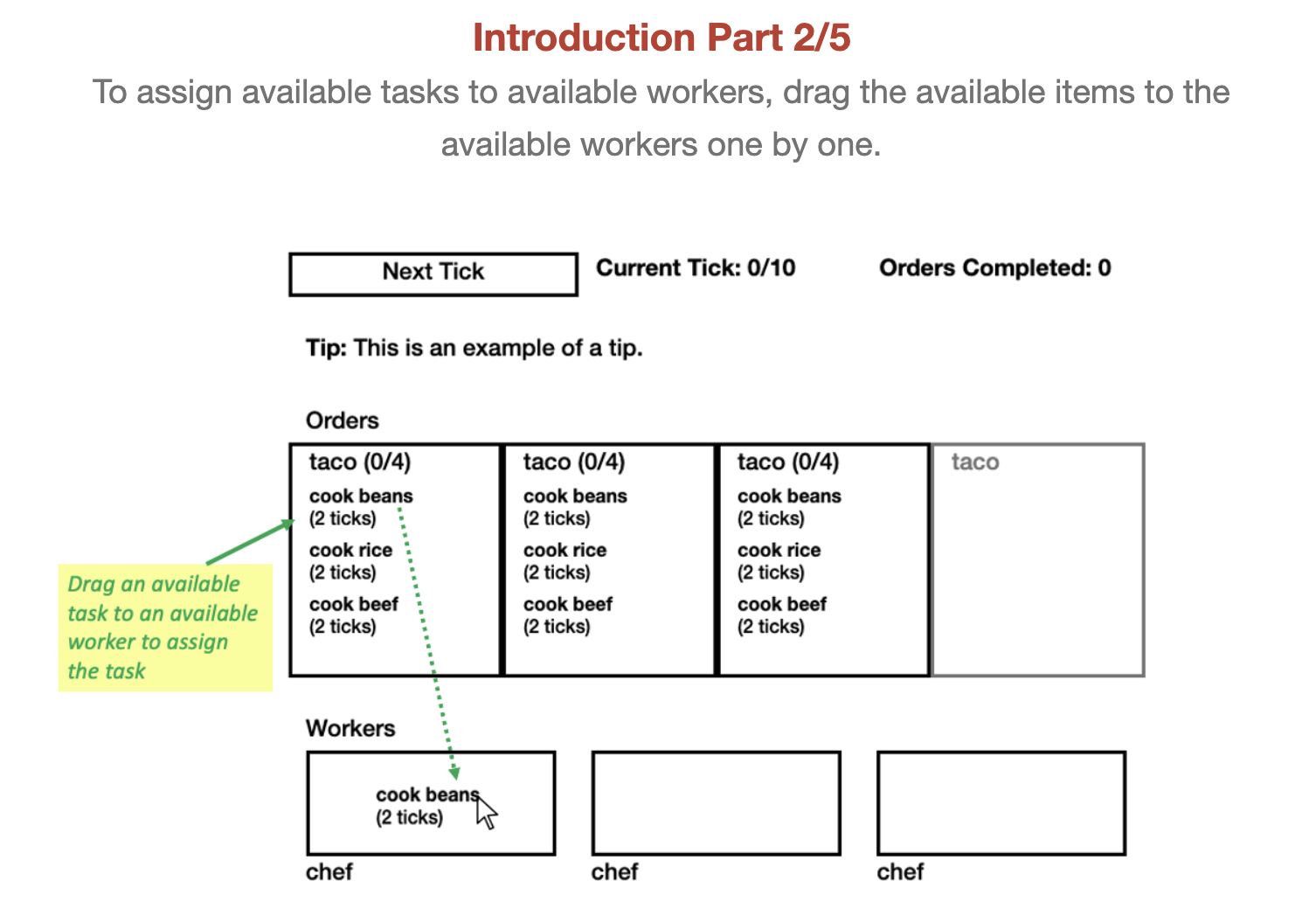}\label{i4}}\quad
\subfloat[Introduction to task assignment (cont.)]{
\includegraphics[width=0.46\columnwidth]{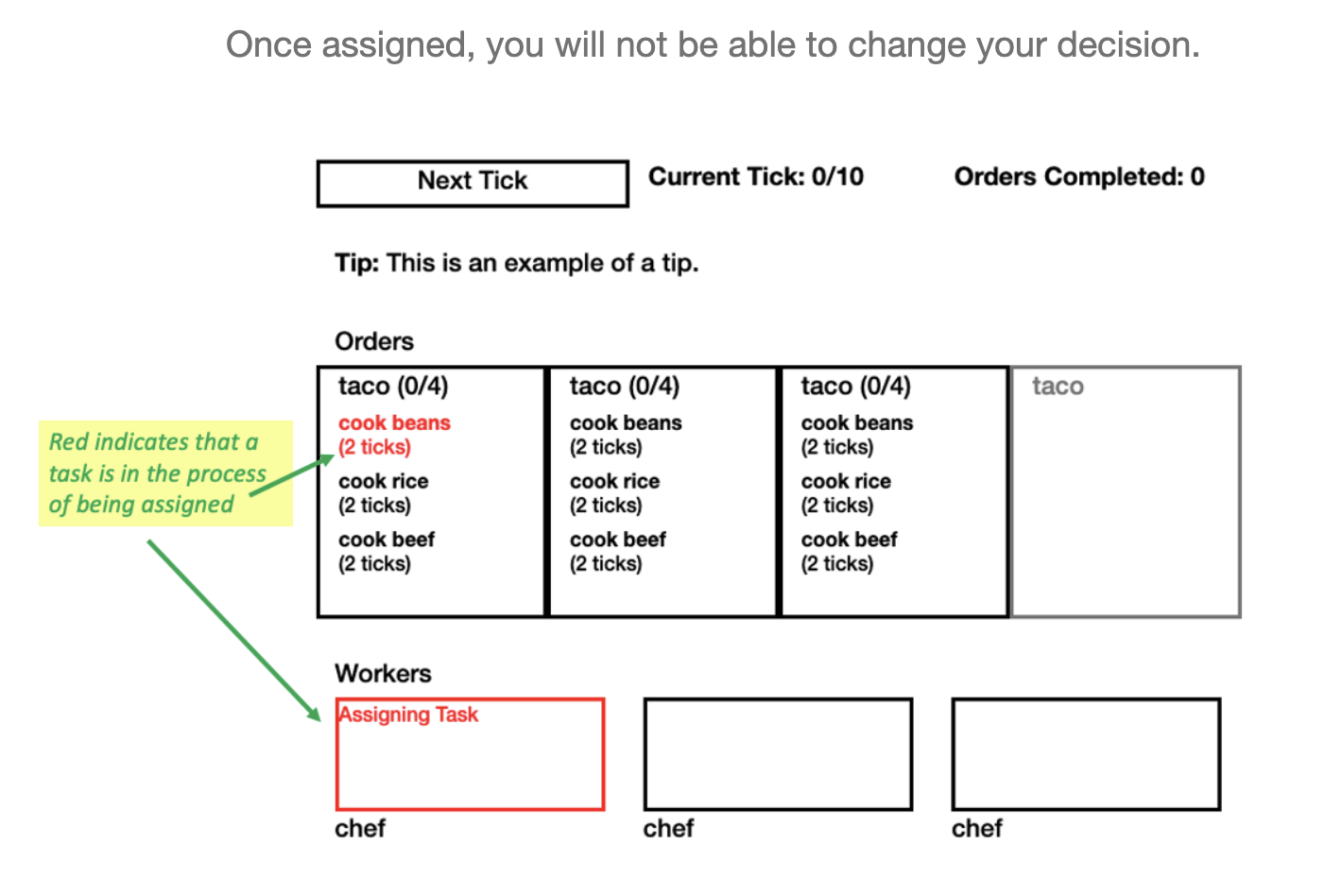}\label{i5}}\\
\subfloat[Introduction to task assignment (cont.)]{
\includegraphics[width=0.46\columnwidth]{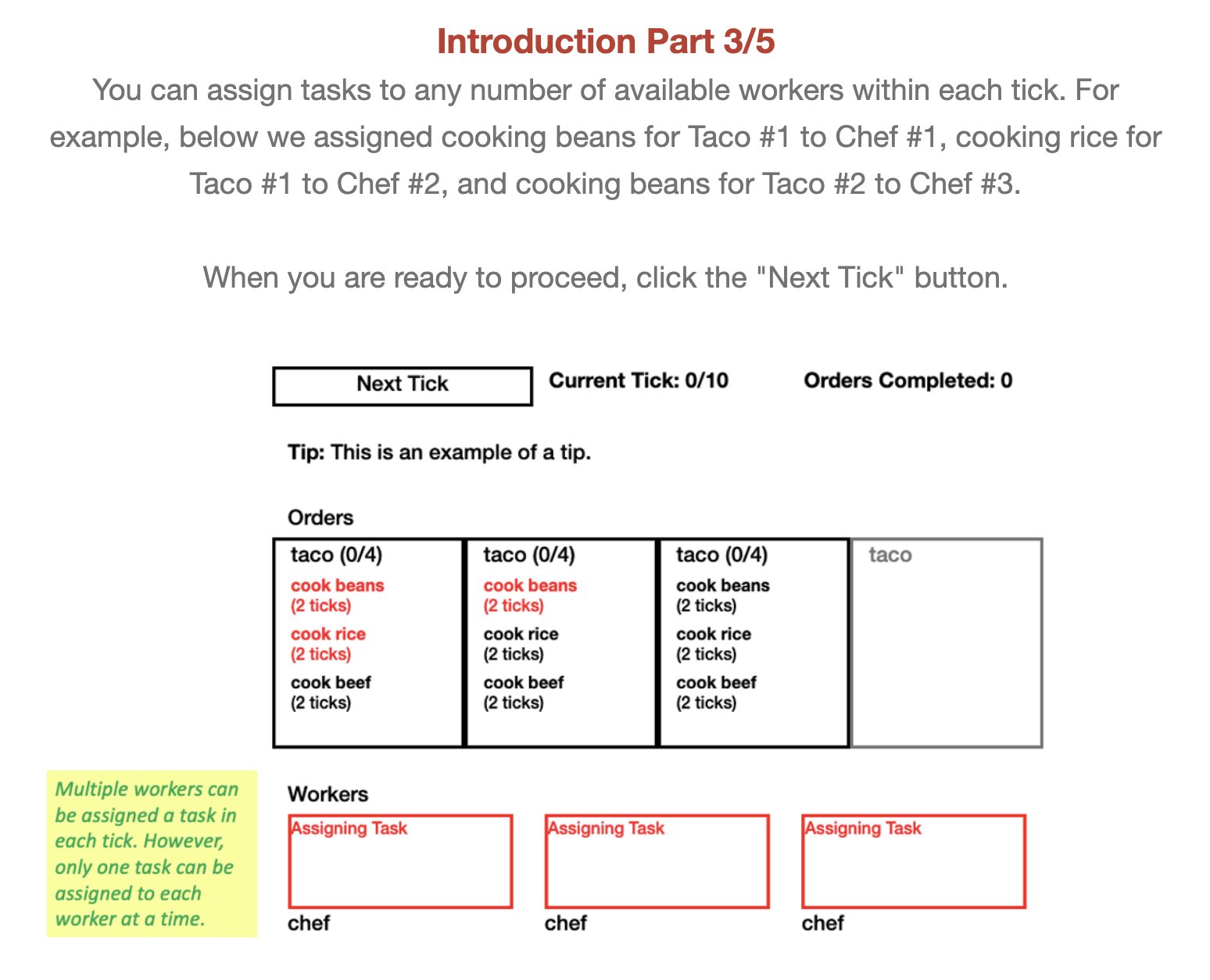}\label{i6}}\quad
\subfloat[Introduction to task assignment (cont.)]{
\includegraphics[width=0.46\columnwidth]{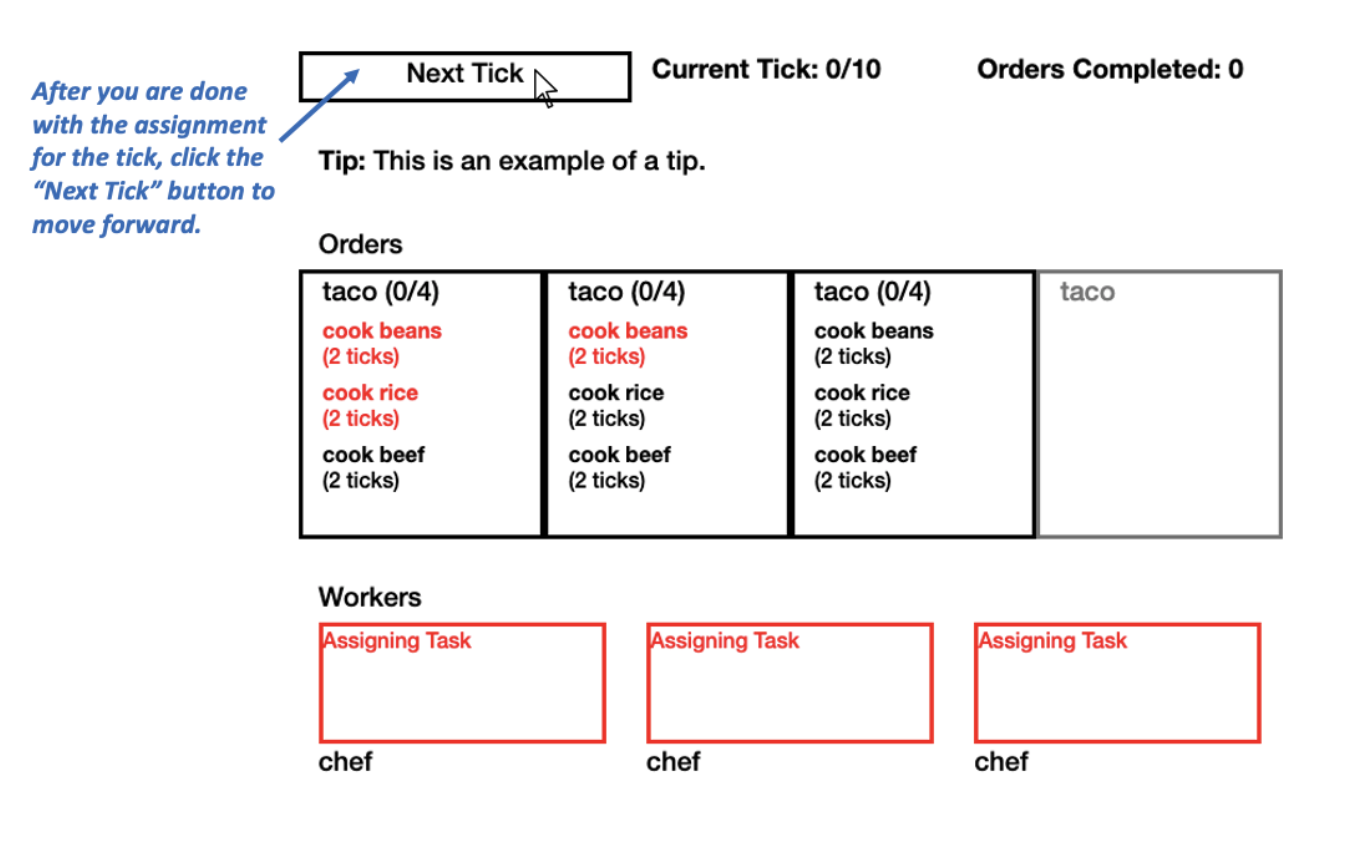}\label{i7}}\quad
\caption{Screenshots of the game introduction.}
\label{fig:intro}
\end{figure}

\begin{figure}
\centering
\subfloat[Introduction to workers' skill levels]{
\includegraphics[width=0.47\columnwidth]{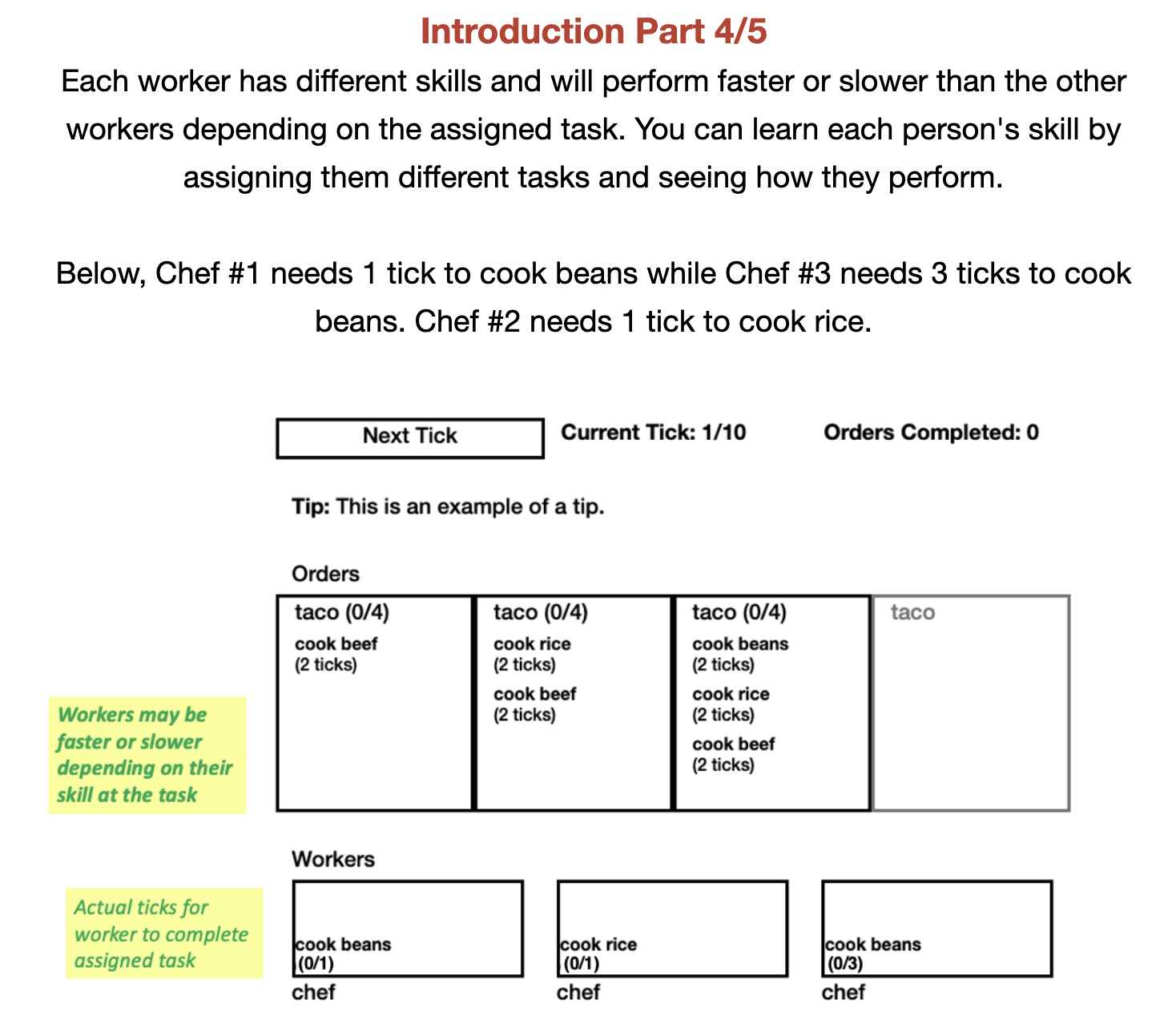}\label{i8}}\\
\subfloat[Introduction to the tip]{
\includegraphics[width=0.47\columnwidth]{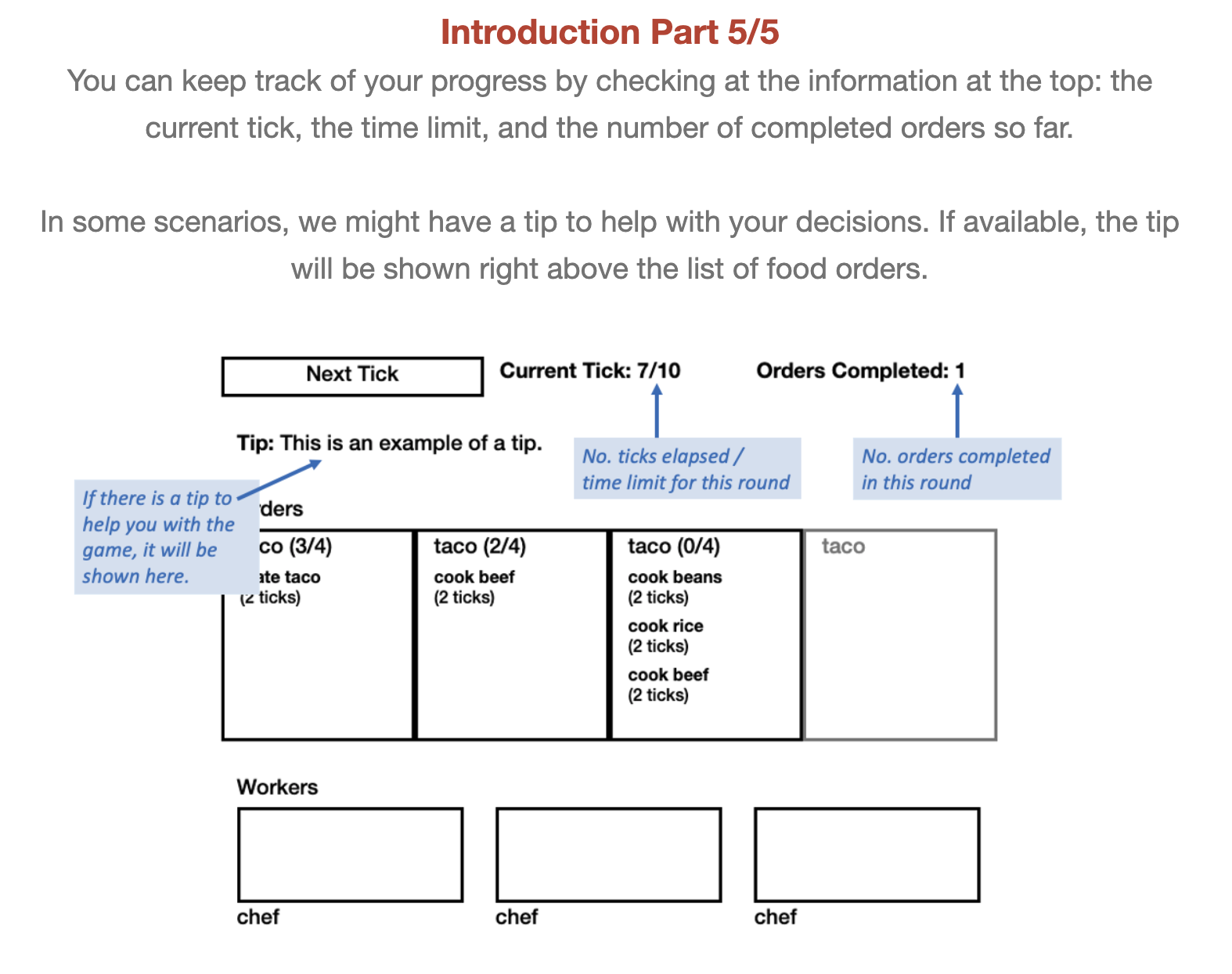}\label{i9}}\quad
\subfloat[Introduction to round completion]{
\includegraphics[width=0.47\columnwidth]{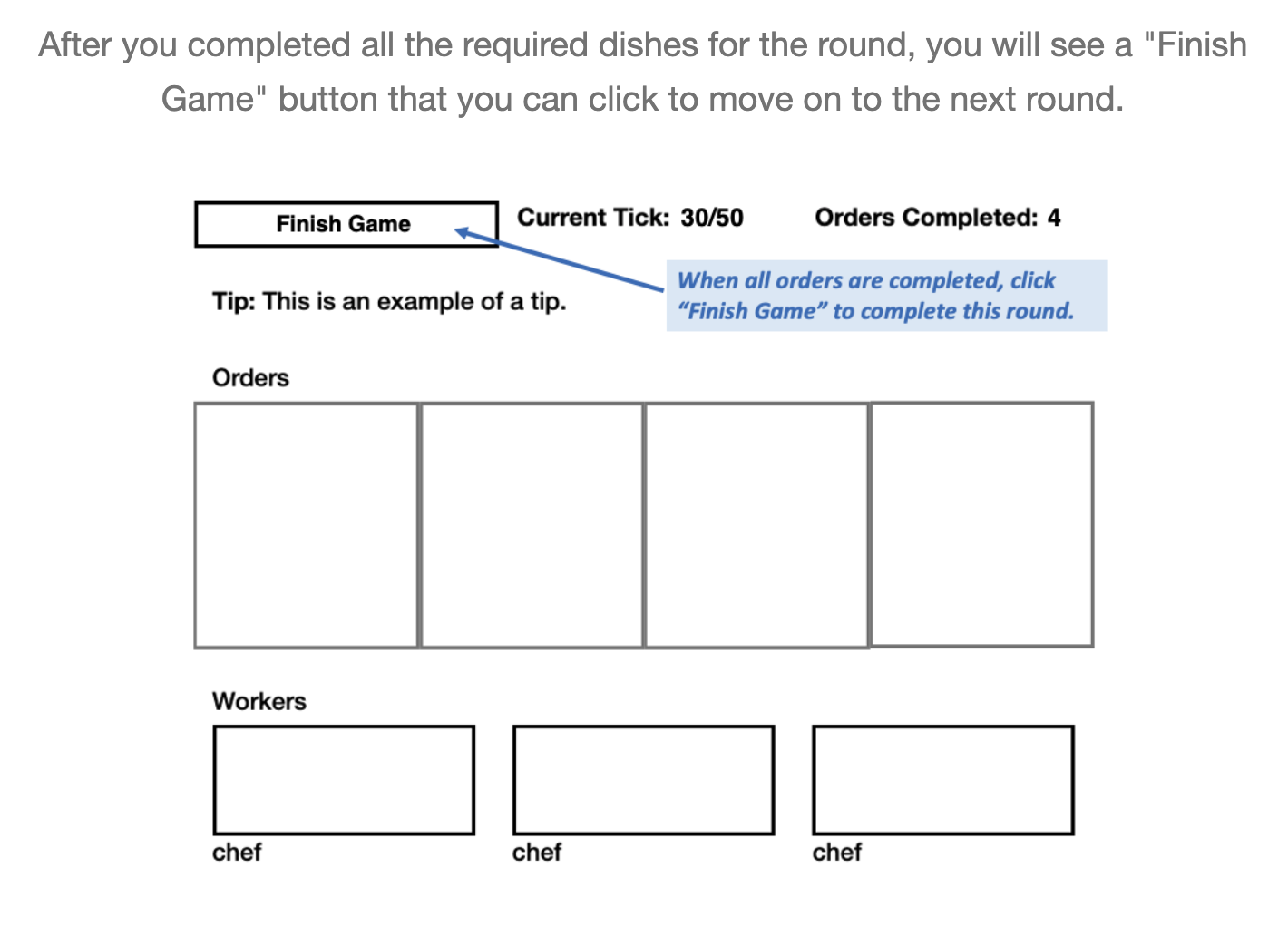}\label{i10}}
\caption{Screenshots of the game introduction (continued).}
\label{fig:intro2}
\end{figure}

\begin{figure}
\centering
\subfloat[Burger's subtasks and available workers]{
\includegraphics[width=0.47\columnwidth]{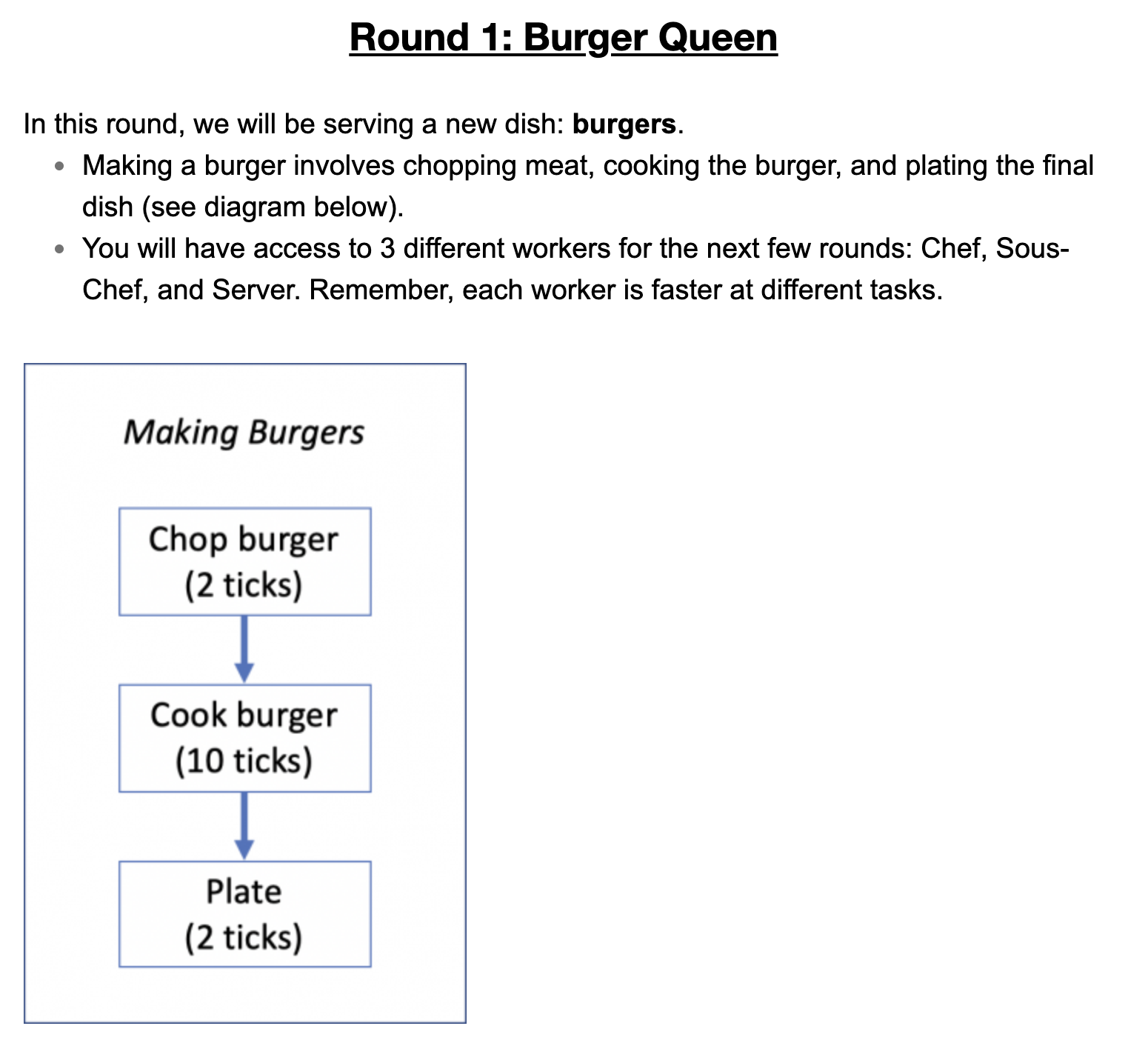}\label{g1}}\quad
\subfloat[Goal, incentives, and reminder]{
\includegraphics[width=0.47\columnwidth]{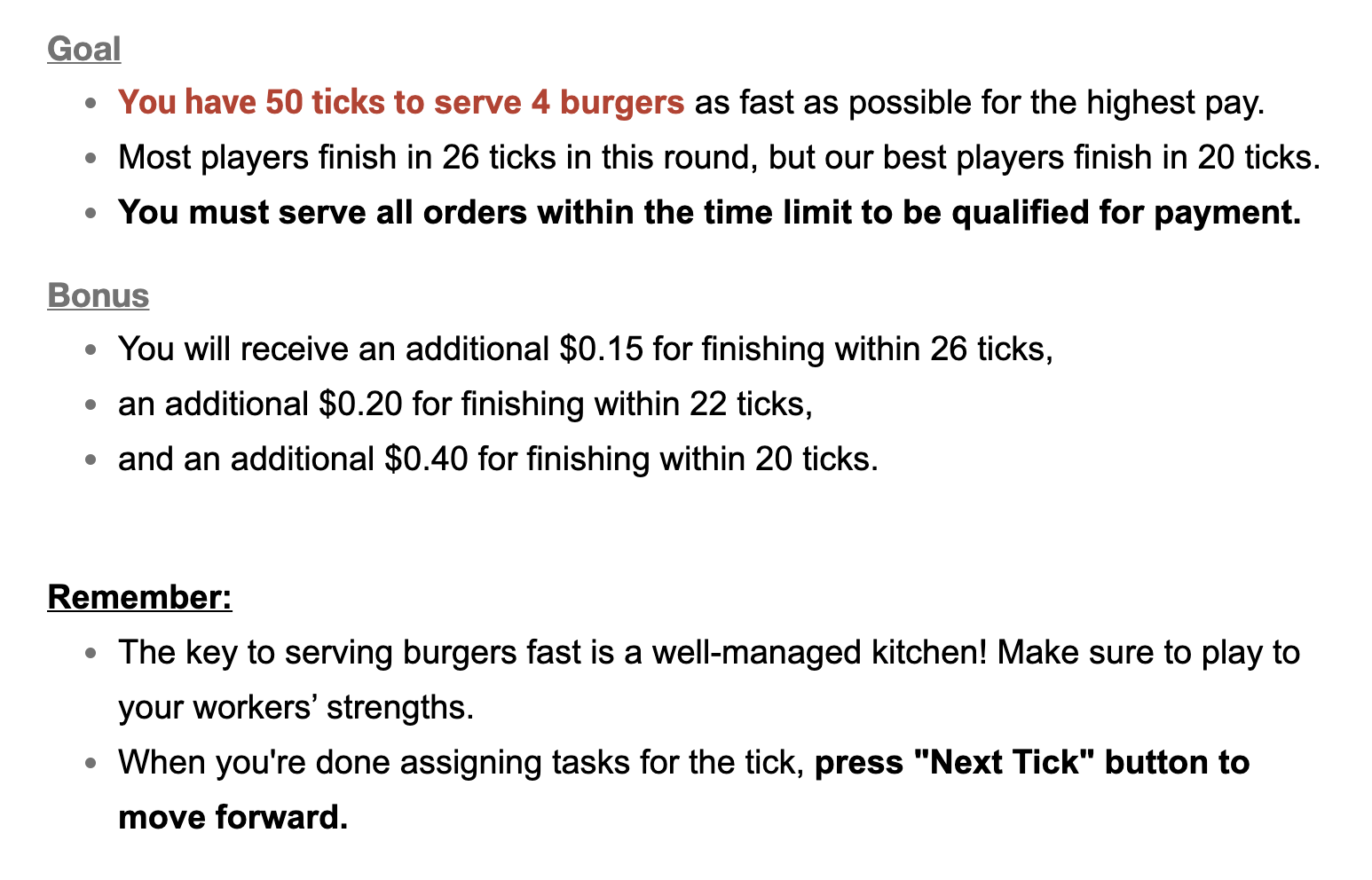}\label{g2}}
\caption{Screenshots of the instructions for the fully-staffed scenario.}
\label{fig:game1}
\end{figure}

\begin{figure}
\centering
\subfloat[Updated instructions following the in-game disruption]{
\includegraphics[width=0.47\columnwidth]{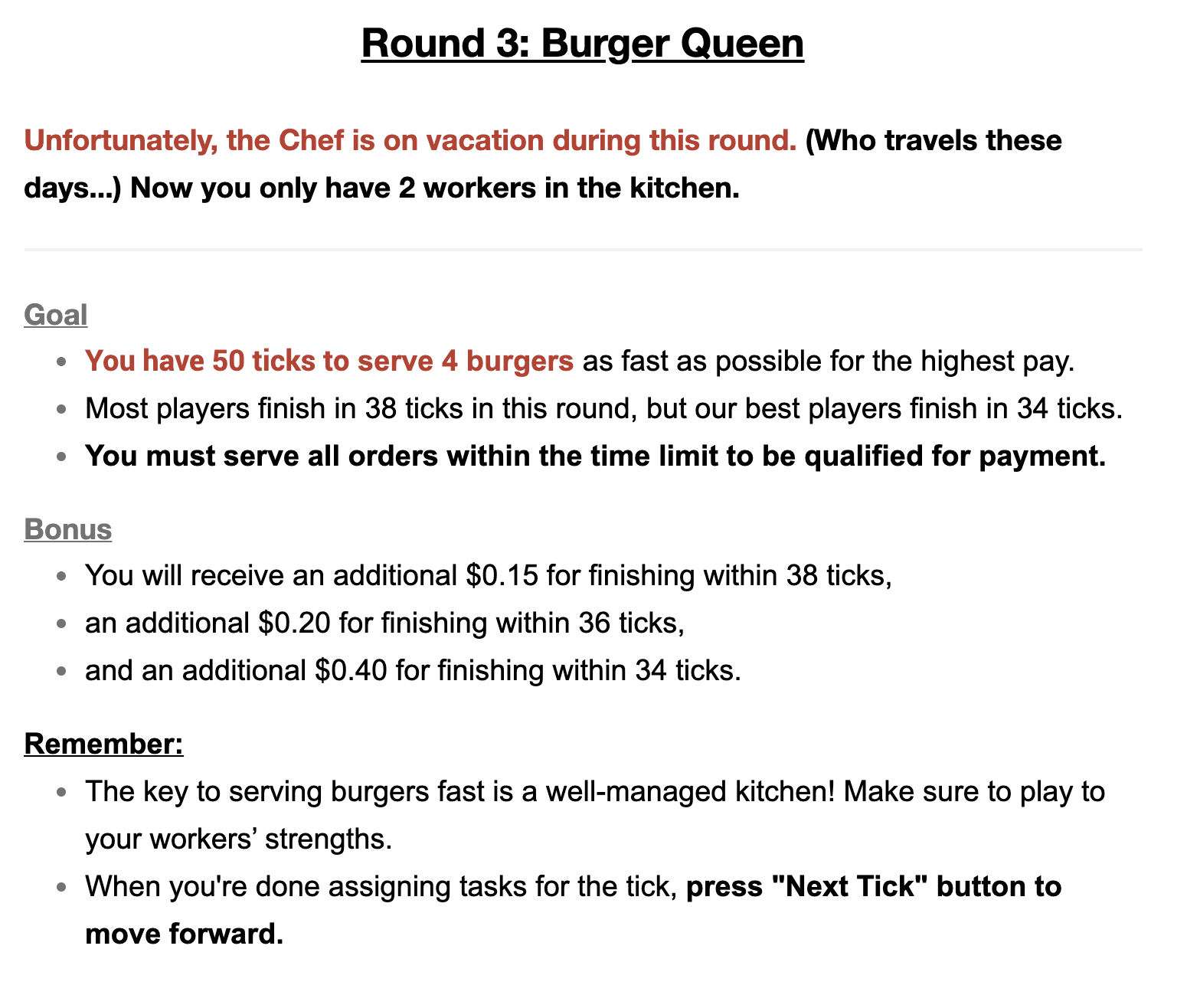}\label{g7}}\quad
\subfloat[Game interface (with the algorithm tip)]{
\includegraphics[width=0.47\columnwidth]{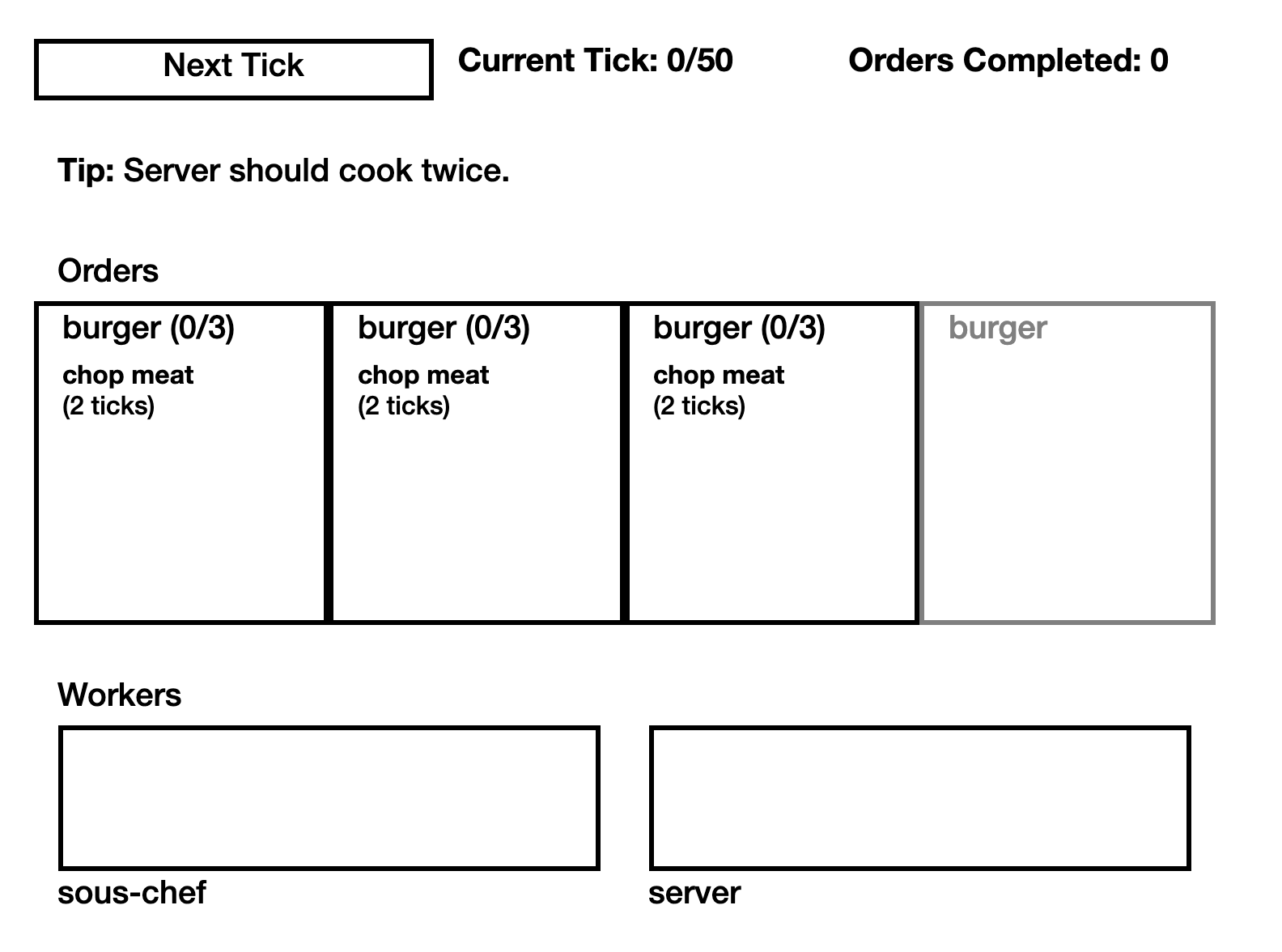}\label{g8}}
\caption{Screenshots of the instructions for the understaffed scenario.}
\label{fig:game3}
\end{figure}

\begin{figure}
\centering
\subfloat[Individual round pay information]{
\includegraphics[width=0.48\columnwidth]{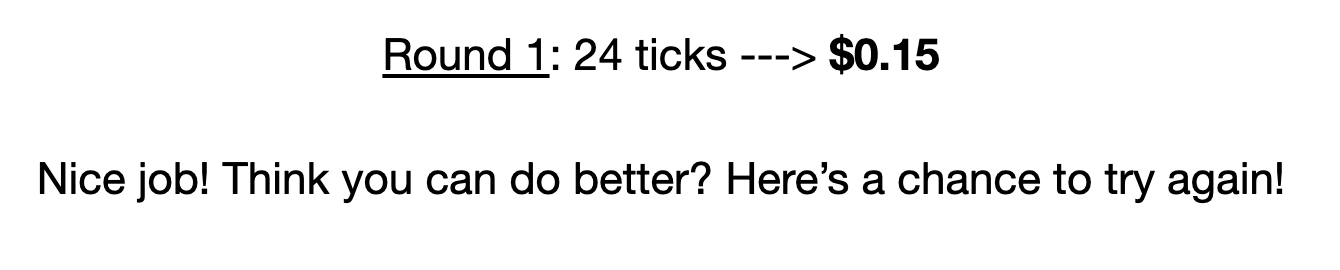}\label{g6}}\quad
\subfloat[Summary of total pay]{
\includegraphics[width=0.2\columnwidth]{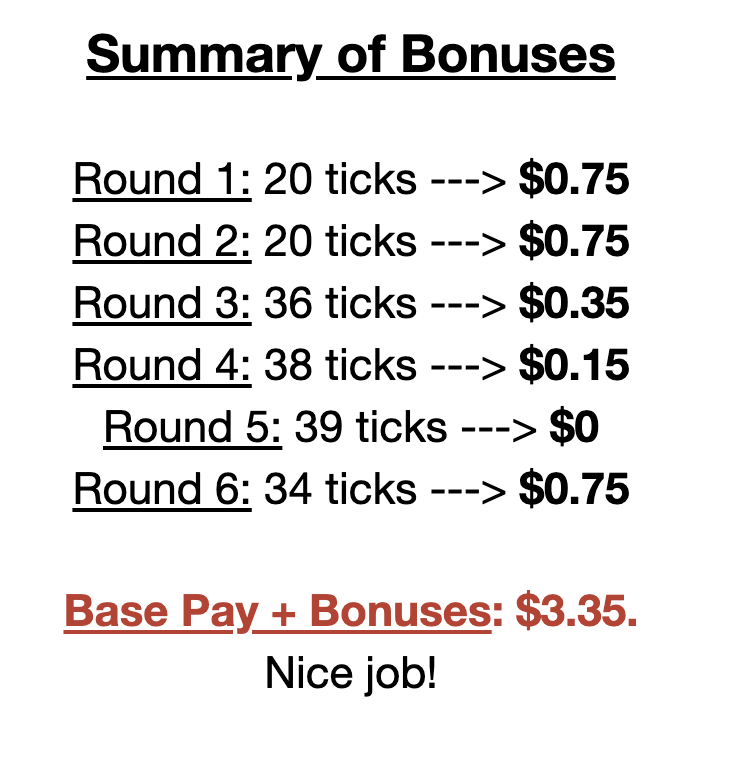}\label{g9}}
\caption{Screenshots of the pay information.}
\label{fig:game4}
\end{figure}
\end{APPENDICES}
\end{document}